\documentclass{article}
\usepackage{amsthm}

\PassOptionsToPackage{numbers}{natbib}
\usepackage[preprint]{neurips_2026}

\usepackage[utf8]{inputenc} 
\usepackage[T1]{fontenc}    
\usepackage{hyperref}       
\usepackage{url}            
\usepackage{booktabs}       
\usepackage{amsfonts}       
\usepackage{nicefrac}       
\usepackage{microtype}      
\usepackage{xcolor}         
\usepackage{graphicx} 
\usepackage{amsfonts}
\usepackage{amsmath}
\usepackage{float}
\usepackage{cleveref}
\usepackage{minitoc}
\usepackage[section]{placeins}
\usepackage{tikz}
\newcommand{\envelopeSymbol}{%
  \begin{tikzpicture}[baseline=0ex]%
    \draw[line width=0.4pt] (0,0) rectangle (1.6ex,1ex);
    \draw[line width=0.4pt] (0,1ex) -- (0.8ex,0.4ex) -- (1.6ex,1ex);
  \end{tikzpicture}%
}

\title{How Much is Brain Data Worth for Machine Learning?}

\author{%
  Lane Lewis$^{1,2,3}$\textsuperscript{\envelopeSymbol} \\
  \texttt{lrlewis@andrew.cmu.edu} \\[0.5em]
  Zhixin Wang$^{4}$ \\
  \texttt{zhixinwa@andrew.cmu.edu} \\[0.5em]
  David Schwab$^{3,5}$ \\
  \texttt{davidjschwab@gmail.com} \\[0.5em]
  Xaq Pitkow$^{1,2,3}$\textsuperscript{\envelopeSymbol} \\
  \texttt{xaq@cmu.edu} \\[1em]
  $^1$Neuroscience Institute, Carnegie Mellon University, Pittsburgh, PA, USA \\
  $^2$Department of Machine Learning, Carnegie Mellon University, Pittsburgh, PA, USA \\
  $^3$NSF AI Institute for Artificial and Natural Intelligence (ARNI) \\
  $^4$Carnegie Mellon University, Pittsburgh, PA, USA \\
  $^5$CUNY Graduate Center, New York, NY, USA\\
  \textsuperscript{\envelopeSymbol} Corresponding authors
}
\newtheorem{theorem}{Theorem}
\newtheorem{lemma}{Lemma}

\theoremstyle{definition}

\theoremstyle{remark}

\begin{document}
\maketitle

\begin{abstract} \label{sec:abstract}
If a person can solve a task, can measuring their brain make it easier to train a model to solve that task too? Recent NeuroAI work suggests that supplementing task training with neural recordings can modestly improve model performance and robustness. However, it is unclear when there should be a benefit from using neural data and how much benefit to expect. We formulate this question mathematically, and begin to address it theoretically using a simple, analytically tractable linear gaussian model of task targets and neural recordings. For a multimodal estimator trained on both brain data and task labels, we derive scaling laws for how performance scales with the numbers of brain and task samples. From these laws we derive relative value and exchange rates between brain samples and task samples, quantifying how much extra task samples neural data is worth as a function of task-brain alignment, neural and task noise, latent dimension, and brain data sample size. We also analyze test distribution shift, to identify conditions where brain-regularized learning can produce substantial robustness gains through learned invariances. Finally, under a fixed collection budget, we characterize the regimes in which brain data is worth collecting. Our results provide a foundation for understanding how valuable brain data could be for improving machine learning.
\end{abstract}

\section{Introduction} \label{sec:intro}
Modern machine learning (ML) systems often improve predictably as training resources scale \cite{kaplan2020scaling}. In many settings, test performance depends systematically on factors such as dataset size, model capacity, and compute, giving rise to empirical and theoretical scaling laws. Understanding these laws is important both scientifically and practically: they help identify which resources are limiting performance and which interventions can most effectively improve sample efficiency and generalization.

A natural question is whether brain data can act as another useful training resource. Humans and animals solve many tasks that overlap with those studied in machine learning, and neural recordings provide a partial view into the internal representations supporting this behavior. This suggests a form of \textit{brain distillation}, in which a learner has access not only to input-output task data, but also to neural measurements from an expert biological system. Recent NeuroAI work has explored this idea by regularizing machine learning models with neural recordings, encoding models, or other brain-derived signals, with several studies reporting modest gains in task performance or robustness \cite{federer2020improved,li2019learning,fong2018using}.

Despite this promise, it remains unclear when brain data should help at all, and how much improvement should be expected. Existing empirical results are often small in magnitude and difficult to interpret: gains may depend strongly on data regime, recording quality, task difficulty, or the alignment between recorded neural features and the task of interest. In some cases, apparent benefits may arise from relatively simple regularization effects rather than from genuinely useful task relevant structure in neural data \cite{li2023robust}. As a result, current empirical work offers limited guidance on basic questions such as: when does brain data improve sample efficiency over task-only learning? How should the value of brain data scale with the number of task labels? What properties of the recordings determine whether brain data is useful? And when is collecting brain data worth its high cost?

In this paper, we study these questions theoretically through a linear gaussian model of task targets and neural recordings. We analyze a multimodal estimator that uses both brain data and task targets, and compare it to task-only learning. Within this model, we derive explicit test error scaling laws in the numbers of brain and task samples. These scaling laws show how brain data can improve task sample efficiency, and how this improvement depends on quantities such as task-brain alignment, neural and task noise levels, neural latent dimension, and the amount of available brain data. We further derive an exchange rate between brain samples and task samples, which quantifies how much task supervision a given amount of neural data is worth. We also analyze test distribution shift, where our brain regularized estimator yields robustness gains by inducing useful invariances, and study a fixed-budget setting to characterize the regime when collecting new brain data makes sense.

Our goal is not to provide a fully realistic model of biological representations or recordings. Rather, we aim to develop a tractable theoretical framework that isolates the main factors governing the value of brain data for machine learning. By making these tradeoffs explicit, our results provide a foundation for understanding when brain data should improve learning, how large those gains will be, and when additional recordings are worth collecting.
\section{Related Work} \label{sec:rw}
\textbf{Brain Distillation.} Brain-inspired machine learning dates back to the earliest stages of the field\cite{rosenblatt1962principles}. Existing approaches span a range of strategies, including biologically inspired architectures and learning rules \cite{hopfield1982neural,fukushima1980neocognitron,lecun2002gradient}, connectomics-based approaches \cite{schmidgall2022biological} as well as choosing models/data based on brain predictiveness \cite{toneva2019interpreting,kubilius2019brain,zhou2024divergences}. Our work is most closely related to a more recent NeuroAI direction that uses neural recordings directly during training to guide machine learning models \cite{federer2020improved,li2019learning,moussa2024improving,vattikonda2025brainwavlm}. This paradigm has the practical advantage of being compatible with standard ML training pipelines and does not require a detailed mechanistic understanding of the underlying neural system.

Within this line of work, several approaches have been explored, including fine-tuning pretrained models on brain data \cite{vattikonda2025brainwavlm,freteault2025alignment,moussa2024improving,federer2020improved}, regularizing task models using neural encoding models \cite{li2019learning}, and using neural data to guide decision boundaries\cite{fong2018using}. Empirical studies have reported modest gains in task performance and robustness in some settings \cite{federer2020improved,li2023robust}. However, these gains are often difficult to interpret, since they may reflect generic regularization effects (such as noise \cite{federer2020improved} or low pass filtering \cite{li2023robust}) rather than genuinely task-relevant information extracted from neural recordings. Recent work further suggests that the value of brain data may be concentrated in low or hard to collect task sample regimes \cite{mineault2026cognitivedarkmattermeasuring}. Despite this empirical literature, there is limited theoretical understanding of when brain data helps, by how much, and which neural signal properties determine its value. Our work addresses this gap by providing explicit scaling law analyses for a type of brain regularized estimator.

\textbf{Scaling Laws.} Scaling laws have played a central role in modern machine learning, especially in language modeling, where they have been used to derive optimal training prescriptions under limited resources \cite{hoffmann2022training,kaplan2020scaling}. Related ideas have also begun to appear in neuroscience, including scaling analyses for brain decoding \cite{ye2023neural,azabou2023unified,banville2025scaling} and language encoding models in fMRI \cite{antonello2023scaling}. Multimodal scaling work further studies how performance depends jointly on multiple data sources \cite{aghajanyan2023scaling}. Our work is distinct in that it studies multimodal scaling over \textit{brain data} and \textit{task data}, and derives an explicit exchange rate between these resources. 

The estimator we analyze is a structured form of generalized ridge regression with a learned positive semidefinite subspace penalty. Ridge and generalized ridge estimators have been studied extensively, including analyses for how the error scales with task samples \cite{hoerl1970ridge,wu2020optimal}. Our estimator is also related to prior work where previous data is used to learn a generalized ridge regularizer for downstream prediction \cite{jin2024meta}, as well as to restricted regression where prediction is constrained or biased toward a lower-dimensional subspace \cite{zhang2010projection,gross2003restricted}. Our setting combines similar ideas: a neural encoding model learned from brain data defines the subspace penalty used for downstream task prediction. To the best of our knowledge this style of two-stage ridge regularization has not been studied previously, nor have theoretical scaling laws been studied over joint brain and task optimization.

\section{Problem Setup} \label{sec:setup}
\begin{figure}
    \centering
    \includegraphics[width=1.0\linewidth]{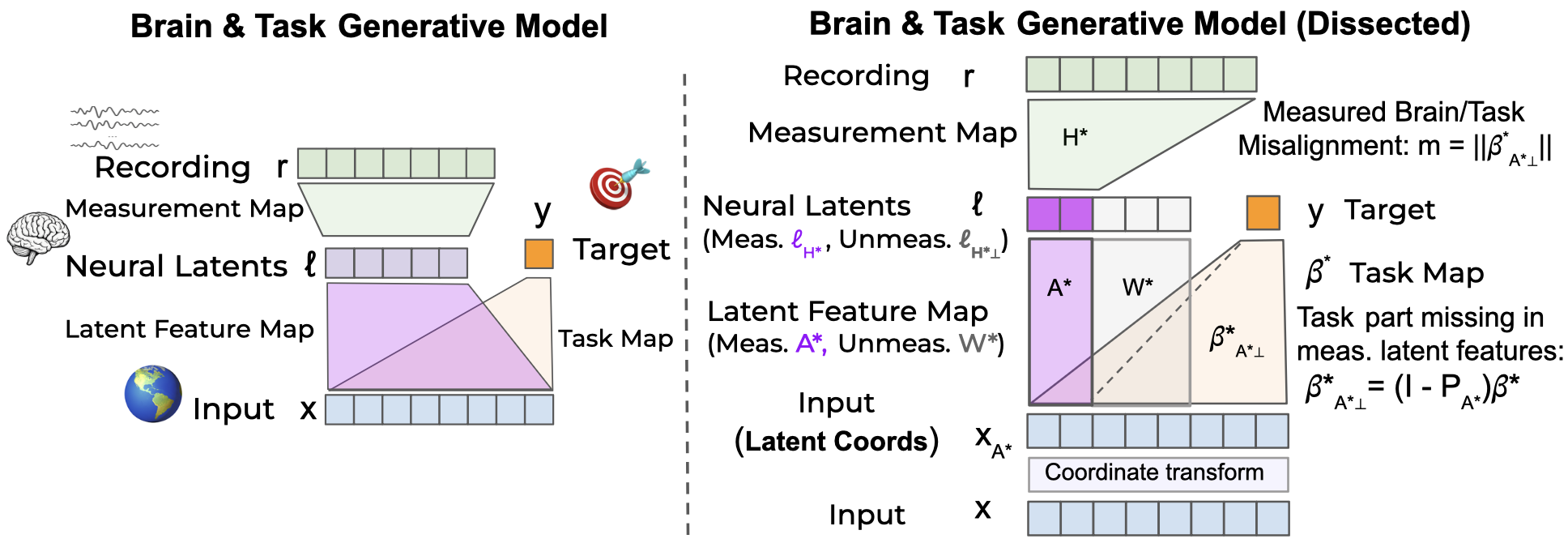}
    \caption{ \textit{Left}: Generative model for brain activity and ML task data. Inputs generate latent representations in the brain which are partially captured by neural recordings. The same inputs drive the response of a task target. \textit{Right:} Brain latents are driven by features that partially capture all relevant task features. Additionally, latents are partially observed through a measurement device. Both effects create misalignment $m$ between the brain and task features.}
    \label{fig:fig1}
\end{figure}
\textbf{Generative Model.} 
Our setting contains four objects: environmental inputs, latent neural features, neural recordings, and task targets. The central idea is that the biological system may contain intermediate representations --- latent neural responses ---  that are lower-dimensional than the input but useful for its own behavior and partially aligned with the target machine learning task. Neural recordings provide only a noisy and partial view of these representations. By ``measured latents'' we indicate the part of the latent representations that are observable with the recording method.

To make this question analytically tractable, we work in a linear-Gaussian model (Figure \ref{fig:fig1}). Although real neural systems and machine learning tasks are highly nonlinear, this model isolates several important statistical factors: task-brain alignment, latent dimension, neural variability, recording noise, task difficulty, and the relative amounts of brain and task data.

For each sample $i$, let $x_i \in \mathbb R^{d_x}$ denote the input, $\ell_i \in \mathbb R^{d_\ell}$ the latent neural representation, $r_i \in \mathbb R^{d_r}$ the neural recording, and $y_i \in \mathbb R$ the task target. We assume $d_x$ and $d_r$ may be large, while the latent dimension $d_\ell$ is smaller. Only a subset of all latent brain features are measured in the recordings due to imperfect capture of the all neural activity. We denote the measured latent subset by $\ell_{H^*, i}\in \mathbb R^{\ell_{H^*, i}}$. Since other latents are not observed in recordings, the relevant generative model components are:
\begin{equation}
\begin{aligned}
x_i &\sim N(0, I_{d_x}), \\
\ell_{H^*,i} &= A^{*T}x_i + \eta_{\ell_{H^*},i}, \\
r_i &= H^{*T}\ell_{H^*,i} + \eta_{r,i}, \\
y_i &= \beta^{*T}x_i + \eta_{y,i},
\end{aligned}
\qquad
\begin{aligned}
\eta_{\ell_{H^*},i} &\sim N(0,\Sigma_{\ell_{H^*}}), \\
\eta_{r,i} &\sim N(0,\sigma_r^2 I_{d_r}), \\
\eta_{y,i} &\sim N(0,\sigma_y^2).
\end{aligned}
\end{equation}
Here $A^* \in \mathbb{R}^{d_x \times d_{\ell_{H^*}}}$ maps inputs to measured latent neural features, $H^* \in \mathbb{R}^{d_{\ell_{H^*}} \times d_r}$ maps measured latents to observed recordings, and $\beta^* \in $ is the ground-truth task predictor (Figure \ref{fig:fig1}). We assume $A^*$ and $H^*$ have rank $d_{\ell_{H^*}}$ and hence are full rank on the subspace of measured latents.

This model separates two sources of noise in neural data. First, the latent representation itself is noisy through $\eta_{{\ell_{H^*}},i}$, which captures variability in the underlying neural state. Second, the recording process is also noisy and potentially higher-dimensional through $H^*$ and $\eta_{r,i}$. As a result, neural recordings need not expose all latent representation structure equally well.

A useful feature of the model is that the task target and the neural representation may be only partially aligned. The target depends on $\beta^*$, while the measured neural latents respond to the subspace spanned by $A^*$.  When $\beta^*$ lies largely in this subspace, the brain contains features that are useful for the task. When $\beta^*$ has substantial mass outside it, the task depends on features that are absent from, or poorly captured by, the recorded neural representation. We quantify the misaligned task features by $\beta_{A_\perp^{*}}^* = (I - P_{A^*})\beta^*$, where the matrix $P_{A^*}$ projects $\beta^*$ onto the measured subspace $A^*$. We can then quantify the misalignment size by
$m = \|\beta^*_{A_\perp^{*}}\|$.
This alignment structure will play a central role in determining the value of brain data.

Note that the parameterization of the latent space is not unique. For any invertible matrix $G$, the transformed parameters $A^{*'} = A^*G$ and $H^{*'} = G^{-1}H^*$ induce the same observable model. Accordingly, only the latent subspace is identifiable. For convenience, we fix a canonical coordinate system in which $A^*$ is orthonormal.

\textbf{Evaluation Setup.}
Given $n$ samples, we write $X$ for the matrix of stacked inputs, $R$ for the stacked neural recordings, and $y$ for the stacked task targets. Let $n_B$ denote the number of brain samples --- pairs of inputs and recorded responses. Let $n_T$ be number of task samples --- pairs of input and task targets. We evaluate predictors in the setting where neural recordings are available only at training time, not at test time. Thus the learned model must ultimately predict targets $y$ from inputs $x$ alone, using knowledge gleaned from neural recordings. We measure performance by mean squared error $\varepsilon$ under a Gaussian test distribution with covariance $\Sigma_{\text{test}}$. For a predictor $\hat\beta$, the test risk is
$$\varepsilon
=
\mathbb{E}\big[(y_{\text{test}}-x_{\text{test}}^\top\hat\beta)^2\big]$$
where $x_{\text{test}}\sim N(0,\Sigma_{\text{test}}), \ \   \eta_{\text{test}}\sim N(0,\sigma_{\text{test}}^2)$, and $y_{\text{test}}=x_{\text{test}}^\top\beta^*+\eta_{\text{test}}$

\textbf{Exchange Rate between Brain Data and Task Data.} To directly evaluate how useful brain data is for solving a task, we define an exchange rate, $\rho$, between the numbers of brain samples and task samples. This exchange rate describes how many extra task samples would be needed for a task-data-only model to match the error of a model trained jointly on brain and task data.
\begin{equation}
\varepsilon(n_B,n_T)=\varepsilon(0,n_T+\rho\cdot n_B)
\end{equation}
We also define the `value' of $n_B$ samples of brain data as the number of additional task samples to reach equivalent performance, $v_T = \rho\cdot n_B$. These quantities provide an interpretable currency of how much brain data helps or hurts learning in units of task samples. In particular, they let us characterize when brain data is useful, how large its benefit is, and how its marginal value changes as more brain data is used.

This quantity can also be converted to a percent `savings' of task data: training with $n_T$ task samples plus $n_B$ brain samples achieves the same test error as a task-only model trained with $n_T + v_T$ task samples. So using brain data along with task data uses only $\frac{n_T}{n_T+v_T} \times 100\%$ of the task samples needed to reach the same performance without using brain data, or equivalently we saved $(1-\frac{n_T}{n_T+v_T})\times100\%$ task data. Many of our figures below show how this savings depends on various parameters.

\section{Results} \label{sec:result}
\begin{figure}
    \centering
    \includegraphics[width=0.9\linewidth]{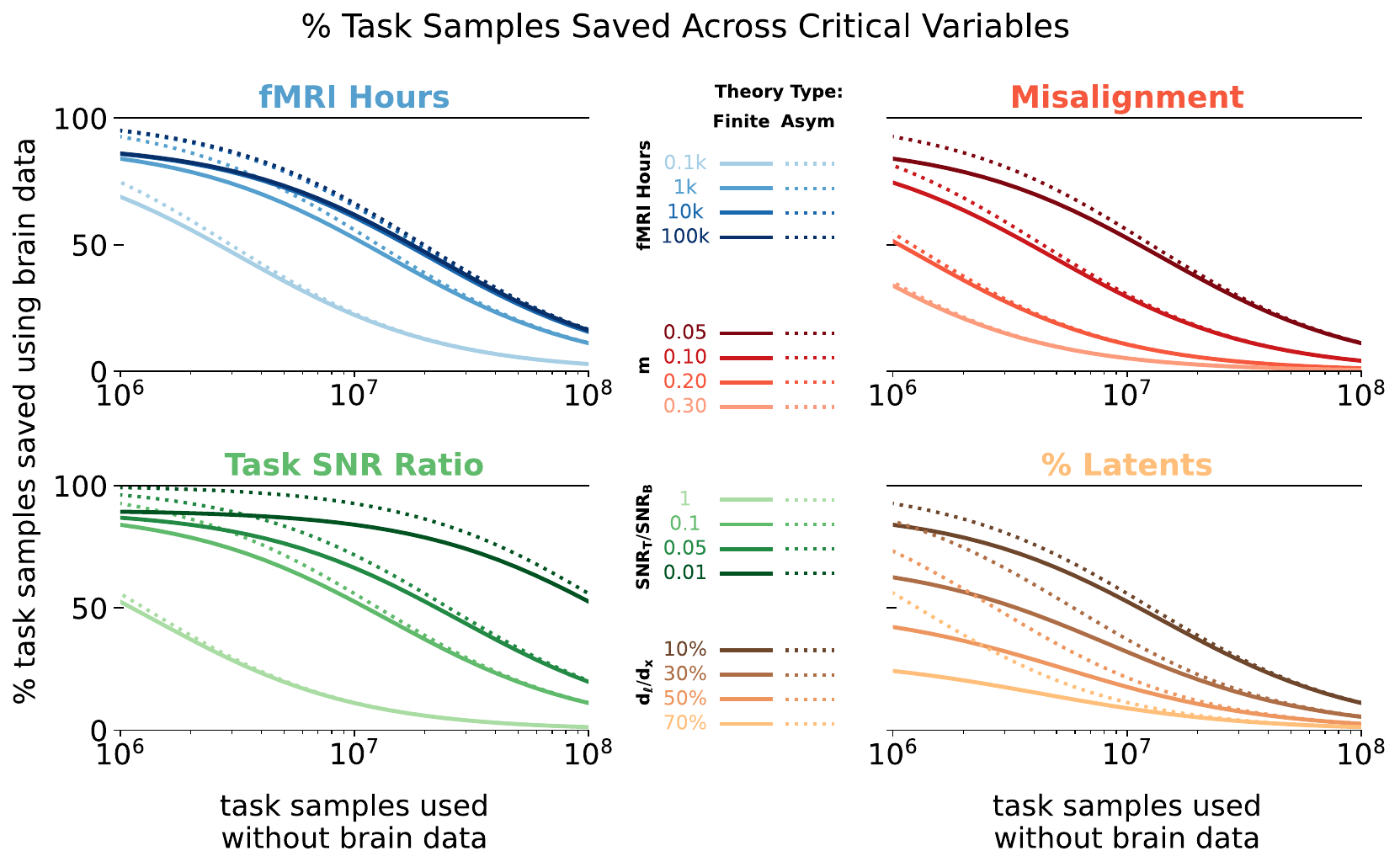}
    \caption{Brain data can substitute for some task data, yielding equal performance while saving a percentage of task samples (dashed lines: asymptotic dependence at large $n_T$ using equation \ref{eq:exchange_rate}; solid lines: finite size corrections using Appendix \cref{thm:total_scaling}). 
    The savings decreases with the number of task samples. 
    Different panels show how the savings depends on various factors (colors) in a simple fMRI model of recordings (Methods \ref{subsec:linear-fmri}). In each panel, the remaining fixed parameters are given by: alignment $m= 0.05$, relative signal to noise ratios $\text{SNR}_{T}/\text{SNR}_{B}=0.1$, fMRI data volume = 1k hrs (1800 brain samples/hr), relative dimensionality $d_x/d_{\ell_{H^*}} = 10\%$. \textit{Top Left}: Increasing brain samples increases task sample savings, but adding more brain data gives diminishing returns. \textit{Top Right}: Better alignment between brain and task increases savings. \textit{Bottom Left}: Increasing task SNR vs brain SNR (Methods \ref{subsec:linear-fmri}) improves savings. \textit{Bottom Right}: Decreasing the latent neural dimension for fixed misalignment produces higher task savings.}
    \label{fig:saved_task_data}
\end{figure}

\textbf{Overview.}
 We analyze the scaling of a particular estimator which uses an encoding model trained on $n_B$ samples to predict neural responses. Internal features from the encoding model are then used to regularize task learning over $n_T$ task pairs. The value and exchange rates are derived under optimal hyperparameters for an isotropic test distribution.
A common constant quantity appearing though our results is $\delta$ (Appendix \ref{app:delta}), a term that depends on the various noises and alignment between brain and task. Ultimately $\delta$ controls the difficulty in using brain data to help solve an ML task.

\textbf{Scaling laws.} We derive multimodal scaling laws for the performance of an estimator trained on both brain and task data (Methods, BEFS). By definition, the scaling law of the estimator with zero brain samples, $\varepsilon(0,n_T)$, is given by the familiar behavior of ordinary least squares training on only task data, which scales as $\sim\sigma_y^2 d_x/n_T$ (Methods, TOS). 

For nonzero brain data, we derive the scaling law for performance as a function of numbers of brain samples and task samples:
$\varepsilon(n_B,n_T) = \varepsilon(0,n_T) - c(\sigma_y, n_B,d_x,d_{\ell_{H^*}},m, \delta)/n_T^2 + o(n_T^{-2})$ (Appendix \cref{thm:optimal_lambda}), where $c$ is a function that captures the dependence on all parameters, and for optimal hyperparameters. 
This scaling law underlies all of the following results. Since empirical simulations for exchange rates are infeasible at the neuroscience-scale sample sizes, we instead use a highly accurate form of our scaling law as a stand-in proxy to characterize non-asymptotic task data regimes. A derivation sketch and full proofs of the scaling laws are provided in the appendix (\ref{app:scaling_law}, \cref{thm:total_scaling}); we also verify our laws empirically in a smaller system \ref{app:sims}.

\textbf{Brain data scaling exchange rate and effective task data value.} We can use the above scaling law to derive an asymptotic exchange rate of brain to task data as well as the exchanged effective task sample value (Appendix, \cref{thm:exchange_rate}):
\begin{equation}\label{eq:exchange_rate}
    \rho = \left(\frac{d_x - d_{\ell_{H^*}}}{d_x}\right)\left(\frac{\sigma_y^2}{n_B[m^2/(d_x - d_{\ell_{H^*}})]   + \delta + o_{n_B}(1)}\right) + o_{n_T}(n_{B}^{-1}), \quad v_{T} = \rho \cdot n_B
\end{equation}
An exchange rate less than 1 indicates that the $n_B$ brain samples are worth less than an equal number of extra task samples for lowering test error. Conversely an exchange rate greater than 1 indicates that these brain samples are more valuable. Our theory suggests that both regimes can occur depending on the quality of the brain data, the difficulty of learning the brain vs learning the task, and how many brain samples are being exchanged. The exchange rate in the large task sample dataset regime depends by the following crucial parameters:
\begin{itemize}
    \item Brain samples $(n_B)$: The exchange rate decreases with brain samples, meaning brain data provides the largest marginal benefits at low to moderate quantities.
    \item Misalignment ($m$): Misalignment critically changes the decay speed of the exchange rate in the number of added brain samples. Additionally, it characterizes the limit of effective extra task sample value of brain data (see below for the limiting expression for $v_T$).
    \item Relative difficulty of learning the task vs the brain ($\sigma_y^2/\delta$): As the relative difficulty of learning the task becomes larger, the exchange rate becomes more favorable. A large ratio allows few brain samples to substitute for many task samples. 
    \item Latent dimension ratio ($d_{\ell_{H^*}}/d_x$): Fewer latent brain dimensions produces better exchange rates. The dimensionality affects the exchange rate by a multiplicative constant, and affects the speed at which the exchange rate decays to zero with brain samples.
\end{itemize}

In the limit of infinite brain and task data, the effective task data value goes to a constant 
$v_T^\infty=\frac{\sigma_y^2 (d_x-d_{\ell_{H^*}})^2}{d_xm^2 }$. Thus for large task samples, savings from brain data drops to zero. Still, for moderate numbers of task samples relative to the input dimension, the key quantities governing the exchange rate can produce substantial savings (Figure \ref{fig:saved_task_data}). 

Our theory predicts that fitting to completely misaligned brain data ($m = \|\beta^*\|$) can still produce a small regularization benefit. This recalls results like \cite{federer2020improved} where fitting to structured noise may explain some of the apparent gains seen from brain regularization empirically.

\textbf{Brain data's value comes from what the brain ignores.}
\begin{figure}
    \centering
    \includegraphics[width=0.9\linewidth]{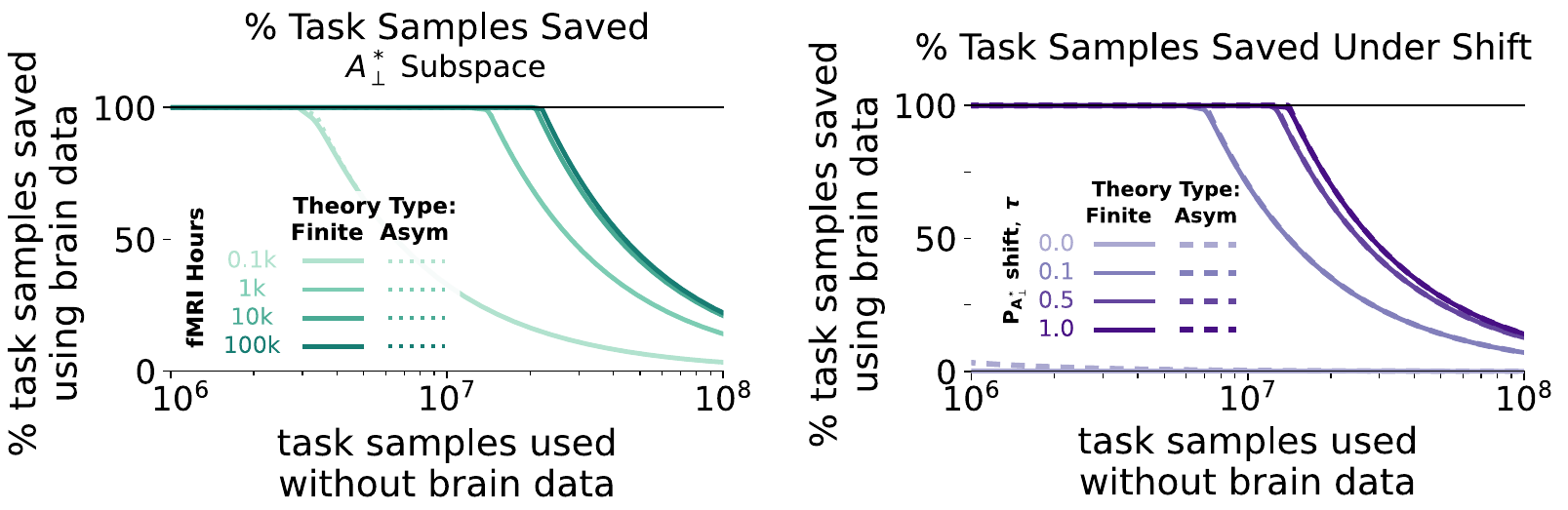}
    \caption{The amount of task data that brain data can substitute changes depending on the test time input covariance distribution. Benefits come from test covariance mass shifted in the brain-insensitive part of the input space, $\operatorname{col}(A^*_\perp)$ (dashed lines: asymptotic dependence at large $n_T$ using Appendix  \cref{lem:robustness_value}; solid lines: finite size corrections using Appendix \cref{thm:total_scaling}). Regularization is chosen optimally for an isotropic test covariance during training. Both panels show data savings in a simple fMRI model of recordings (Methods \ref{subsec:linear-fmri}) with model parameters: $m= 0.05$, relative signal to nosie ratios $\text{SNR}_{T}/\text{SNR}_{B}=0.1$, fMRI data volume = 1k hrs (1800 brain samples/hr), relative dimensionality $d_x/d_{\ell_{H^*}} = 10\%$. \textit{Left panel}: The equivalent task sample value of brain data evaluated under the part of an isotropic distribution in the brain insensitive part of the space provides even greater task sample savings than under an isotropic covariance over all inputs (compare to Figure \ref{fig:saved_task_data} fMRI Hours panel). Task data savings still saturate with large brain data. \textit{Right panel}: The percent task data saved increases as $\tau$ increases and the mass of the test input covariance , $\Sigma_{shift}(\tau) = (1-\tau)P_{A^*} + \tau P_{A^*\perp}$, shifts towards the brain insensitive part of the inputs. Conversely, task data savings become small when most of the test covariance mass is in the brain sensitive part of the input space $\operatorname{col}(A^*)$.}
    \label{fig:robustness}
\end{figure}
How does the value of brain data change across test distributions? Answering this helps clarify what neural data provides beyond in-distribution generalization and what produces its value in the first place. The brain-sensitive subspace is the part of input space to which measured brain activity responds, $\operatorname{col}(A^*)$; the brain-insensitive subspace is the complement to which measured brain activity does not respond, $\operatorname{col}(A^*_\perp)$. Similarly, the task defines task-sensitive and task-insensitive directions in the input. If misalignment is small, then the true task map is approximately contained in the latent features, and task-insensitive directions are partially aligned with brain-insensitive directions. Thus, brain data may approximately reveal a subset of input dimensions to ignore. This makes the brain-insensitive subspace a natural place to look for the source of brain data value.

To analyze where brain data has value, we consider the limit of large sample sizes, and partition the isotropic covariance used in the previous section into the brain-sensitive and brain-insensitive subspaces. 
Surprisingly, in the brain-sensitive subspace, brain data provides no benefit: $\lim_{n_T, n_B\to \infty} v_{T, {A^*}}= 0$. On the brain-insensitive part of the inputs, the value of brain data is even larger (Figure \ref{fig:robustness} left) than under an isotropic test $\lim_{n_T, n_B\to \infty} \ v_{T,{A^*_\perp}}= v_T^\infty\frac{d_x}{d_x - d_{\ell_{H^*}}}$. 

Evaluating under a more general test distribution shift shows a similar effect. Moving mass to the brain-sensitive parts of the space decreases value while increasing mass on the brain-insensitive parts usually increases value (Figure \ref{fig:robustness} right). However, adversarial inputs
can even drive the exchange rate to be negative (Appendix \cref{thm:robustness_on_subspace,thm:robustness_off_subspace}).

\textbf{When should brain data be collected?}
\begin{figure}
    \centering
    \includegraphics[width=0.8\linewidth]{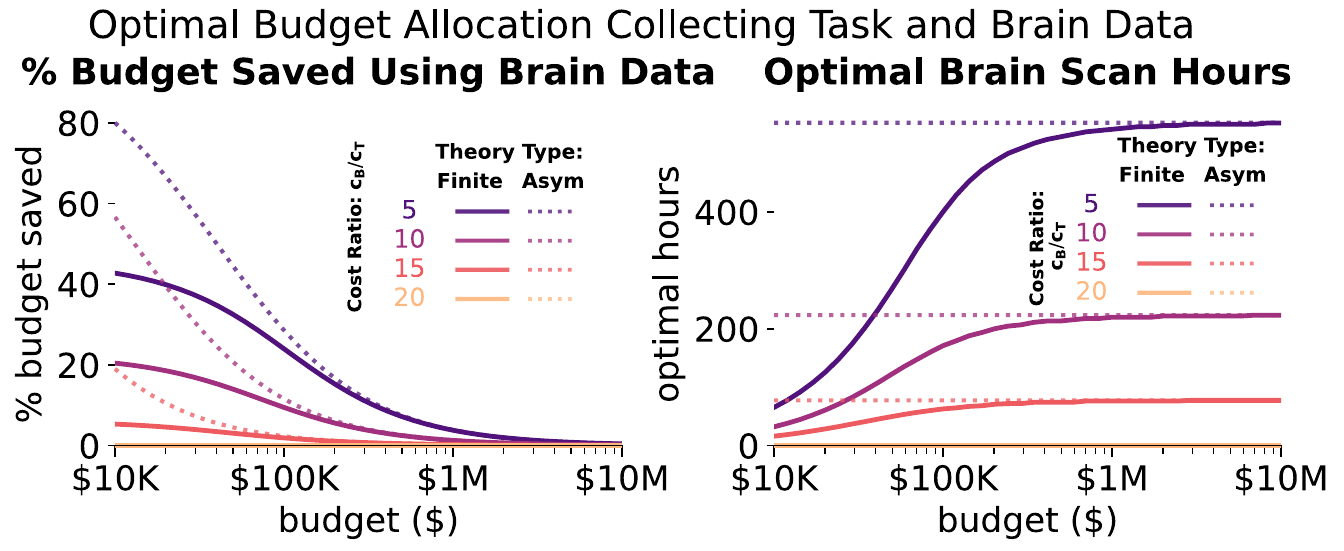}
    \caption{Budget scaling under optimal allocation of task and brain data with different cost ratios: cost of a brain sample from placing a person in a scanner and showing them stimuli $c_B$, over the cost of obtaining a task sample label generated by a human labeler $c_T$. Empirically optimized budget allocation of the joint brain-task scaling law (Appendix \cref{thm:total_scaling}) in solid lines; asymptotic theory \ref{eq:budget} in dotted lines. Linear fMRI parameters ($m=0.05$, $\text{SNR}_T/\text{SNR}_{B} = 0.1$, $d_{\ell_{H^*}}/d_x = 10\%$, $c_T = \$15/\$1800$- \$15 an hour at 1 label every 2 seconds). \textit{Left Panel}: The percent of budget saved with brain data drops in both budget and cost ratio. At a high enough cost ratio (in this case $c_B/c_T = 20$), no brain data should be collected, hence no budget savings. A realistic fMRI ratio in this setting would be $\$500/\$15\approx 33$, in which case we predict no brain data should be collected. \textit{Right Panel}: The optimal number of hours to collect saturates in large budget. Even under large budget, brain data should only be collected in relatively small quantities.}
    \label{fig:budget}
\end{figure}
Suppose you have a budget to solve a problem, but brain data isn't available for a desired stimulus set yet. Should you spend your budget to collect brain data in order to improve your ML model, or should you spend that budget on collecting even more task data? In real neuroscience data collection, high fidelity recordings from the brain are expensive, however the dollar cost of collection depends on the method used and recording quality: EEG data may be cheap but noisy while inter-cranial data is much more precious but more accurate. We could also collect task labels from humans (e.g. Amazon Mechanical Turk for naming images). 
We denote the cost of collecting a stimulus-brain response pair $c_B$, the cost of collecting an input-label pair $c_T$, and the total \$ budget $\mathcal{B}$. We show that an estimator trained using brain and task data under a fixed budget can give the same test error as one that only uses task data at a larger budget \ref{eq:budget}. The amount of budget savings is driven by a brain-favorability equation, $F$, which measures how good conditions are for brain data  (bigger is more favorable) and depends on the cost ratio of task vs brain data collection, dimensionality savings, and the relative task learning difficulty for the brain and task.
\begin{equation}\label{eq:favorability}
    F = \frac{c_T}{c_B}\left(\frac{d_x - d_{\ell_{H^*}}}{d_x}\right)
\frac{\sigma_y^2}{\delta}
\end{equation}
Non-zero amounts of brain data should be collected under large budget when the following conditions hold: $ F > 1$ and $\delta > 0$ (Appendix \cref{thm:budget_scaling}).
Under these conditions, we show that brain data buys you an equivalent extra amount of budget to spend on task data collection, giving budget savings for equal performance (Figure \ref{fig:budget} left, Equation \ref{eq:budget} left). This quantity behaves asymptotically like a constant, so the total percent budget saved drops to zero in a large budget. The equivalent extra budget increases with brain favorability and depends on the value of brain data for an isotropic test and on the cost of a task sample (Appendix \cref{thm:extra_budget}).
Finally, we show that the amount of brain samples that should be collected, $n_{B}^{opt}$, asymptotes in large budget and increases as the brain data becomes more favorable to collect, (Figure \ref{fig:budget} right).
\begin{equation}\label{eq:budget}
    \text{Equiv. Extra Task \$} = c_Tv_T^\infty\left[1-
\sqrt{1/F}
\right]^2 + o_{\mathcal B}(1),\quad n^{opt}_{B}=\frac{d_x-d_{\ell_{H^*}}}{m^2}
\left[
\sqrt{\text{F}
}
-
\delta
\right]+o_{\mathcal B}(1)
\end{equation}
Hence, brain data should be collected only under narrow conditions on the cost, and only as a small auxiliary dataset. Under current high cost neuroscience data collection limitations, there must be significant savings in dimensionality and a large difference in the task-brain learning difficulty to justify brain data collection. Given the challenge of obtaining neural data, this can be seen as a benefit --- it may not need to be collected in massive quantities to obtain most of its value.

\section{Methods} \label{sec:method}
\begin{figure}
    \centering
    \includegraphics[width=0.6\linewidth]{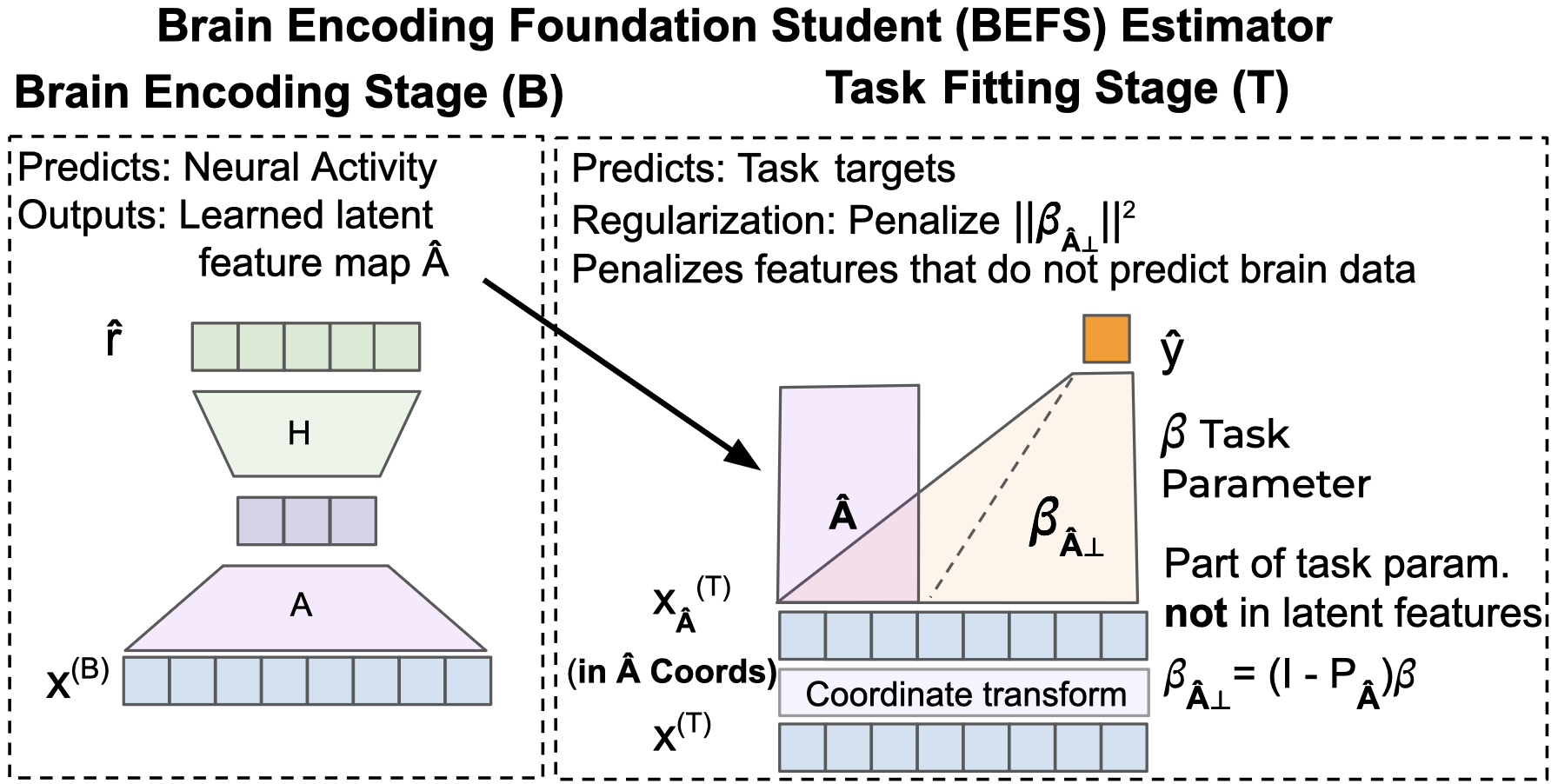}
    \includegraphics[width=0.39\linewidth]{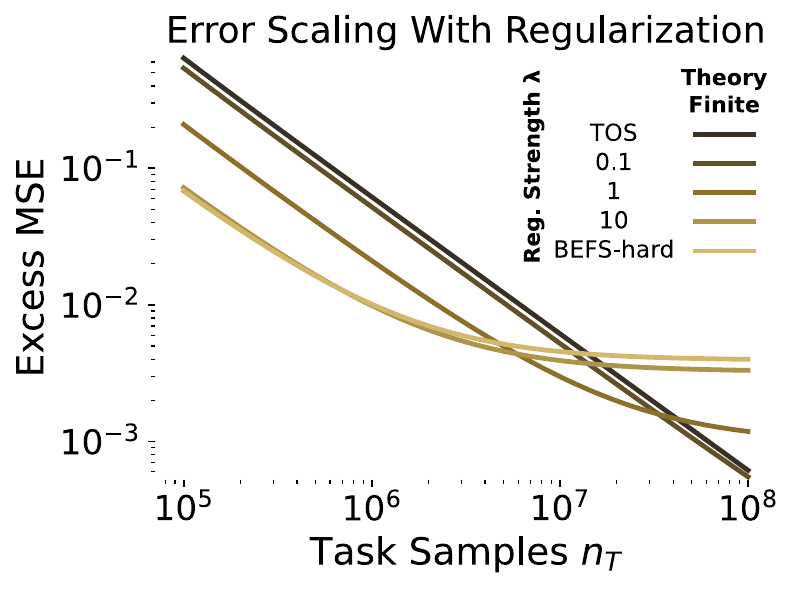}
    \caption{\textit{Left Panel} BEFS estimator model configuration. In the first stage, an encoding model is learned to predict neural activity in recordings using an autoencoder. In the second stage, the learned brain features are used to regularize task learning. \textit{Right Panel} BEFS test error scaling over regularization $\lambda$. A strong fixed regularization under a low misalignment improves test error at low task samples. However, eventually this regularization leads to a floor which gives worse performance than a task only model. This figure uses a linear fMRI model (Methods \ref{subsec:linear-fmri}) with parameters of $m= 0.05$, $\text{SNR}_{T}/\text{SNR}_{B}=0.1$, fMRI = 1k hrs (1800 brain samples/hr) and $d_x/d_{\ell_{H^*}} = 10\%$}
    \label{fig:befs}
\end{figure}
We compare a task only baseline to a two-stage estimator that uses neural recordings and task labels.

\textbf{Task Only Student (TOS)} To characterize the baseline of learning with zero brain data, we construct a task only student estimator that learns only from paired inputs and task targets. Given $n_T$ task samples, $(X, y)$, the estimator is ordinary least squares, $\hat \beta^{TOS} = \text{argmin}_{\beta} \frac{1}{n}\|y - X\beta\|^2$.
This estimator serves as the reference point for quantifying the task data value of brain data.

\textbf{Brain Encoding Foundation Student (BEFS)} We next consider a two-stage estimator that uses neural recordings to learn features and then uses them to regularize downstream task learning (Figure \ref{fig:befs}). This construction is motivated by empirical NeuroAI approaches where a neural encoding model is first learned from stimulus-response data and then used to guide a task model.

Brain Encoding Stage: 
In the brain encoding stage, the learner observes $n_B$ paired inputs and neural recordings, giving the dataset $(X^{(B)}, R^{(B)})$. It fits a low-rank linear encoding model by solving 
\begin{equation}
   \hat A, \hat H = \text{argmin}_{A, H} \frac{1}{n_B}\|R^{(B)} - X^{(B)}AH\|^2 
\end{equation}
Here $\hat A$ represents the learned latent feature map from inputs to a low-dimensional neural representation, while $\hat H$ maps these learned latents to observed recordings. Throughout this work, we assume the latent dimension is known, correctly specified, so that $\hat A \in \mathbb{R}^{d_x \times d_{\ell_{H^*}}}$.

Task Stage - 
In the task stage, the learner observes $n_T$ paired inputs and task targets, giving the dataset $(X^{(T)}, y^{(T)})$. The learned latent feature space from the brain stage is then used to regularize the task predictor. We encourage alignment of learned task features to brain features by penalizing task components that lie outside the learned feature. Mathematically, we write this regularization penalty as $\|(I - P_{\hat A})\beta\|$, the projection of the task parameters onto the non-brain predictive features. 

A hard constraint version of this estimator forces the task predictor to lie only in the learned neural feature space. A softer version replaces this constraint with a quadratic penalty:
\begin{equation}
    \hat\beta^{BEFS}
=
\arg\min_{\beta}
\frac{1}{n_T}\|y^{(T)} - X^{(T)}\beta\|^2
+
\lambda \|(I - P_{\hat A})\beta\|^2
\end{equation}
This is a generalized ridge objective with penalty matrix $I - P_{\hat A}$. The parameter $\lambda$ controls the strength of alignment to the learned neural features. As $\lambda \to 0$, the estimator approaches the task only student behavior. As $\lambda$ becomes large, it approaches the behavior of the hard constraint (Appendix, \cref{thm:BEFS_hard_lambda}). A fixed positive lambda can produce useful test error benefits by shrinking a subset of task dimensions, however this eventually becomes detrimental as task samples increase (Figure \ref{fig:befs}).

\textbf{Interpretation}
The BEFS estimator biases learning toward task predictors that are supported on features useful for explaining neural recordings. Its benefit depends on two factors: how accurately the brain stage recovers the neural subspace, and how strongly the task aligns with that subspace.

\textbf{Value derivation sketch}
TOS has the scaling law of ordinary least squares and BEFS has the scaling law under optimal regularization of $\varepsilon(n_B,n_T) = \varepsilon(0,n_T) - c(\sigma_y, n_B,d_x,d_{\ell_{H^*}},m, \delta)/n_T^2 + o(n_T^{-2})$ for a fixed function $c$ of critical parameters such as noise and dimensionality.
Adding a fixed number of \textit{extra} task samples, $\Delta _T$, to the TOS at large $n_T$ also produces a quadratic correction.
$
\varepsilon(0, n_T+\Delta_{T})
=
\varepsilon(0,n_T)
-
 \Delta_{T}\sigma_y^2d_x/n_T^2
+
o(n_T^{-2})
$
Equating the second order $n_T$ corrections lets us solve for the asymptotic exchange rate $\Delta_{T} \approx c/(\sigma_y^2d_x) = v_T = \rho n_B$. Similar style of proofs produce the results obtained for the test shift and  budget results. See Appendix for details.
\subsection{Linear fMRI Model}\label{subsec:linear-fmri}
To obtain coarse scaling predictions in a regime roughly matched to modern visual fMRI, we use a stylized linear simulation of voxel responses. This is a major simplification of real fMRI, but it lets us ask what scaling behavior would arise if stimulus-to-voxel responses were approximately linear. We use input dimension 4096, corresponding to 64 by 64 images, latent dimension 410, and 10,000 stimulus-sensitive voxels. We calibrate the variance so that 40\% of single-trial variance is stimulus driven, while the remaining 60\% is split into 40\% measurement noise and 20\% neural variability. This toy calibration is broadly consistent with recent visual fMRI datasets reporting roughly 20\%–60\% stimulus-driven single-trial variance \cite{gifford20267}, and with modeling results showing that measurement noise is a significant contributor to prediction error\cite{prince2022improving}. For data collection, we assume one stimulus response every 2 seconds, corresponding to 1800 samples per hour. We define the SNR of the task as $\text{SNR}_{T} = \|\beta^*\|^2/\sigma_y^2$ and the SNR of the brain as the average channel SNR, $\text{SNR}_{B} = 1/d_r\cdot\sum^{d_r}_{i=1}(H^{*\top}H^*)_{ii}/(H^{*\top}\Sigma_\ell H^{*} + \sigma_r^2I)_{ii}$. For the downstream task, we fix the true task vector to have norm 1 and vary label noise to change SNR. The main-text simulations use deliberately brain-favorable regimes. For additional details see Appendix \ref{app:extra_theory_figs}.

\section{Discussion} \label{sec:discussion}
How much is brain data worth for ML? Our work suggests that brain data has some worth in task sample efficiency, however its value is highly dependent on the training data regime, testing distribution shift and critical parameters like the misalignment of the recorded brain and task. We suggest that brain data is most valuable in small to moderate amounts when solving the task is much harder than estimating the brain, and when a small number of highly task-aligned latents are well exposed by or selected from a brain recording. We also demonstrate that the benefits are best seen at low to moderate task samples. Through this work, we provide foundational results for more complex theory to build on as well as provide initial guiding principles for empirical NeuroAI practitioners. 

The obvious limitation of our work is that we analyzed an analytically tractable linear model in simplified settings while real neural data and tasks are highly nonlinear and operate on far more complicated distributions. Still, simple linear theory can expose a surprising number of useful learning structures seen in nonlinear settings \cite{saxe2013exact,ding2024understanding,schaeffer2023double}. Despite our model's simplicity, we were able to capture several qualitative behaviors observed in the NeuroAI literature. We are able to demonstrate that brain data can improve robustness, which is claimed to be a dominant reason to perform brain distillation \cite{mineault2024neuroai}. We also show that fitting to uninformative brain data can produce structured noise regularization effects that can lead to apparent performance benefits \cite{federer2020improved}. Additionally, we find similar results to suggestions from recent perspective papers that brain data should be used when task data is very difficult to collect or hard \cite{mineault2026cognitivedarkmattermeasuring}. In future work, we seek to extend this theory to nonlinear settings and investigate scaling on real neural data in the regimes explored in this paper.

While our application in this problem was NeuroAI, our method generally characterizes a form of noisy, partially observed knowledge distillation. We believe our work could be extended to distillation in ML generally for cases when performing full knowledge distillation may be too computationally expensive given model sizes. Our theory would provide insight into performance from passing a more efficient, corrupted partial view of teacher representations to the student during learning.

\textbf{Author Contributions}: Conceptualization, XP; methodology, LL, ZW; software, LL, ZW; writing—original draft preparation, LL; writing—review and editing, LL, ZW, DS, XP; visualization, LL, XP; supervision, XP, DS. All authors have read and agreed to the published version of the manuscript.

\textbf{Acknowledgements}:
This work is supported through funds to XP and DS provided by the National Science Foundation and DoD OUSD (R \& E) under Cooperative Agreement PHY-2229929 (The NSF AI Institute for Artificial and Natural Intelligence, ARNI).

\bibliographystyle{unsrt}
\bibliography{bibliography}


\newpage
\raggedbottom
\section*{Appendix}
\addcontentsline{toc}{section}{Appendix}
\appendix
\section{Code}\label{app:code}
All code used to run simulations and generate the figures is provided at \url{https://github.com/LaneLewis/brain-distillation-theory}. The codebase contains a readme with the commands used to generate the figures as well as additional figures not shown.
\section{Simulations}\label{app:sims}
To provide evidence of our theory tracking empirically, we perform simulations on a smaller scale than those given in the main paper. The reason for this is that estimating brain data values as we have defined them is numerically unstable as it requires solving an inverse expression in $n_T$ to describe an empirically averaged estimated risk. So, at the scale described in the paper, the number of samples and dimensionality of the different components makes empirical curves infeasible. Provided below are simulations for $d_x = 8, d_r=5, d_r = 8$ and a task SNR of $1.0$. We use the same latent pooling measurement matrix as the linear-fMRI model results presented in the paper.

To empirically estimate the error $\varepsilon$, we averaged the closed form of the error over independent draws from the generative model and fitting $\hat \beta^{BEFS}$. We call the number of independent dataset draws the number of trials. Additionally, we performed multiple replicate runs with different random seeds to obtain mean and confidence interval statistics. For most of the simulations, we additionally fit $\lambda$ empirically by fitting on a log spaced grid of lambda points and choosing the lambda with lowest estimated test error. The only simulation where we did not do this was for the budget simulations. In the budget simulations, we empirically estimated the risk over many different feasible cost samples of $n_B$ and $n_T$ for the conditions on $c_B, c_T$. We used the theoretically optimal $\lambda$ in this case since the double grid search was too expensive and we had previously verified that empirical and theoretical lambda schedules track very closely. The estimated $n_B, n_T$ combination that produced the lowest test error was kept as the optimal allocation and used to derive the budget scaling results. All runs use an average over 30 separate run replications to generate a mean estimate and a 95\% confidence interval through boostrapping. 

We used CPUs to run all our simulations. In total for the empirical simulations shown here, around 2 days on 64 HPC CPUs with 32GB of memory for non-budget simulations and 400GB of memory for budget ones.
\begin{figure}
    \centering
    \includegraphics[width=0.9\linewidth]{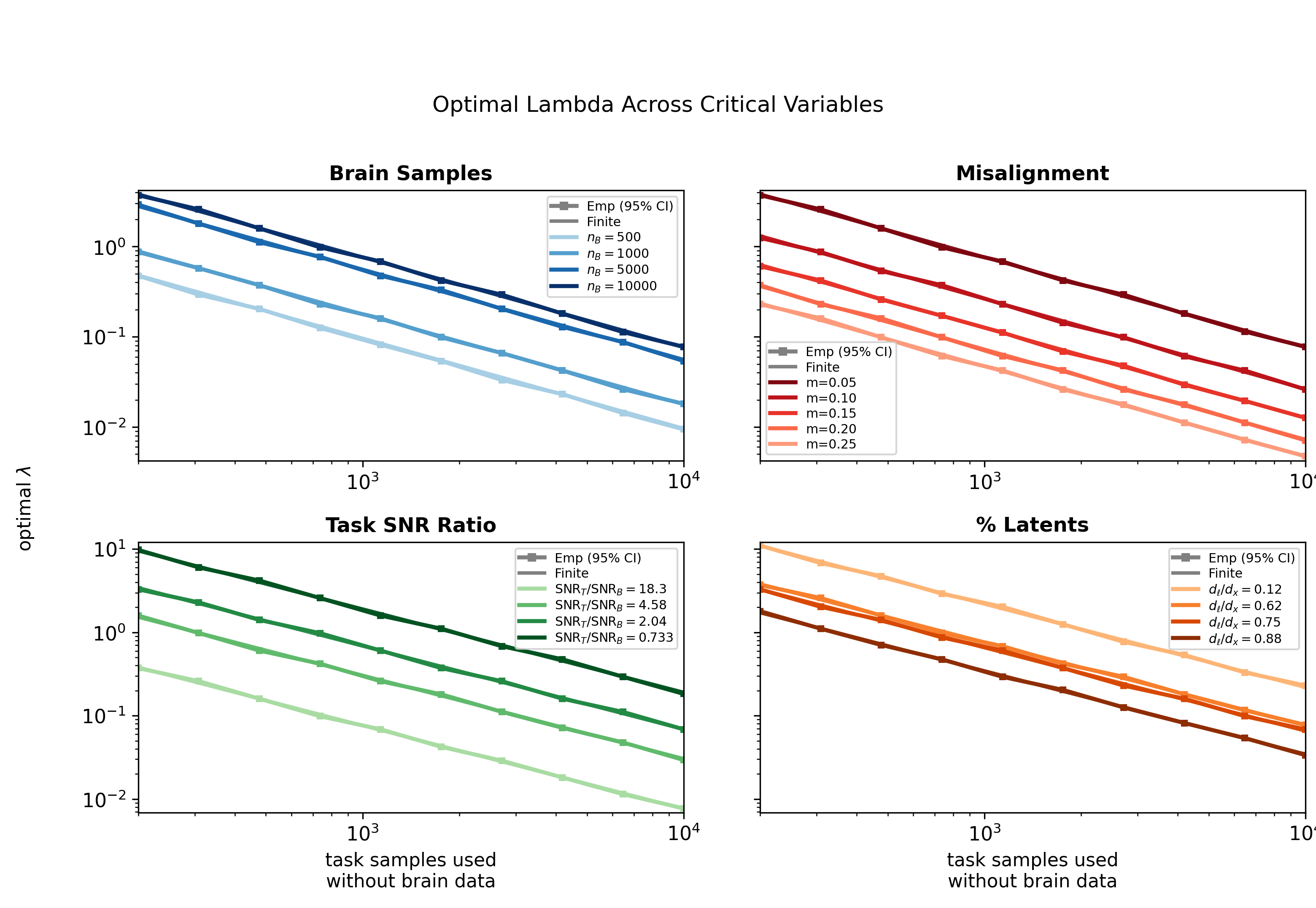}
    \caption{Empirically fit optimal $\lambda$ empirically matches the theoretical schedule derived in \cref{thm:optimal_lambda}. 100k independent trials used to generate each replicate and 30 replicates averaged to generate the mean and confidence interval. Parameters used: $m=0.05 $, $\text{SNR}_T/\text{SNR}_B=1.83$, $n_B =10000$ samples $d_{{\ell_{H^*}}}/d_x=62\%$, $100,000$ trials. Empirical curves (Emp) are plotted as solid with a square at evaluated points with confidence intervals, asymptotic curves (Asym) \cref{eq:exchange_rate} are plotted dashed, and finite sample theory curves (Finite) \cref{thm:total_scaling} are plotted in solid. }
    \label{fig:empirical_optimal_lambda}
\end{figure}
\begin{figure}
    \centering
    \includegraphics[width=1.0\linewidth]{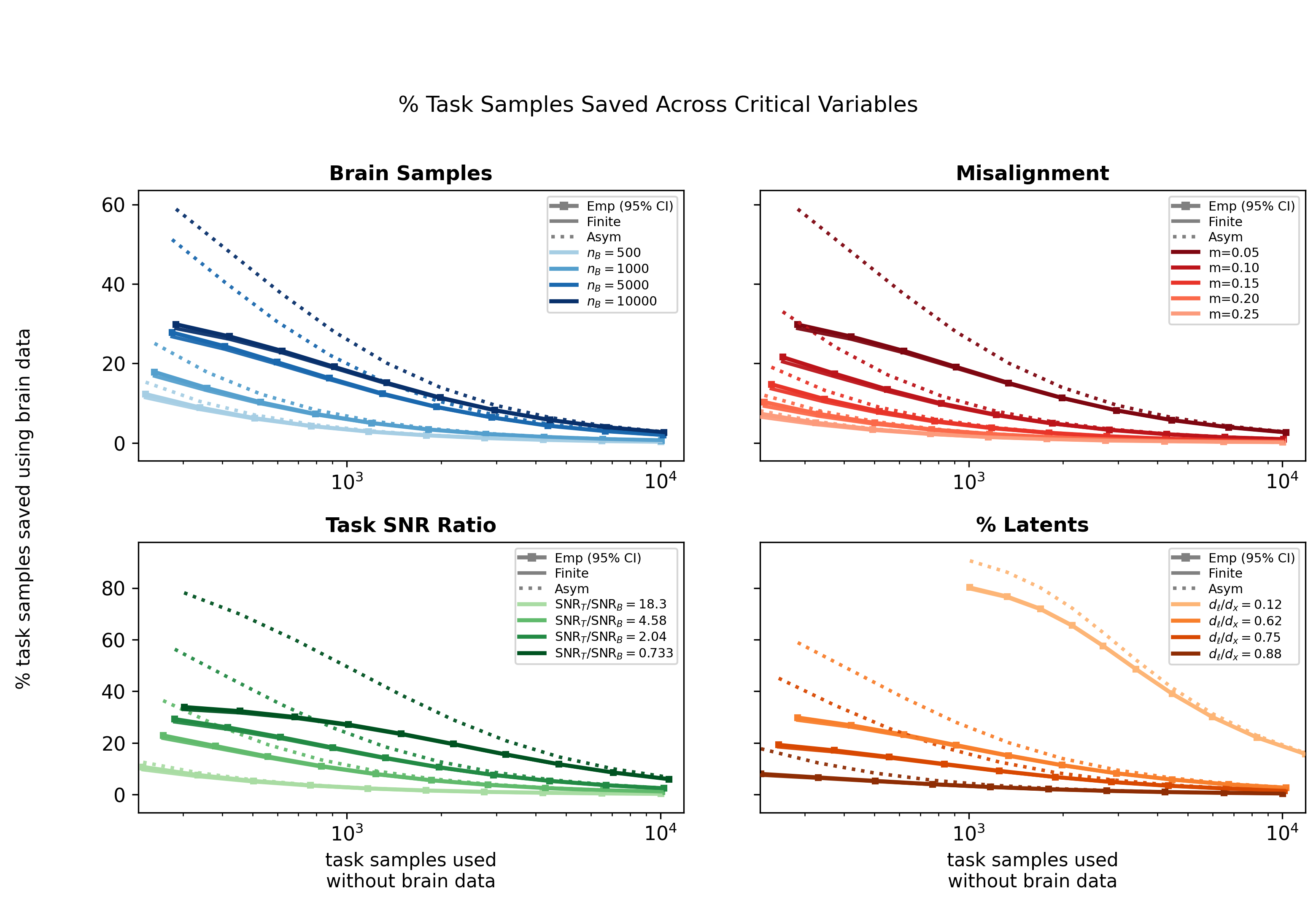}
    \caption{Empirically fit data savings match the finite sample theory curves (\cref{thm:total_scaling}) even at moderate task samples. 100k independent trials were used to generate each MSE estimate replicate and 30 replicates were averaged to generate the mean and confidence interval. Parameters used: $m=0.05 $, $\text{SNR}_T/\text{SNR}_B=1.83$, $n_B =10000$ samples $d_{{\ell_{H^*}}}/d_x=62\%$, $100,000$ trials. Empirical curves (Emp) are plotted as solid with a square at evaluated points with confidence intervals, asymptotic curves (Asym) \cref{eq:exchange_rate} are plotted dashed, and finite sample theory curves (Finite) \cref{thm:total_scaling} are plotted in solid. }
    \label{fig:empirical_percent_task_saved}
\end{figure}
\begin{figure}
    \centering
    \includegraphics[width=0.49\linewidth]{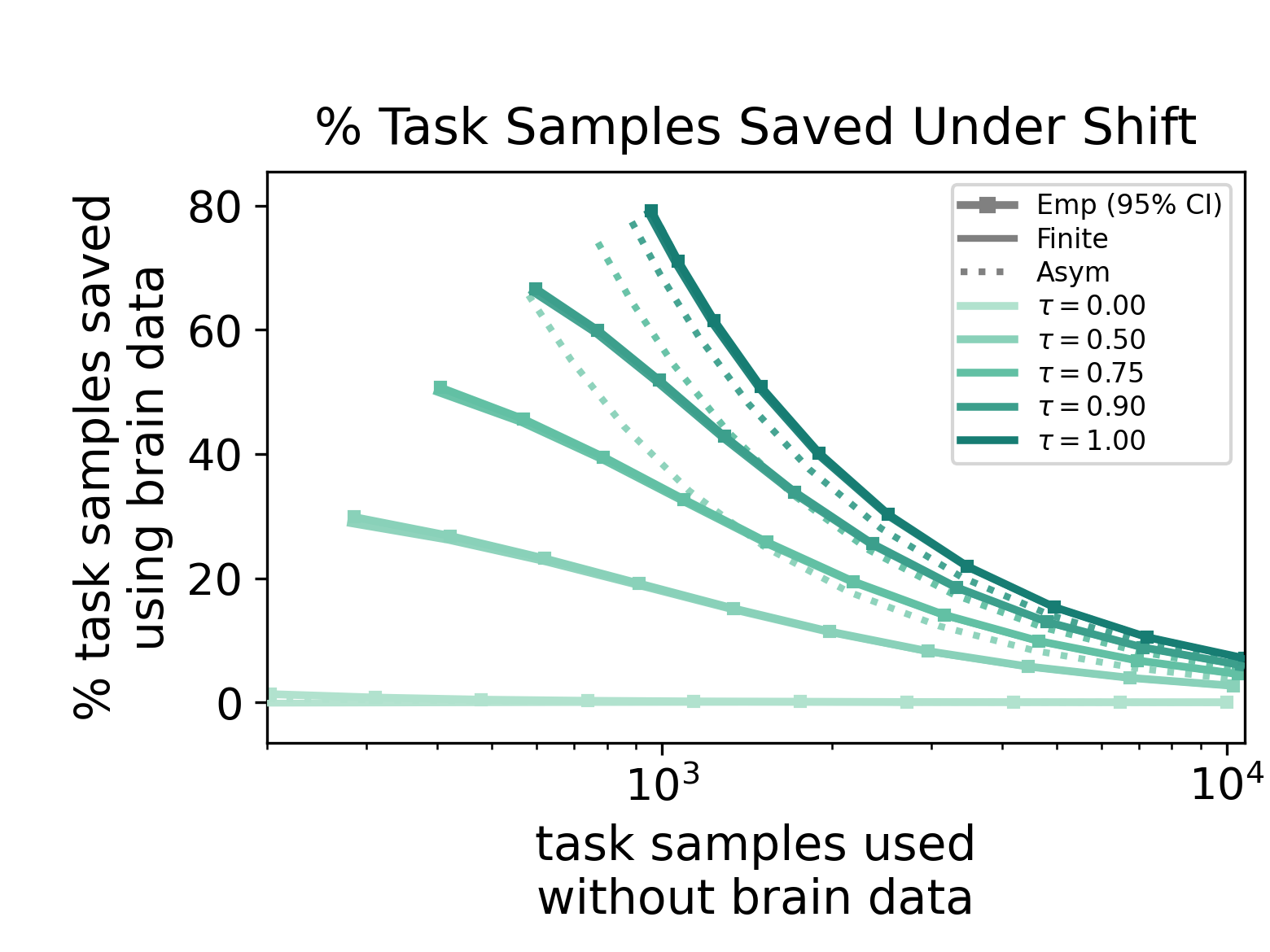}
    \includegraphics[width=0.49\linewidth]{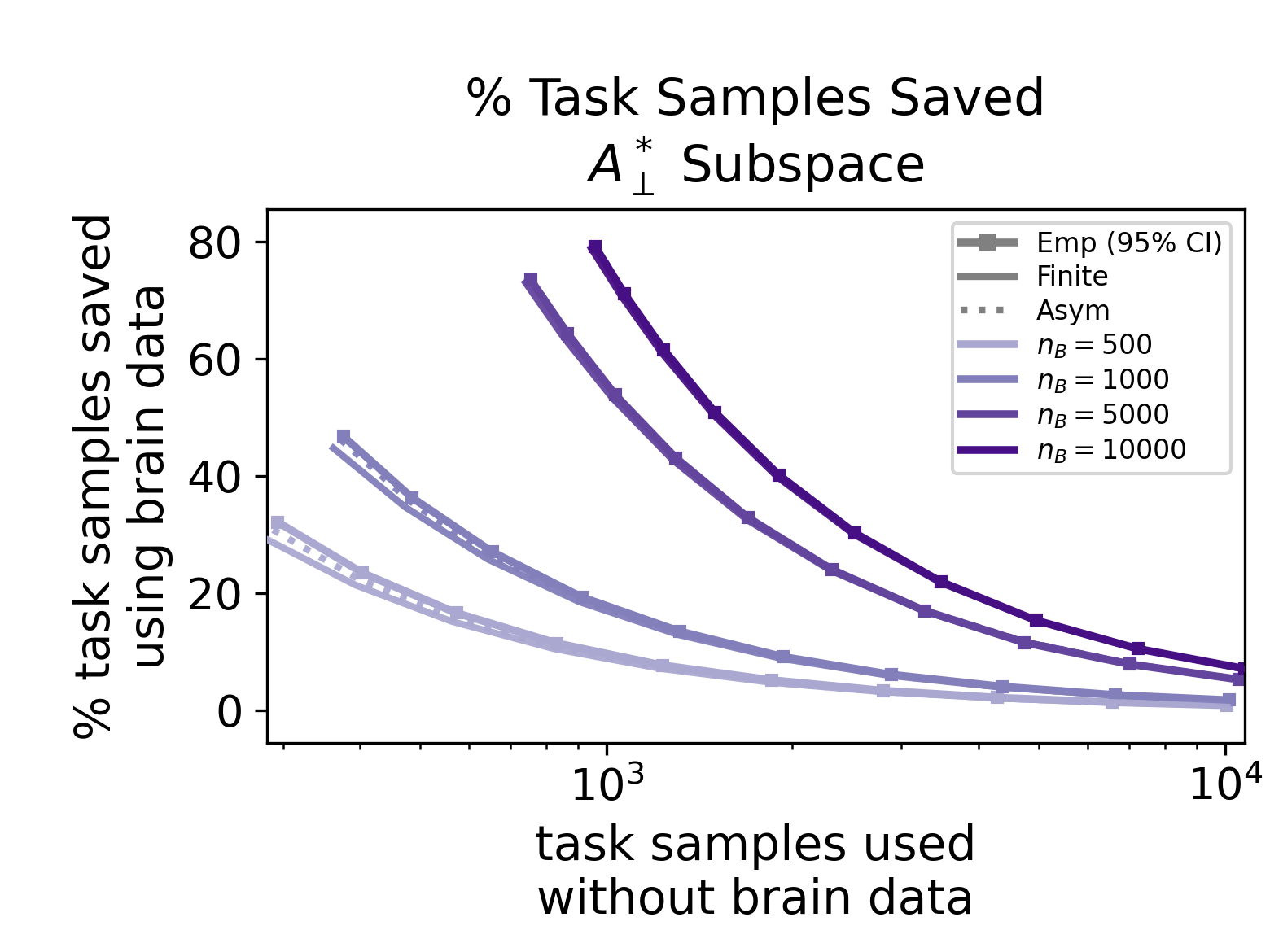}
    \caption{\textit{Left panel}: Empirical test error under test shift towards $P_{A^*_\perp}$ task data savings match scaling theory even at moderate task samples.\textit{Right panel}: Empirical test error for isotropic covariance closely matches the finite sample theory curves (\cref{thm:total_scaling}) under optimal regularization. 100k independent trials were used to generate each MSE estimate replicate and 30 replicates averaged to generate the mean and confidence interval. Parameters used ($m=0.05 $, $\text{SNR}_T/\text{SNR}_B=1.83$, $n_B =10000$, samples $d_{{\ell_{H^*}}}/d_x=62\%$, $100,000$ trials). Empirical curves (Emp) are plotted as solid with a square at evaluated points with confidence intervals, asymptotic curves (Asym) \cref{eq:exchange_rate} are plotted dashed, and finite sample theory curves (Finite) \cref{thm:total_scaling} are plotted in solid. }
    \label{fig:empirical_tau_on_off_subspace}
\end{figure}

\begin{figure}
    \centering
    \includegraphics[width=1.0\linewidth]{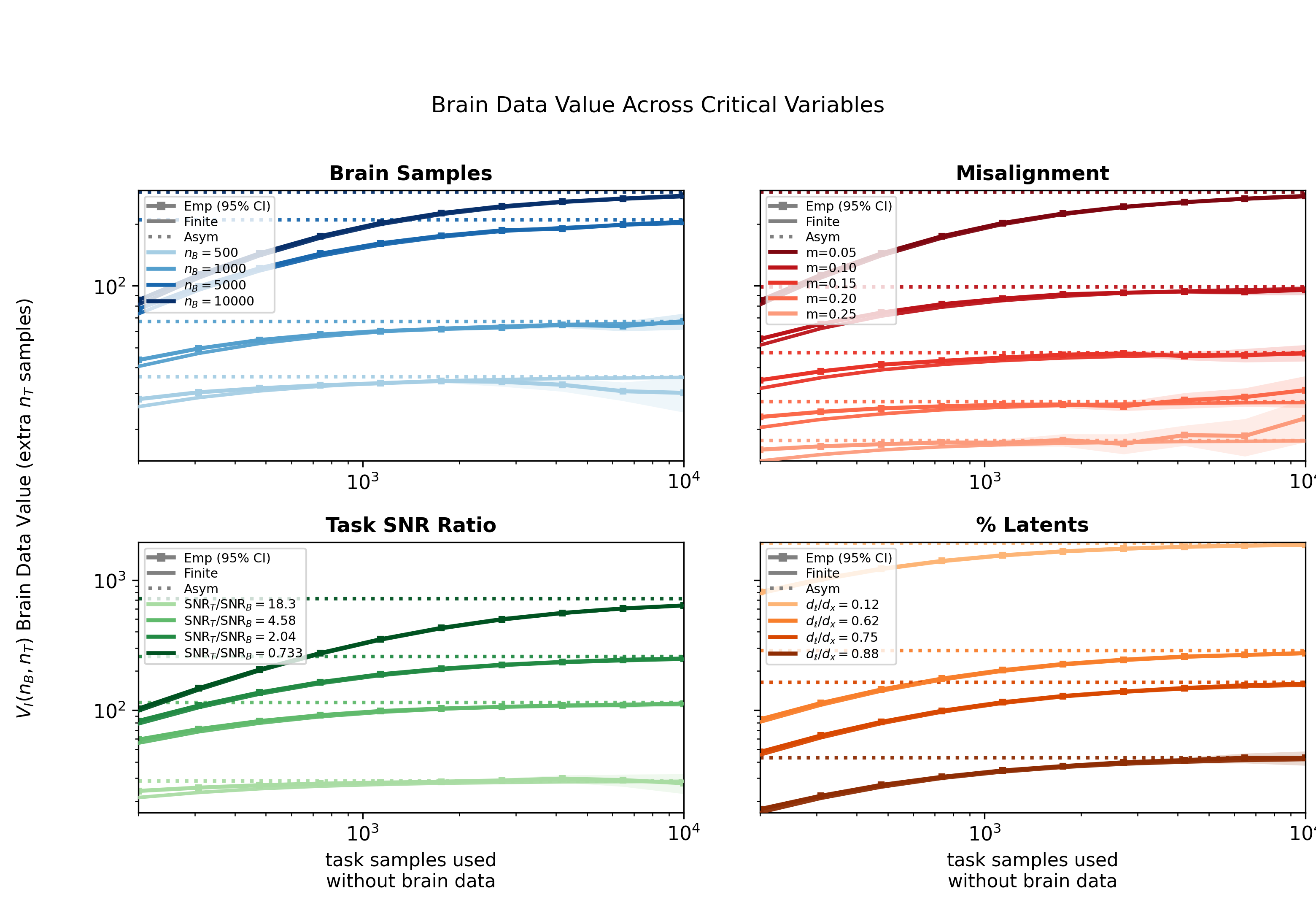}
    \caption{Estimated brain data value matches the finite sample theory curves (\cref{thm:total_scaling}) even in moderate task samples. 100k independent trials used to generate each MSE estimate replicate and 30 replicates averaged to generate the mean and confidence interval. Parameters used ($m=0.05 $, $SNR_T/SNR_B=1.83$, $n_B =10000$ samples $d_{{\ell_{H^*}}}/d_x=62\%$, $100,000$ trials). Empirical curves (Emp) are plotted as solid with a square at evaluated points with confidence intervals, asymptotic curves (Asym) \cref{eq:exchange_rate} are plotted dashed, and finite sample theory curves (Finite) \cref{thm:total_scaling} are plotted in solid. }
    \label{fig:empirical_brain_value}
\end{figure}
\begin{figure}
    \centering
    \includegraphics[width=1.0\linewidth]{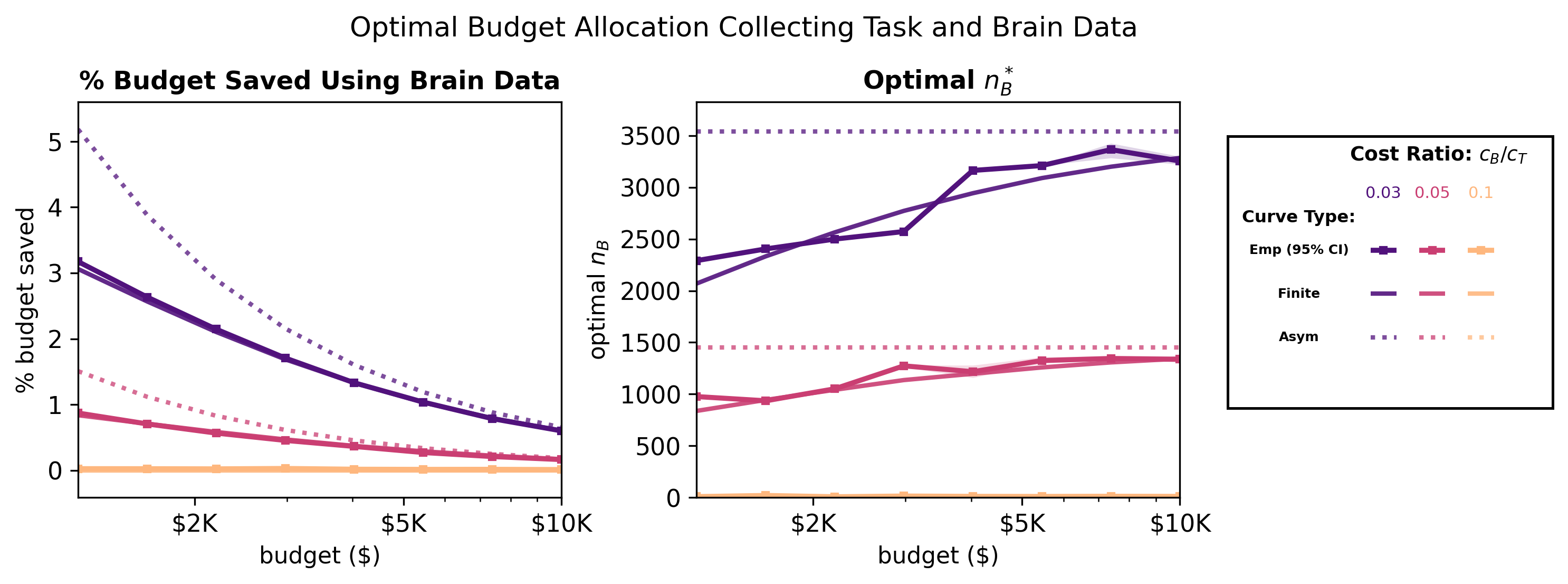}
    \caption{ Empirical curves closely follow the finite scaling law theory (\cref{thm:total_scaling}), coarseness stems from the number of brain and task samples explored in the cost minimization. Parameters used: $m=0.05 $, $SNR_T/SNR_B=1.83$, $n_B =10000$ samples $d_{{\ell_{H^*}}}/d_x=62\%$, 1.5 million estimator trials. Empirical curves (Emp) are plotted as solid with a square at evaluated points with confidence intervals, asymptotic curves (Asym) \cref{eq:exchange_rate} are plotted dashed, and finite sample theory curves (Finite) \cref{thm:total_scaling} are plotted in solid. }
    \label{fig:empirical_budget}
\end{figure}
\FloatBarrier

\section{Extra Theory Figures}\label{app:extra_theory_figs}
\subsection{More fMRI theory details} 
In order to obtain the theory plots described in the main paper, we used a random orthonormal projection for the first layer $A$, $\Sigma_{\ell_{H^*}} = 0.5I$, $\sigma^2_r = 0.4$, and a pooling measurement matrix $H^*$ such that each voxel receives the sum of 4 latents. 
\begin{figure}[H]
    \centering
    \includegraphics[width=0.9\linewidth]{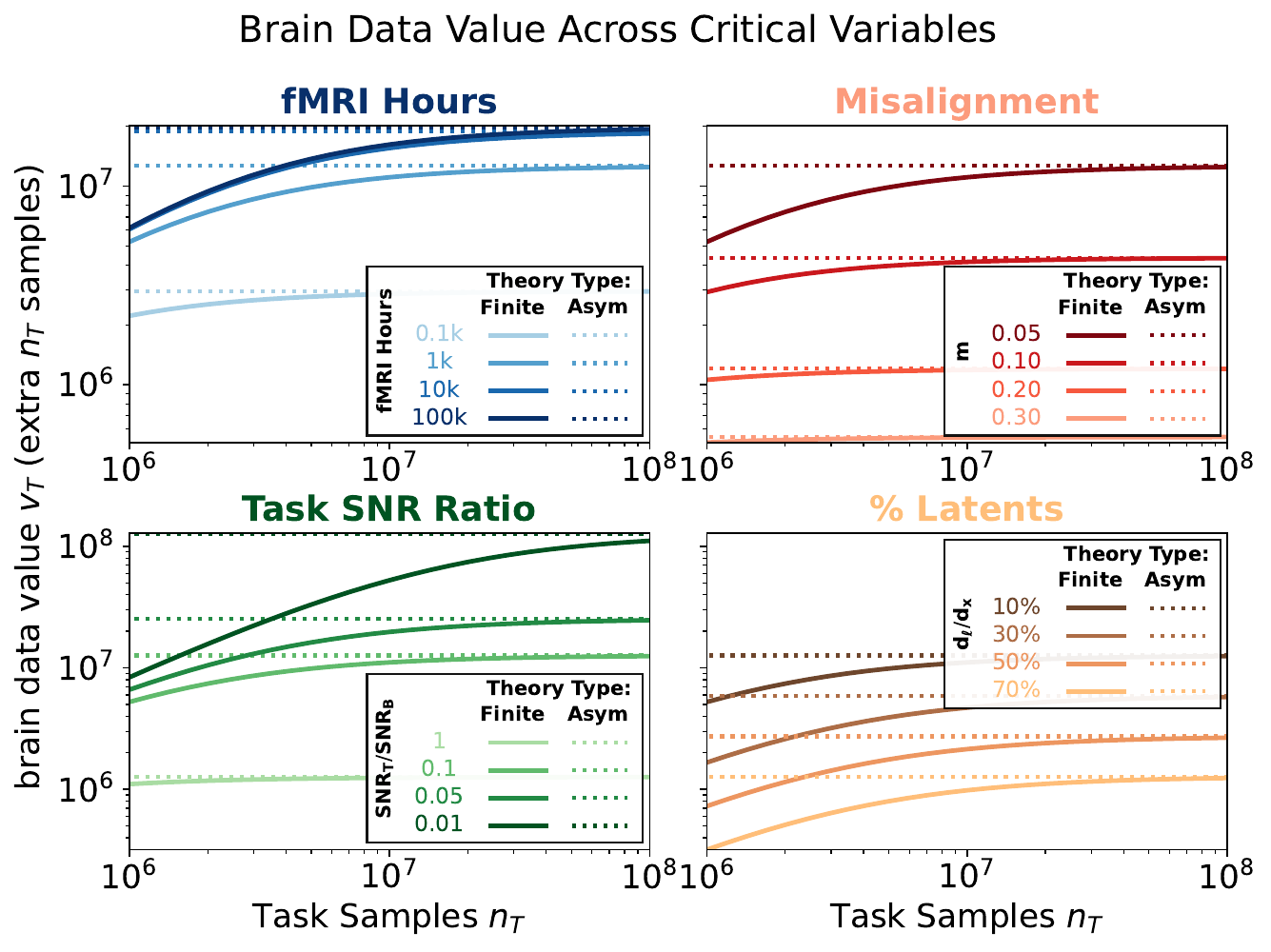}
    \caption{Brain data value approaches the exchange rate theory in large $n_T$. Top left: Adding more brain samples increases value, however the value asymptotes in large brain samples. Top right: Lower misalignment increases the value of brain data. Bottom left: Increasing the task difficulty compared to the difficulty of estimating the brain drives up the value of brain data. Bottom right: Smaller ratios of latents to ambient dimension under fixed misalignment. }
    \label{fig:extra_value_theory}
\end{figure}
\subsection{$\delta$ definition and interpretation}\label{app:delta}
$\delta$ is defined as:
\[
\Sigma_{est} = A^*(\sigma_r^2(H^{*T}H^*)^{-1} +\Sigma_{\ell_{H^*}})A^{*T}, \quad \delta = \left(
\beta^{*T}\Sigma_{est}\beta^*
-
\|\beta^*_{A_\perp^{*}}\|^2\frac{\mathrm{Tr}(\Sigma_{est})}{d_x-d_{\ell_{H^*}}}
\right)
\]
$\Sigma_{est}$ provides the noise, in the encoding feature space, of an optimal measurement map going backward from recordings to latents. So it captures the amount of estimator latent noise from an optimal recording to latent decoder. $\delta$ measures the noise in estimating task relevant features from a feature subspace learned in finite brain data. The norm of $\delta$ controls the constant on the rate of scaling with brain data, similar to a variance term in OLS scaling. 

In a simple case, suppose that the latent noise is given by $\Sigma_{{\ell_{H^*}}} = \sigma^2_{\ell_{H^*}} I$, and the number of recording dimensions is a multiple of ${\ell_{H^*}}$, $d_r = kd_{\ell_{H^*}}$ with $H^* = \omega_{H} [I^{(1)}_{d_{\ell_{H^*}},d_{\ell_{H^*}}},I^{(2)}_{d_{\ell_{H^*}},d_{\ell_{H^*}}},..I^{(k)}_{d_{\ell_{H^*}},d_{\ell_{H^*}}}]$ then $\Sigma_{est} = (\frac{\sigma_r^2}{\omega^2_{H}} +  \sigma_{{\ell_{H^*}}}^2) P_{A^*}$, then 
\[
\delta = d_{\ell_{H^*}}\left(\frac{\sigma_r^2}{k\omega^2_{H^*}} + \sigma^2_{\ell_{H^*}}\right)\left(\frac{\|\beta^*_{A^*}\|^2}{d_{\ell_{H^*}}} - \frac{\|\beta^*_{A_\perp^*}\|^2}{d_x - d_{\ell_{H^*}}}\right)
\]
So increasing the number of recorded dimensions is able to decrease the effective recording noise in estimating the latents, however more recordings do not help suppress latent noise. Note that this scales with $\|\beta^*\|$. If the norm of the task is large, then small misalignment means that task error is large. If the alignment is moderately high $\frac{\|\beta_{A^*}^{*}\|}{\|\beta^*_{A^*\perp}\|} > \sqrt{\frac{d_{\ell_{H^*}}}{d_x - d_{\ell_{H^*}}}}$, $\delta$ has a positive scaling sign controlled by the effective latent estimation noise. However, this term can be negative if the brain is highly misaligned. The intuition behind this is that if a brain is very misaligned, finite sample fluctuations are more aligned than the population quantity, so adding more brain samples actually would hurt performance. Note that $\delta$ only controls the rate of learning the task from from the brain, a poor misalignment will reduce the total amount of brain task data value. The special case when $\delta = 0$ corresponds to the case when the aligned portion of the task $\beta_{A^*}^*$ and the misaligned portion $\beta^*_{A^*\perp}$ have equal relative mass to their dimensions. This means that the population projection is behaving like a random projection of the task map and finite sample fluctuations are not detrimental at first order.
\section{Proofs}\label{app:proofs}
\subsection{Notation}\label{app:notation}
In order to have better precision in the proofs than the notation in the paper we adopt a more verbose notation in some areas:
\begin{itemize}
    \item $\varepsilon_{\Sigma_{test}}(n_B, n_T)$ denotes the mse with respect to the input covariance test distribution $\Sigma_{test}$. $\varepsilon^{TOS}_{\Sigma_{test}}(n_T)$ is used to denote the task only student and $\varepsilon^{BEFS}_{\Sigma_{test}}(n_B, n_T)$ is used to denote the brain encoding foundation student.
    \item $V_{\Sigma_{test}}(n_B, n_T)$ denotes the effective task sample of brain data with respect to the input covariance test distribution $\Sigma_{test}$. 
\end{itemize}
We use several special names throughout these proofs for useful quantities that appear many times
\begin{itemize}
\item The $j_{th}$ eigenvalue of a matrix is given by $\mu_{j}$.
\item The pseudoinverse is denoted by $\dagger$.
    \item $
\mathbb{E}_{X^{(B)}, R^{(B)}}[\|
\beta_{\hat A\perp}^{*}\|^2]\approx \gamma_{I}(n_B)$ see \cref{thm:total_scaling}
\item $\mathbb{E}_{X^{(B)}, R^{(B)}}[
\beta_{\hat A\perp}^{*T}\Sigma_{test}\beta^*_{\hat A\perp}]\approx \gamma_{\Sigma_{test}}(n_B)
 $  see \cref{thm:total_scaling}
 \item $K = (I - P_{A^*})P_{\hat A}P_{A^*}$
 \item $K_{lin}$ first order approximation of $K$
 \item $\alpha = \frac{1}{1+\lambda}$
 \item $J_{\hat A} = P_{\hat A} + \alpha P_{\hat A_\perp}$
 \item $M_i, M$ denote remainder terms, we consider these indices local to each lemma/thm for notational cleanliness. $Z_{main}$ we also use locally to denote the main scaling terms that are not event control remainders.
 \item $Q_{ols} = (X^TX)^{-1}X^TR$
 \item $E_r$ to denote the row stacked $\eta_{r,i}$, $E_{{\ell_{H^*}},i}$ to denote the row stacked $\eta_{\ell_{H^*}}$, $e_{y}$ to denote the row stacked $\eta_{y,i}$. 
 \item $\Delta_R = (X^{(B)T}X^{(B)})^{-1}X^{(B)T}(E_r + E_{\ell_{H^*}} H^*)$
 \item $\Delta_y = (X^{(T)T}X^{(T)})^{-1}X^{(T)T}e_y$
 \item $\mathcal{E}$ denotes a statistical event condition, $\mathcal{E}_c$ denotes the complement of that event and $\mathbf{1}_{\mathcal{E}}$ denotes the indicator function of that event.
\end{itemize}
\subsection{BEFS Scaling Law}\label{app:scaling_law}
\textbf{BEFS Scaling Law Proof Sketch}\\
We show that the optimal brain encoding model is given by the low rank regression solution (\cref{lem:BEFS_1st_stage_P_hat_form}):
\[
\hat A, \hat H = LRR_{rank = \hat l}(X,R)
\]
and derive the first order expansion for $K = (I - P_{A^*})P_{\hat A}P_{A^*}\approx K_{lin}$ for \[K_{lin} = (I - P_{A^*})\Delta_{R}H^{*T}(H^*H^{*T})^{-1}A^{*T}\] (\cref{lem:1st_stage_BEFS_K_expansion}) and,
\[
P_{\hat A} \approx K_{lin} + K_{lin}^T 
\]
(\cref{lem:BEFS_1st_stage_P_hat_A_expansion})
We demonstrate that  (\cref{lem:K_expectation}) $\mathbb{E}[K] = 0$ under our model assumptions and as a consequence of this and our block test covariance structure, all scaling quantities in $n_B$ can be shown to depend only on $K_{lin}$ up to second order in the noise (\cref{lem:BEFS_1st_stage_trace_second_order_expansion}, \cref{lem:BEFS_stage_1_beta_norm_expansion}, \cref{lem:befs_1st_stage_mixed_expansion}). This allows us to derive the first order scaling quantity for
\[
\mathbb{E}[\beta_{\hat A\perp}^T\Sigma_{test}\beta_{\hat A\perp}] \approx \gamma_{\Sigma_{test}}(n_B)
\]
As well as other key $n_B$ dependent scaling quantities (\cref{lem:BEFS_1st_stage_beta_norm_scaling}, \cref{lem:BEFS_1st_Stage_trace_scaling_law},\cref{lem:BEFS_stage_1_mixed_scaling}).
We show that the solution of BEFS in stage 2 under a fixed brain latent encoding model $\hat A$ has the generalized positive semi-definite constraint ridge regression solution:
\[
 \hat \beta^{BEFS} = \beta^* + \Delta_{y} - \lambda(\hat \Sigma + \lambda( I - P_{\hat A}))^{-1}( I - P_{\hat A})(\beta^* + \Delta_{y})
\]
Which we approximate to second order in the noise (\cref{lem:BEFS_2nd_stage_soft_beta}). This allows us to get a first order closed form for $\mathbb{E}[y_{test} - x_{test}^{T}\hat \beta^{BEFS} | \hat A]$ in terms of stage 1 quantities such as the alignment of the encoding map $\beta_{\hat A\perp}^T\Sigma_{test}\beta_{\hat A\perp}$ (\cref{thm:befs_stage_2_scaling}).\\

Combining the scaling quantities from stage 1 and 2 gives us the total scaling law. Since we are operating in a gaussian regime, we are able to show explicit remainder control on the scaling law (\cref{thm:total_scaling}).
\subsection{General}
\begin{lemma}\label{lem:basics}
We use the following basic facts for gaussian distributions
Under $\Delta = (X^TX)^{-1}X^TE$, where $E_i \sim N(0,S)$,
\[
\mathbb{E}_{E}[\Delta \Delta^T|X] = \text{Tr}(S)(X^{T}X)^{-1}
\]
Additionally, for fixed $G$,
\[
\mathbb{E}_{E}[\Delta G\Delta^T|X] = \text{Tr}\left(S G\right)(X^{T}X)^{-1}
\]
\[
\mathbb{E}_{E}[\Delta^T G\Delta|X] = \text{Tr}(G(X^{T}X)^{-1})S
\]
Finally for $\hat \Sigma = \frac{1}{n}x_ix_i^T$, $x_i \sim N(0,\Sigma)$,
\[
\mathbb{E}[\hat \Sigma G\hat \Sigma] = \frac{n+1}{n}\Sigma G\Sigma + \frac{1}{n}\text{Tr}(\Sigma G)\Sigma
\]
And 
\[
\mathbb{E}_X[(\hat \Sigma  - \Sigma) G(\hat\Sigma - \Sigma)] = \mathbb{E}_X[\hat \Sigma G\hat\Sigma] + \Sigma G \Sigma = \frac{1}{n}\left(\Sigma G \Sigma + \text{Tr}(\Sigma G)\Sigma\right)
\]
Finally when $x_i \sim N(0, I)$ under the event $\|\hat \Sigma - I\|\leq 1/2$
\[
\hat \Sigma^{1/2} = I + \frac{1}{2}(\hat \Sigma - I) + O(\|\hat \Sigma - I\|^2_{op})
\]
And 
\[
\hat \Sigma^{-1/2} = I - \frac{1}{2}(\hat \Sigma - I) + O(\|\hat \Sigma - I\|^2_{op})
\]
\end{lemma}
\begin{lemma}[Gaussian Bounds]\label{lem:gaussian_bounds}
$Z \sim N(0, \Sigma_z)$ and $g \sim N(0,I)$, $X_{i,j} \sim N(0,1)$,
\[
\mathbb{E}[\|Z\|^k] \leq \|\Sigma_z\|^{k/2}_{op}\mathbb{E}[\|g\|^k] 
\le
C\|\Sigma_z\|^{k/2}_{op}
\]
$\Delta = (X^TX)X^Te$ for $(X_i)^T \sim N(0,\Sigma)$ and $e_i \sim N(0,\sigma_y^2)$. So $\Delta|X \sim N(0, \sigma_y^2(X^TX)^{-1})$. Then
\[
\mathbb{E}_{X}[\mathbb{E}_{e_y}[\|\Delta\|^{2k}|X]] \leq C_1(d,k)\sigma^{k}\mathbb{E}_{X}[\|(X^TX)^{-1}\|^{k}_{op}] \leq C_2(d,k,\sigma)n^{-k}\mathbb{E}_{X}[\|\hat \Sigma^{-1}\|^{k}_{op}]
\]
\[
\|\hat \Sigma^{-1}\|_{op} \leq \text{Tr}(\hat \Sigma^{-1}) = n \text{Tr}((X^TX)^{-1})
\]
And from \cite{matsumoto2012general}
\[
\mathbb{E}[\text{Tr}((X^TX)^{-1})^2] = O(n^{-1})\quad 
\mathbb{E}[\text{Tr}((X^TX)^{-1})^2] = O(n^{-2})
\quad
\mathbb{E}[\text{Tr}((X^TX)^{-1})^4] = O(n^{-4})
\]
Then 
\[
\mathbb{E}[\|\hat\Sigma^{-1}\|_{op}],  \mathbb{E}[\|\hat\Sigma^{-1}\|^2_{op}],\mathbb{E}[\|\hat\Sigma^{-1}\|^4_{op}]= O(1)
\]
And 
\[
\mathbb{E}_{X,e_y}[\|\Delta\|^2] = O(n^{-1}) \quad \mathbb{E}_{X,e_y}[\|\Delta\|^4] = O(n^{-2})
\]
Finally, we use the standard gaussian concentration inequalities that 
\[
P\left(\|\hat \Sigma - I\| > L_1(q)\sqrt \frac{\log n }{n}\right) < t_1 n^{-q}
\]
and 
\[
P\left(\|\hat \beta^{OLS} - \beta^*\| > L_2(q)\sqrt \frac{\log n }{n}\right) < t_2 n^{-q}
\]
\end{lemma}
\begin{lemma}\label{lem:general_projection_expansion}
Suppose we have a symmetric matrix $\hat G \in R^{d\times d}$ with $\hat G = G + E$ with $G$ having $1..k$ non-zero eigenvalues $\mu_{i}$ and a multiplicity $d - k$ zero eigenvalue where $E$ is an arbitrary error matrix. $G = BB^T$. Call the top k eigenspace of $\hat G$, $\hat U_{k}$ and the top k eigenspace of $G$, $ U_{k}$. We show that under the event
$2\|E\|_{op}/\mu_{k}(BB^T)  < 1$, we obtain the first order expansion for the projection onto the eigenvectors of $\hat G$:
\[
P_{\hat U_{k}} = P_{B} +(BB^T)^\dagger E(I - P_{B}) + (I - P_{B})E(BB^T)^\dagger + M
\]
\[
\|M\|_{op} \leq k \frac{(2\|E\|_{op}/\mu_{k}(BB^T))^2}{1-(2\|E\|_{op}/\mu_{k}(BB^T))}
\]
Proof:\\
From \cite{jirak2020perturbation} equation 1.3 with positive eigenvalue set $\mathcal{I} = \{1..k\}$ and zero padding eigenvalues $\mathcal{I}^c = \{k+1..d\}$.
\[
P_{\mathcal{I}} = \sum^k_{i=1}P_i = P_{U_k} = P_{B}
\]
\[
P_{\mathcal{I}^c} = \sum^{d_x}_{j=k+1}P_j = I - P_{U_k} =  I - P_{B}
\]
Call $g_\mathcal{I} = \min_{i\in \mathcal{I}, j \in \mathcal{I}^c} |\mu_{i} - \mu_{j}| = \mu_{k}$.
Under the event  $\delta_{\mathcal{I}} = 2\|E\|_{op}/g_{\mathcal I} = 2\|E\|_{op}/\mu_{k}(BB^T)  < 1$, then for \[
\|S_{\mathcal{I}}(E)\|_{op} \leq |\mathcal{I}|\frac{\delta^2_{\mathcal{I}}}{1-\delta_{\mathcal{I}}} = k\frac{\delta^2_{\mathcal{I}}}{1-\delta_{\mathcal{I}}}
\]
\[
P_{\hat U_{k}} - P_{U_k} = \sum^k_{i=1}\sum^{d}_{j=k+1}\frac{1}{\mu_{i} - \mu_{j}}(P_{i}EP_{j} + P_{j}EP_{i}) + S_{\mathcal{I}}(E)
\]
Since the $\mathcal{I}^c$ eigenvalue is zero,
\[
=\sum^k_{i=1}\sum^{d}_{j=k+1}\frac{1}{\mu_{i}}(P_{i}EP_{j} + P_{j}EP_{i}) + S_{\mathcal{I}}(E)
\]

\[
= \left(\sum^k_{i=1}\frac{1}{\mu_i}P_i\right)E(I - P_{B}) + (I - P_B)E\left(\sum^k_{i=1}\frac{1}{\mu_i}P_i\right)+  S_{\mathcal{I}}(E)
\]
Note that $(BB^T)^\dagger = S^\dagger = \sum^k_{i=1}\frac{1}{\mu_i}P_i$ is a pseudoinverse,
\[
= (BB^T)^\dagger E(I - P_{B}) + (I - P_{B})E(BB^T)^\dagger + S_{\mathcal{I}}(E)
\]
\end{lemma}
\begin{lemma}\label{lem:general_K_expansion}
We derive the following expressions for the projection matrix $P_{\hat A}$ under the event that $\|P_{A^*} - P_{\hat A}\|_{op} \leq 1/2$ for $K = (I - P_{A^*})P_{\hat A}P_{A^*}$ and $\|M\|_{op} \leq 4\|K\|^4_{op}$:\\
1)
\[
P_{A^*} - P_{A^*}P_{\hat A}P_{A^*} = K^TK + M
\]\\
2)
\[
(I -P_{A^*})P_{\hat A} (I -P_{A^*}) = KK^T + M
\]
Proof of 1)\\
\[
K^TK = P_{A^*}P_{\hat A}(I - P_{A^*})P_{\hat A}P_{A^*} = P_{A^*}P_{\hat A}P_{A^*} - (P_{A^*}P_{\hat A}P_{A^*})^2
\]
\[
K^TK + (P_{A^*} - P_{A^*}P_{\hat A}P_{A^*})^2 = P_{A^*}P_{\hat A}P_{A^*} - (P_{A^*}P_{\hat A}P_{A^*})^2 + P_{A^*} - 2(P_{A^*}P_{\hat A}P_{A^*}) + (P_{A^*}P_{\hat A}P_{A^*})^2 = P_{A^*} - P_{A^*}P_{\hat A}P_{A^*}
\]
So, $P_{A^*} - P_{A^*}P_{\hat A}P_{A^*} = K^TK + (P_{A^*} - P_{A^*}P_{\hat A}P_{A^*})^2$, and under the condition that $\|P_{\hat A} - P_{A^*}\|_{op}\leq 1/2$, 
\[
\|P_{A^*} - P_{A^*}P_{\hat A}P_{A^*}\|_{op} \leq \|(P_{A^*} - P_{A^*}P_{\hat A}P_{A^*})\|^2_{op} + \|K^TK\|_{op}
\]
and
\[
\|P_{A^*} -  P_{A^*}P_{\hat A}P_{A^*}\|_{op} = \|P_{A^*}(I - P_{\hat A})P_{A^*}\|_{op}\leq\|P_{A^*} - P_{\hat A}\|_{op} \leq 1/2
\]
Therefore
\[
\|P_{A^*} -  P_{A^*}P_{\hat A}P_{A^*}\|_{op} \leq \frac{1}{2}\|P_{A^*} -  P_{A^*}P_{\hat A}P_{A^*}\|_{op} + \|K^TK\|_{op}
\]
\[
\|P_{A^*} -  P_{A^*}P_{\hat A}P_{A^*}\|_{op}\leq 2\|K^TK\|_{op}\leq 2\|K\|^2_{op}
\]
and 
\[
\|(P_{A^*} - P_{A^*}P_{\hat A}P_{A^*})^2\|_{op}\leq 4\|K\|^4_{op}
\]
So, 
\[
P_{A^*} -  P_{A^*}P_{\hat A}P_{A^*} = K^TK + B
\]
for $\|B\|_{op} \leq 4\|K\|^4_{op}$\\

Proof of 2) \\
\[
KK^T = (I - P_{A^*})P_{\hat A}P_{A^*}P_{\hat A}(I - P_{A^*}) = (I - P_{A^*})P_{\hat A}(I - (I - P_{A^*}) )P_{\hat A}(I - P_{A^*})
\]
\[
 = (I - P_{A^*})P_{\hat A}(I - P_{A^*}) - \left((I - P_{A^*})P_{\hat A}(I - P_{A^*})\right)^2
\]
So, $(I -P_{A^*})P_{\hat A} (I -P_{A^*}) = KK^T + \left( (I -P_{A^*})P_{\hat A} (I -P_{A^*})\right)^2 $
And
\[
\|(I - P_{A^*})P_{\hat A}(I - P_{A^*})\|_{op} \leq \|K^TK\|_{op} + \|(I - P_{A^*})P_{\hat A}(I - P_{A^*})\|^2_{op}
\]
Under the event $\|P_{\hat A} - P_{A^*}\|_{op} \leq 1/2$, 
\[
\|(I -P_{A^*})P_{\hat A} (I -P_{A^*})\|_{op} = \|(I -P_{A^*})(P_{\hat A} - P_{A^*}) (I -P_{A^*})\|_{op}\leq \|P_{A^*} - P_{\hat A}\|_{op}\leq 1/2
\] 
So, 
\[
\|(I -P_{A^*})P_{\hat A} (I -P_{A^*})\|_{op} \leq \frac{1}{2}\|(I -P_{A^*})P_{\hat A} (I -P_{A^*})\|_{op} + \|KK^T\|_{op}
\]
\[
\|(I -P_{A^*})P_{\hat A} (I -P_{A^*})\|_{op} \leq 2\|KK^T\|=2\|K\|^2_{op}
\]
and 
\[
\|((I -P_{A^*})P_{\hat A} (I -P_{A^*}))^2\|_{op}\leq 4\|K\|^4_{op}
\]
Which gives the final result
\[
(I -P_{A^*})P_{\hat A} (I -P_{A^*}) = KK^T + B
\]
for $\|B\|_{op} \leq 4 \|K\|_{op}^4$
\end{lemma}
\begin{lemma}\label{lem:K_expectation}
$\mathbb{E}[K] = 0$
Take an orthogonal matrix $T$ such that $X' = XT^T$ and $R$ fixed. Then:
\[
\hat \Sigma' = T\hat \Sigma T^T
\]
\[
\hat \Sigma'^{1/2} = T\hat \Sigma^{1/2} T^T \quad \hat \Sigma'^{-1/2} = T\hat \Sigma^{-1/2} T^T 
\]
Building the estimator for $\hat A$, for $Q_{ols} = (X^TX)^{-1}X^TR$
\[
Q_{ols}' = (TX^TXT^T)^{-1}TX^TU^TR = T(X^TX)X^TR = TQ_{ols}
\]
\[
(\hat\Sigma Q_{ols})' = T\hat\Sigma Q_{ols}
\]
Then the top k left subspace is: $\tilde U_{k}' = T\tilde U_{k}$. 
Since $\hat A$ is the left singular space of $\hat S = \hat \Sigma^{-1/2}P_{\tilde U_k}\hat \Sigma^{-1/2}$
\[
\hat S'
=
T\hat\Sigma^{-1/2}U^T \, T P_{\tilde U_k}U^T \, T\hat\Sigma^{-1/2}U^T
=
T\hat\Sigma^{-1/2}P_{\tilde U_k}\hat\Sigma^{-1/2}T^T
\]
Then $\hat A' = T\hat A$ and $P_{\hat A}' = TP_{\hat A}T^T$ and $K'=(I-P_{A^*})T P_{\hat A}T^T P_{A^*}$
Suppose now we add the extra condition $TA^* = A^*$, then $TP_{A^*} =P_{A^*} =  P_{A^*}T^T$ 
Since,
\[
T^T(TA^*) = T^TA^* = A^*
\]
\[
PT = A^*A^{*T}T = A^*(T^TA^*)^T = A^*A^{*T}
\]
And this implies our estimator has the following relationship with $T$,
\[
K' = (I-P_{A^*})T P_{\hat A}T^T P_{A^*} = T(I-P_{A^*}) P_{\hat A}P_{A^*}
\]
Finally we show that the data distribution is the same under the transformation $T$,
Since $X_i \sim N(0, I)$ is isotropic gaussian, $X_i' = X_iT^T \sim N(0, T^TT) = N(0, I)$ $X$ has the same generating distribution. $R = XA^*H^* + \eta_R$ so $R' = XT^TA^*H^* + \eta_R = XA^*H^* + \eta_R$. Therefore the distributions of $(X',R')$ and $(X,R)$ are equal. \\
Now choose $T = 2P_{A^*} - I$, then $TA^* = A^*$ and $T^TT = (2P_{A^*} - I)(2P_{A^*} - I) = 4P_{A^*}^2 - 4P_{A^*} + I = I$ so the conditions of $T$ are satisfied. Additionally, $K' = T(I - P_{A^*})P_{\hat A}P_{A^*} = -(I - P_{A^*})P_{\hat A}P_{A^*}$.

Since $(X',R')$ has the same distribution as $(X,R)$
\[
\mathbb{E}_{X,R}[K] = \mathbb{E}_{X',R'}[K]
\]
And since $K(X',R') = K(X',R) = TK(X,R)$,
\[
\mathbb{E}_{X',R'}[K] = \mathbb{E}_{X,R}[TK] = \mathbb{E}_{X,R}[-K] = -\mathbb{E}_{X,R}[K]
\]
So from this we obtain
\[
\mathbb{E}_{X,R}[K] = -\mathbb{E}_{X,R}[K]
\]
So 
\[
\mathbb{E}_{X,R}[K]  = 0
\]
\end{lemma}
\subsection{BEFS - 1st Stage}
Assume the following conditions hold:
\[
\mu_1(H^*H^{*T})\leq C_{H^*,1}<\infty \quad \mu_k(H^*H^{*T})\geq C_{H^*,k}>0
\]
\[
\|\beta^*\|\leq C_{\beta^*}<\infty
\]
Define: 
\[L_{max} = \max{\{L_1(2), L_2(2)\}}\]
\[C_E = 8\sqrt{C_1}+4+3C_1\]
\[
C_S
=
17+\frac{18}{\sqrt{C_k}}+9C_{\tilde U_k}
\]
\[
C_{\tilde U_k}=L_{\max}^2\left(\frac{4C_{H^*,1}+12\sqrt{C_{H^*,1}}+8}{C_{H^*,k}}+\frac{8kC_E^2}{C_{H^*,k}^2}\right)
\]
\[
C_{\hat A}=18C_{\tilde U_k}+\left(36+\frac{32}{\sqrt{C_{H^*,k}}}\right)L_{\max}^2+8kC_S^2L_{\max}^2
\]
\[
C_{BEFS,1}=\frac{C_{H^*,k}}{4C_E}\]
\[
C_{BEFS,2}=\frac{1}{4C_S}\]
\[
C_{BEFS,3}=\frac{L_{\max}}{2\left(\frac{2L_2(2)}{\sqrt{C_{H^*,k}}}+C_{\hat A}\right)}
\]
And assume $n_B$ large enough such that:
\[
L_{\max}\sqrt{\frac{\log n_B}{n_B}}<\min\left\{1,\frac12,C_{BEFS,1},C_{BEFS,2},C_{BEFS,3}\right\}
\]
\[
n_B > d_x -1
\]
Here we define an event $\mathcal{E}^{BEFS-1}$, and assume throughout this section that the event holds. $\mathcal{E}^{BEFS-1}$ is defined as the following:
\[
\|\hat\Sigma - I\|_{op} \leq L_1(2)\sqrt{\frac{\log n_B}{n_B}}
\]
\[
\|\Delta_R\|_{op} \leq L_2(2)\sqrt{\frac{\log n_B}{n_B}}
\]
Using gaussian concentration from \cref{lem:gaussian_bounds},
\[
P\left(\|\hat\Sigma - I\|_{op} \leq L_1(2)\sqrt{\frac{\log n_B}{n_B}}\right) \geq t_1(2)n^{-2}_B
\]
\[
P\left(\|\Delta_R\|_{op} \leq L_2(2)\sqrt{\frac{\log n_B}{n_B}} \right)\geq t_2(2)n^{-2}_B
\]
So by a union bound, for some constant $C$,
\[
P(\mathcal{E}^{BEFS-1}) \geq 1 - Cn^{-2}
\]
\begin{lemma}[$P_{\hat A}$ Exact Form]\label{lem:BEFS_1st_stage_P_hat_form}
We show that 
\[
\hat A, \hat H = \text{argmin}_{A, H} \mathcal{L}(A, H) = \text{argmin}_{A, H} \frac{1}{n_B}\|R^{(B)} - X^{(B)}AH\|_{F}^2
\]
Has an exact solution. For  $AH = Q$ and $Q_{ols}= (X^{(B)T}X^{(B)})^{-1}X^{(B)T}R^{(B)}$ for the top k SVD truncation $\Pi_{r_k}$,
\[
Q_{min} = \hat \Sigma^{-1/2}\Pi_{r_k}(\hat \Sigma^{1/2}Q_{ols})
\]
Which produces the un-normalized $\hat A = \hat\Sigma^{-1/2} \tilde{U}_{k}$ and $\hat H = \tilde{D}\tilde{V}_{k}$ for $\Pi_{r_k}(\hat \Sigma ^{1/2}Q_{ols}) = \tilde{U}_k\tilde{D}\tilde{V}_k^T$. So $P_{\hat A} = U_kU_k^T$ for $U_k$ as the top k eigenspace of $\hat \Sigma^{-1/2}P_{\tilde U_k}\hat \Sigma^{-1/2}$.\\

Proof:\\
Rewriting the objective as a low rank problem:
\[
\text{min}_{A, H} \mathcal{L}(A, H) = \text{min}_{Q, rank(Q) = k}\frac{1}{n_B}\|R^{(B)} - X^{(B)}Q\|_F^2 = \text{min}_{Q, rank(Q) = k}\frac{1}{n}\|R^{(B)} - X^{(B)}Q_{ols} + X^{(B)}Q_{ols} - X^{(B)}Q\|_F^2 
\]
Due to orthogonality of OLS residuals, 
\[
= \frac{1}{n}\|R^{(B)} - XQ_{ols}\|_F^2 +\text{min}_{Q, rank(Q) = k}\frac{1}{n} \|\hat{\Sigma}^{1/2}(Q_{ols} - Q)\|_F^2
\]
Call $\tilde{Q}_{ols} = \hat{\Sigma}^{1/2}Q_{ols}$ and $\tilde{Q} = \hat{\Sigma}^{1/2}Q$. Then since $\hat{\Sigma}^{1/2}$ is full rank, $\tilde{Q}$ is also rank k.
\[
 \frac{1}{n}\|R^{(B)} - XQ_{ols}\|_F^2 +\text{min}_{\tilde{Q}, rank(\tilde{Q}) = k}\frac{1}{n} \|\tilde{Q}_{ols} - \tilde{Q}\|_F^2
\]
Which has the known SVD solution:
\[
\tilde{Q}_{min} = \Pi_{r_k}(\tilde{Q}_{ols})
\]
So, 
\[
Q_{min} = \hat \Sigma^{-1/2}\Pi_{r_k}(\hat \Sigma^{1/2}Q_{ols})
\]
Which describes $\hat A = \hat\Sigma^{-1/2} \tilde{U}_{k}$ and $\hat H = \tilde{D}\tilde{V}_{k}$ for $\Pi_{r_k}(\hat \Sigma ^{1/2}Q_{ols}) = \tilde{U}_k\tilde{D}\tilde{V}_k^T$.
\end{lemma}

\begin{lemma}[$P_{\tilde{U}_k}$ Expansion]\label{lem:BEFS_1st_stage_P_U_expansion}
From \cref{lem:BEFS_1st_stage_P_hat_form}, we determined the need for a first order expansion of $P_{\tilde U_k}$ where $\tilde U_k$ is the top k left singular vectors of $\hat \Sigma^{1/2}Q_{ols}$. So $\tilde{U}_{k}$ is defined by the top k eigenvectors of $\hat \Sigma^{1/2}Q_{ols}Q_{ols}^T\hat \Sigma^{1/2}$. From the definition of OLS,  $Q_{ols} = A^*H^* + (X^TX)^{-1}X^T\xi_{R} = A^*H^* + \Delta_{R}$.

Then we derive the first order expansion as:
\[
P_{\tilde U_k} = P_{A^*} + \zeta + M
\]
For 
\[
\zeta = A^*(H^*H^{*T})^{-1}H^*\Delta_{R}^T(I - P_{A^*}) + (I - P_{A^*}) \Delta_{R}H^{*T}(H^*H^{*T})^{-1}A^{*T}\]
\[
 +\frac{1}{2}P_{A^*}(\hat\Sigma - I)(I - P_{A^*})+\frac{1}{2}(I - P_{A^*})(\hat\Sigma - I)P_{A^*}
\]
and 
\[
\|M\|_{op}  = O\left(\frac{\log n_B }{n_B}\right)
\]
Proof:\\
Call $B =(\hat \Sigma^{1/2} - I)$. Note that from \cref{lem:basics}, under $\mathcal{E}^{BEFS - 1}$,
\[
B = \frac{1}{2}(\hat \Sigma - I) + M_3,\qquad \|M_3\|_{op}\leq \frac{1}{2}\|\hat\Sigma-I\|_{op}^2\leq \frac{1}{2}L_1(2)^2\frac{\log n_B}{n_B}
\]
\[
\hat\Sigma^{1/2}Q_{ols}Q_{ols}^T\hat \Sigma^{1/2} = (I + B)(A^*H^* + \Delta_R)(A^*H^* + \Delta_R)^T(I + B)
\]
Then for $\|E\|_{op} = O\left(\sqrt{\frac{\log n_B}{n_B}}\right)$,
\[
= A^*H^*H^{*T}A^{*T} + E
\]
\[
\|E\|_{op} \leq (1 + 2\|B\|_{op} + \|B\|^2_{op})(2\|H^*\|_{op}\|\Delta_R\|_{op} + \|\Delta_{R}\|^2_{op}) + (2 \|B\|_{op} + \|B\|^2_{op})\|H^*\|_{op}^2
\]
and since under $\mathcal{E}^{BEFS -1}$ $\|\hat \Sigma^{1/2} - I\| \leq \|\hat \Sigma - I\|$,
\[
\|E\|_{op} \leq \left(1 + 2L_1(2)\sqrt{\frac{\log n_B}{n_B}} + L_1(2)^2\frac{\log n_B}{n_B}\right)\left(2\sqrt{C_{H^*,1}}L_2(2)\sqrt{\frac{\log n_B}{n_B}} + L_2(2)^2\frac{\log n_B}{n_B}\right) 
\]
\[
+ \left(2 L_1(2)\sqrt{\frac{\log n_B}{n_B}} + L_1(2)^2\frac{\log n_B}{n_B}\right)C_{H^*,1}
\]
Since $L_{\max}\sqrt{\frac{\log n_B}{n_B}} < 1$,
\[
1+2L_1(2)\sqrt{\frac{\log n_B}{n_B}}+L_1(2)^2\frac{\log n_B}{n_B} \le 4,\qquad
2L_1(2)\sqrt{\frac{\log n_B}{n_B}}+L_1(2)^2\frac{\log n_B}{n_B} \le 3L_{\max}\sqrt{\frac{\log n_B}{n_B}}
\]
\[
2\sqrt{C_{H^*,1}}L_2(2)\sqrt{\frac{\log n_B}{n_B}} + L_2(2)^2\frac{\log n_B}{n_B} \le (2\sqrt{C_{H^*,1}}+1)L_{\max}\sqrt{\frac{\log n_B}{n_B}}
\]
\[
\|E\|_{op}
\le
(8\sqrt{C_{H^*,1}}+4+3C_{H^*,1})L_{\max}\sqrt{\frac{\log n_B}{n_B}} = C_E L_{\max}\sqrt{\frac{\log n_B}{n_B}} < \frac{1}{4}C_{H^*,k} \leq \frac{1}{2}\mu_k(H^*H^{*T})
\]
Applying \cref{lem:general_projection_expansion}, since under $\mathcal{E}^{BEFS-1}$ we have that $\|E\|_{op} < \mu_{k}(A^*H^*H^{*T}A^{*T})/2 = \mu_{k}(H^*H^{*T})/2$ since $A$ is orthonormal
\[
P_{\tilde{U}_{k}} = P_{A^*H^*} + (A^*H^*H^{*T}A^{*T})^\dagger E(I - P_{A^*H^*}) + (I - P_{A^*H^*})E(A^*H^*H^{*T}A^{*T})^\dagger + M_1
\]
\[
\|M_1\|_{op} \leq k \frac{(2\|E\|_{op}/\mu_{k}(H^*H^{*T}))^2}{1-(2\|E\|_{op}/\mu_{k}(H^*H^{*T}))}
\]
Simplifying $P_{A^* H^*} = P_{A^*}$ and $(A^*H^*H^{*T}A^{*T})^\dagger = A^*(H^*H^{*T})^{-1}A^{*T}$,
\[
P_{\tilde{U}_{k}} = P_{A^*} +  A^*(H^*H^{*T})^{-1}A^{*T} E(I - P_{A^*}) + (I - P_{A^*})E A^*(H^*H^{*T})^{-1}A^{*T} + M_1
\]
From our earlier derivation,
\[
2\frac{\|E\|_{op}}{\mu_{k}(H^* H^{*T})} \leq \frac{2C_E}{C_{H^*,k}}L_{\max}\sqrt{\frac{\log n_B}{n_B}} <\frac{1}{2}
\]
So 
\[
1 - 2\frac{\|E\|_{op}}{\mu_{k}(H^* H^{*T})} \geq \frac{1}{2}
\]
Which means,
\[
\|M_1\|_{op}  \leq 2k\left(\frac{2\|E\|_{op}}{\mu_{k}(H^* H^{*T})}\right)^2 \leq \frac{8kC_E^2L_{\max}^2}{C_{H^*,k}^2}\frac{\log n_B}{n_B}
\]
Pulling our higher order terms from $E$
\[
E = E_1 + M_2
\]
\[
E_{1} = \Delta_RH^{*T}A^{*T} + A^*H^*\Delta_R^T + BA^*H^*H^{*T}A^{*T} + A^*H^*H^{*T}A^{*T}B
\]
\[
M_2 = \Delta_{R}\Delta_{R}^T + B(A^*H^*\Delta_{R}^T + \Delta_{R}H^{*T}A^{*T}) + (A^*H^*\Delta_{R}^T+\Delta_{R}H^{*T}A^{*T})B + B(\Delta_R\Delta_{R}^T) + (\Delta_R\Delta_{R}^T)B
\]
\[
+ B(A^*H^*H^{*T}A^{*T})B + B(A^*H^*\Delta_{R}^T + \Delta_{R}H^{*T}A^{*T} + \Delta_{R}\Delta_{R}^T)B
\]
Using $\|A^*H^*\|_{op} \leq \sqrt {C_{H^*,1}}$ and cauchy schwartz/triangle inequalities and $L_{\max}\sqrt{\frac{\log n_B}{n_B}} < 1$,
\[
\|M_2\|_{op}
\le
(C_{H^*,1} + 6\sqrt{C_{H^*,1}} + 4)L_{\max}^2 \frac{\log n_B}{n_B}
\]
Finally, using the expansion under the $\mathcal{E}^{BEFS-1}$ that  $B = \frac{1}{2}(\hat \Sigma - I) + M_3$, $\|M_3\|_{op} \leq \frac{1}{2}\|\hat \Sigma - I\|^2_{op} \leq \frac{1}{2}L_1(2)^2\frac{\log n_B}{n_B}$,
\[
E_1 = E_{lin} + M_4
\]
\[
E_{lin} = \Delta_{R}H^{*T}A^{*T} + A^*H^*\Delta_{R}^T + \frac{1}{2}(\hat \Sigma - I)A^*H^*H^{*T}A^{*T} + \frac{1}{2}A^*H^*H^{*T}A^{*T}(\hat \Sigma - I)
\]
\[
M_4 = M_3A^*H^*H^{*T}A^{*T} + A^*H^*H^{*T}A^{*T}M_3
\]
So $\|M_4\|_{op} \leq C_{H^*,1}L_1(2)^2\frac{\log n_B}{n_B}$ and $\|M_5\|_{op} = \|M_4 + M_2\|_{op} \leq (2C_{H^*,1} + 6\sqrt{C_{H^*,1}} + 4)L_{\max}^2 \frac{\log n_B}{n_B}$ so,
\[
E = E_{lin} + M_5
\]
Moving $E$ into $P_{\tilde U_{k}}$,
\[
P_{\tilde{U}_{k}} = P_{A^*} +  A^*(H^*H^{*T})^{-1}A^{*T} (E_{lin} +M_5)(I - P_{A^*}) + (I - P_{A^*}) (E_{lin} +M_5) A^*(H^*H^{*T})^{-1}A^{*T} + M_1
\]
\[
P_{\tilde{U}_{k}} = P_{A^*} +  A^*(H^*H^{*T})^{-1}A^{*T}E_{lin}(I - P_{A^*}) + (I - P_{A^*}) E_{lin} A^*(H^*H^{*T})^{-1}A^{*T} + M_6
\]
\[
M_6 = A^*(H^*H^{*T})^{-1}A^{*T}M_5(I - P_{A^*}) + (I - P_{A^*}) M_5 A^*(H^*H^{*T})^{-1}A^{*T} + M_1
\]
And since $\left\|A^*(H^*H^{*T})^{-1}A^{*T}\right\|_{op} = \left\|(H^*H^{*T})^{-1}\right\|_{op}\leq \frac{1}{C_{H^*,k}}$ 
\[
\|M_6\|_{op} \leq L_{\max}^2\left(\frac{4C_{H^*,1}+12\sqrt{C_{H^*,1}}+8}{C_{H^*,k}}+\frac{8kC_E^2}{C_{H^*,k}^2}\right)\frac{\log n_B}{n_B} = C_{\tilde U_k}\frac{\log n_B}{n_B}
\]
Using the following facts: $A^*(H^*H^{*T})^{-1}A^{*T}(A^*H^*H^{*T}A^{*T}) = (A^*H^*H^{*T}A^{*T})A^*(H^*H^{*T})^{-1}A^{*T} = P_{A^*}$, $(I - P_{A^*})A^*H^* = 0$ and $H^{*T}A^{*T}(I - P_{A^*}) = 0$,
\[
P_{\tilde{U}_{k}} = P_{A^*} + A^*(H^*H^{*T})^{-1}H^*\Delta_{R}^T(I - P_{A^*}) + (I - P_{A^*}) \Delta_{R}H^{*T}(H^*H^{*T})^{-1}A^{*T}
\]
\[
+ \frac{1}{2}P_{A^*}(\hat\Sigma - I)(I - P_{A^*}) + \frac{1}{2}(I - P_{A^*})(\hat\Sigma - I)P_{A^*} + M_6
\]
\end{lemma}
\begin{lemma}[$P_{\hat A}$ Expansion]\label{lem:BEFS_1st_stage_P_hat_A_expansion}
We show 
\[
P_{\hat A} = P_{A^*} + A^*(H^*H^{*T})^{-1} H^*\Delta_R^T(I - P_{A^*}) +  (I - P_{A^*})\Delta_{R}H^{*T}(H^*H^{*T})^{-1}A^{*T}  + M
\]
For 
\[
M \leq  C_{\hat A}\frac{\log n_B }{n_B}
\]
Proof:\\
Using \cref{lem:BEFS_1st_stage_P_hat_form}, we obtained the form of $P_{\hat A}$. We additionally obtained a first order expansion of $P_{\tilde{U}_k} = \tilde{U}_{k}\tilde{U}_{k}^T = P_{A^*} + \zeta + M_1$ from \cref{lem:BEFS_1st_stage_P_U_expansion}, where $\tilde{U}_k$ is the top k left singular space of $\hat \Sigma^{1/2}Q_{ols}$ and
\[\|M_1\|_{op}\leq C_{\tilde U_k}\frac{\log n_B}{n_B}\]
Since $P_{\hat A} =P_{\hat \Sigma^{-1/2}\tilde{U}_k}$, the top k eigenvectors of $\hat{S} = \hat \Sigma^{-1/2}\tilde{U}_k\tilde{U}^T_{k}\hat \Sigma^{-1/2}$ span the top left k singular vectors of $\hat A$.

Call $D = (\hat \Sigma^{-1/2} - I)$. Then
\[\hat S = (I + D)(P_{A^*} + \zeta + M_1)(I +D) = P_{A^*} + E\]
where
\[E = \zeta + DP_{A^*}+P_{A^*}D + M_1+D\zeta+\zeta D+DM_1+M_1D+DP_{A^*}D+D\zeta D+DM_1D\]
Bounding $\|\zeta\|_{op}$ using $\|(H^*H^{*T})^{-1}H^*\|_{op} \leq \frac{1}{\sqrt {C_{H^*,k}}}$ gives
\[\|\zeta\|_{op} \leq \left(1 + \frac{2}{\sqrt {C_{H^*,k}}}\right)L_{\max}\sqrt\frac{\log n_B}{n_B}\]
Also, under $\mathcal E^{BEFS-1}$,
\[\|D\|_{op} \leq 2\|\hat \Sigma - I\|_{op} \leq 2L_1(2)\sqrt\frac{\log n_B}{n_B}\leq 2L_{\max}\sqrt\frac{\log n_B}{n_B}\]
Finally using $L_{\max}\sqrt{\frac{\log n_B}{n_B}} < 1$ along with Cauchy-Schwarz and the triangle inequality,
\[\|E\|_{op} \leq \left(17 + \frac{18}{\sqrt {C_{H^*,k}}} + 9C_{\tilde U_k}\right)L_{\max}\sqrt \frac{\log n_B}{n_B}=C_SL_{\max}\sqrt \frac{\log n_B}{n_B}\]
Using
\[L_{\max}\sqrt \frac{\log n_B}{n_B} < C_{BEFS,2}\]
we have
\[\|E\|_{op} < \frac14 <\frac12\]
Applying \cref{lem:general_projection_expansion}, since $\|E\|_{op} < \mu_k(P_{A^*})/2 = 1/2$,
\[P_{\hat A} = P_{A^*} + P_{A^*}^\dagger E ( I - P_{A^*}) + ( I - P_{A^*})EP_{A^*}^\dagger + M_2\]
Here
\[\|M_2\|_{op} \leq k \frac{4\|E\|_{op}^2}{1 - 2\|E\|_{op}}\]
Since $P_{A^*}^\dagger = P_{A^*}$,
\[P_{\hat A} = P_{A^*} + P_{A^*}E ( I - P_{A^*}) + ( I - P_{A^*})E P_{A^*} + M_2\]
Since $1 - 2\|E\|_{op} > 1/2$, we get
\[\|M_2\|_{op} \leq 8k\|E\|^2_{op} \leq 8kC_S^2L_{\max}^2\frac{\log n_B}{n_B}\]
Call
\[M_3 = M_1 + D\zeta + \zeta D + DM_1 + M_1D + DP_{A^*}D+ D\zeta D + DM_1D\]
Using Cauchy-Schwarz, the triangle inequality, and $L_{\max}\sqrt{\frac{\log n_B}{n_B}} <1$,
\[\|M_3\|_{op} \leq \left(9 C_{\tilde U_k} + \left(12 + \frac{16}{\sqrt {C_{H^*,k}}}\right)L_{\max}^2\right)\frac{\log n_B }{n_B}\]
Therefore
\[E = \zeta + D P_{A^*} + P_{A^*}D + M_3\]
Now pull out the first order part of $D$. Using $D = -\frac{1}{2}(\hat \Sigma - I) + M_4$ and
\[\|M_4\|_{op}\leq 3L_1(2)^2\frac{\log n_B}{n_B}\leq 3L_{\max}^2\frac{\log n_B}{n_B}\]
we get
\[E_{lin} = \zeta  - \frac{1}{2}(\hat \Sigma  - I) P_{A^*}  - \frac{1}{2}P_{A^*}(\hat \Sigma  - I)\]
Thus
\[E = E_{lin} + M_5\]
where $M_5 = M_3 + M_4P_{A^*} + P_{A^*}M_4$. Hence
\[\|M_5\|_{op}\leq \left(9 C_{\tilde U_k} + \left(18 + \frac{16}{\sqrt {C_{H^*,k}}}\right)L_{\max}^2\right)\frac{\log n_B }{n_B}\]
Now,
\[P_{\hat A} = P_{A^*}  + P_{A^*}(E_{lin} + M_5)( I - P_{A^*}) + ( I - P_{A^*})(E_{lin} + M_5)P_{A^*} + M_2\]
Plugging in $E_{lin}$, and using $(I - P_{A^*})P_{A^*} = 0$ and $P_{A^*}(I - P_{A^*}) = 0$,
\[P_{\hat A} = P_{A^*}  + P_{A^*}\zeta ( I - P_{A^*}) + ( I - P_{A^*})\zeta P_{A^*} - \frac{1}{2}P_{A^*}(\hat \Sigma - I)(I - P_{A^*}) -\frac{1}{2} ( I - P_{A^*})(\hat \Sigma - I) P_{A^*} + M_6\]
where
\[M_6 = (I - P_{A^*})M_5P_{A^*} + P_{A^*}M_5(I - P_{A^*}) + M_2\]
Therefore
\[\|M_6\|_{op} \leq C_{\hat A}\frac{\log n_B }{n_B}\]
where
\[C_{\hat A} = 18C_{\tilde U_k}+\left(36+\frac{32}{\sqrt {C_{H^*,k}}}\right)L_{\max}^2+8kC_S^2L_{\max}^2\]
Now plugging in $\zeta$, using $P_{A^*}A^*H^* = A^*H^*$, $(I - P_{A^*})P_{A^*} = 0$, and $P_{A^*}(I - P_{A^*})=0$,
\[P_{A^*}\zeta(I - P_{A^*}) = A^*(H^*H^{*T})^{-1}H^*\Delta_R^T(I - P_{A^*}) + \frac{1}{2}P_{A^*}(\hat \Sigma - I) (I - P_{A^*})\]
and
\[(I - P_{A^*})\zeta P_{A^*} = (I - P_{A^*})\Delta_{R}H^{*T}(H^*H^{*T})^{-1}A^{*T} + \frac{1}{2}(I - P_{A^*})(\hat \Sigma - I)P_{A^*}\]
Plugging these back into $P_{\hat A}$, the covariance fluctuation terms cancel, leaving
\[P_{\hat A} = P_{A^*} + A^*(H^*H^{*T})^{-1} H^*\Delta_R^T(I - P_{A^*}) +  (I - P_{A^*})\Delta_{R}H^{*T}(H^*H^{*T})^{-1}A^{*T}  + M_6\]
\end{lemma}
\begin{lemma}[K Expansion]\label{lem:1st_stage_BEFS_K_expansion}
For $K = (I - P_{A^*})P_{\hat A}P_{A^*}$, we show that for $M = O(\frac{\log n_B}{n_B})$,
\[
K = K_{lin} + M = (I - P_{A^*})\Delta_{R}H^{*T}(H^*H^{*T})^{-1}A^{*T} + M
\]
And it directly follows that $\|K\|_{op} = \sqrt \frac{\log n_B}{n_B}$.\\
Proof:\\
Plugging in the expansion of $P_{\hat A}$ into $K$, \cref{lem:BEFS_1st_stage_P_hat_A_expansion} with $\|M_1\|_{op} = O(\frac{\log n_B}{n_B})$
\[
K = (I - P_{A^*})P_{\hat A}P_{A^*} = (I - P_{A^*})(P_{A^*} + A^*(H^*H^{*T})^{-1} H^*\Delta_R^T(I - P_{A^*}) +  (I - P_{A^*})\Delta_{R}H^{*T}(H^*H^{*T})^{-1}A^{*T} + M_1)P_{A^*}
\]
\[
= (I - P_{A^*})\Delta_{R}H^{*T}(H^*H^{*T})^{-1}A^{*T} + (I - P_{A^*})M_1P_{A^*}
\]
\end{lemma}
\begin{lemma}[$\|P_{\hat A} - P_{A^*}\| \leq 1/2$]\label{lem:BEFS_1st_stage_projection_bound}
From \cref{lem:BEFS_1st_stage_P_hat_A_expansion}, using $\|(H^* H^{*T})^{-1}H^*\|_{op} \leq \frac{1}{\sqrt{C_{H^*,k}}}$, Cauchy-Schwarz, the triangle inequality and $\frac{\log n_B}{n_B} < 1$,
\[\|P_{\hat A} - P_{A^*}\|_{op} \leq \frac{2}{\sqrt {C_{H^*,k}}}\|\Delta_R\|_{op}+C_{\hat A}\frac{\log n_B}{n_B}\]
Under $\mathcal E^{BEFS-1}$,
\[\|\Delta_R\|_{op}\leq L_2(2)\sqrt{\frac{\log n_B}{n_B}}\]
so
\[\|P_{\hat A} - P_{A^*}\|_{op} \leq \frac{2L_2(2)}{\sqrt {C_{H^*,k}}}\sqrt{\frac{\log n_B}{n_B}}+C_{\hat A}\frac{\log n_B}{n_B}\]
Since $\frac{\log n_B}{n_B}<\sqrt{\frac{\log n_B}{n_B}}$,
\[\|P_{\hat A} - P_{A^*}\|_{op} \leq \left(\frac{2L_2(2)}{\sqrt {C_{H^*,k}}}+C_{\hat A}\right)\sqrt{\frac{\log n_B}{n_B}}\]
Equivalently,
\[\|P_{\hat A} - P_{A^*}\|_{op} \leq \left(\frac{2L_2(2)}{L_{\max}\sqrt {C_{H^*,k}}}+\frac{C_{\hat A}}{L_{\max}}\right)L_{\max}\sqrt{\frac{\log n_B}{n_B}}\]
Using $L_{\max}\sqrt{\frac{\log n_B}{n_B}}<C_{BEFS,3}$ and
\[
C_{BEFS,3}=\frac{L_{\max}}{2\left(\frac{2L_2(2)}{\sqrt{C_{H^*,k}}}+C_{\hat A}\right)}
\]
we have
\[\|P_{\hat A} - P_{A^*}\|_{op}<\frac{1}{2}\]
\end{lemma}
\begin{lemma}[$\beta^{*T}_{\hat{A}\perp}\Sigma_{test}\beta^{*}_{\hat{A}\perp}$ Second Order Expansion ]\label{lem:BEFS_stage_1_beta_norm_expansion}
We show that:
\[
\beta_{\hat A\perp}^{*T}\Sigma_{test}\beta^*_{\hat A\perp} = \beta^{*T}K_{lin}\Sigma_{A^*}K_{lin}^T\beta^* 
\]
\[
+ \beta_{A^*\perp}^{*T}\Sigma_{A^*\perp}\beta^*_{A^*\perp}-2\beta^{*T}_{A^*\perp}\Sigma_{A^*\perp}K\beta^*_{A^*}-2 \beta_{A^*\perp}^{*T}\Sigma_{A^*\perp}K_{lin}K_{lin}^T\beta_{A^*}^* + \beta_{A^*\perp}^{*T}K_{lin}^T\Sigma_{A^*\perp}K_{lin}\beta^*_{A^*\perp} + M
\]
where $M = O\left(\left(\frac{\log n_B}{n_B}\right)^{3/2}\right)$\\
Proof: \\

Using $K = (I - P_{A^*})P_{\hat A}P_{A^*}$, 
\[
P_{A^*}(I - P_{\hat A}) = P_{A^*} - P_{A^*}P_{\hat A} =  P_{A^*} - P_{A^*}P_{\hat A}(P_{A^*} + (I - P_{A^*}) = (P_{A^*} - P_{A^*}P_{\hat A}P_{A^*}) -K^T
\]
From \cref{lem:BEFS_1st_stage_projection_bound}, the event $\|P_{\hat A} - P_{A^*}\|_{op} \leq 1/2 $ holds under $\mathcal{E}^{BEFS-1}$.

Then using \cref{lem:general_K_expansion}, for $\|B\|_{op} \leq 4\|K\|^4_{op}$
\[
P_{A^*}(I - P_{\hat A})= K^TK - K^T + B
\]
Since from \cref{lem:1st_stage_BEFS_K_expansion}, $\|K\|_{op} = O(\sqrt\frac{\log n_B}{n_B})$ and $K = K_{lin} + M$ where $M = O(\frac{\log n_B}{n_B})$.
Then to 
And similarly
\[
(I - P_{A^*})(I - P_{\hat A}) = (I - P_{A^*}) - (I - P_{A^*})P_{\hat A} =  (I - P_{A^*}) - (I - P_{A^*})P_{\hat A}(P_{A^*} + (I - P_{A^*})) = 
\]
\[
= (I - P_{A^*}) - (I -P_{A^*})P_{\hat A} (I -P_{A^*}) - K
\] Using \cref{lem:general_K_expansion},
for $\|B\|^4_{op} \leq 4\|K\|_{op}$
\[
= (I - P_{A^*}) - KK^T - K - B
\]
Using the block test structure:
\[
\beta^{*T}_{\hat{A}\perp}\Sigma_{test}\beta^{*}_{\hat{A}\perp} = \beta^{*T}_{\hat{A}\perp}\Sigma_{A^*}\beta^{*}_{\hat{A}\perp} + \beta^{*T}_{\hat{A}\perp}\Sigma_{A^*\perp}\beta^{*}_{\hat{A}\perp}
\]
\[
\beta^{*T}_{\hat{A}\perp}\Sigma_{A^*}\beta^{*}_{\hat{A}\perp} = \beta^{*T}(I - P_{\hat A})P_{A^*}\Sigma_{A^*}P_{A^*}(I - P_{\hat A})\beta^* =  \beta^{*T}( K^TK - K^T + B)^T\Sigma_{A^*}(K^TK - K^T + B)\beta^*
\]
Then for $M_1 = O(\|K\|_{op}^2)$
\[
\beta^{*T}_{\hat{A}\perp}\Sigma_{A^*}\beta^{*}_{\hat{A}\perp} = \beta^{*T}K\Sigma_{A^*}K^T\beta^* + M_1
\]
And 
\[
\beta^{*T}_{\hat{A}\perp}\Sigma_{A^*\perp}\beta^{*}_{\hat{A}\perp} = \beta^{*T}(I - P_{\hat A})(I -P_{A^*})\Sigma_{A^*\perp}(I - P_{A^*})(I - P_{\hat A})\beta^*
\]
\[
 =  \beta^{*T}( (I - P_{A^*}) - KK^T - K - B)^T\Sigma_{A^*\perp}((I - P_{A^*}) - KK^T - K - B)\beta^*
\]
Then for $M_2 = O(\|K\|_{op}^2)$
\[
= \beta_{A^*\perp}^{*T}\Sigma_{A^*\perp}\beta^*_{A^*\perp}-2\beta^{*T}_{A^*\perp}\Sigma_{A^*\perp}K\beta^*_{A^*}-2 \beta_{A^*\perp}^{*T}\Sigma_{A^*\perp}KK^T\beta_{A^*\perp}^* + \beta_{A^*}^{*T}K^T\Sigma_{A^*\perp}K\beta^*_{A^*} + M_2
\]
Then the total expression is given by $M_3 = M_1 + M_2 =O(\|K\|_{op}^2)$
\[
\beta_{\hat A\perp}^{*T}\Sigma_{test}\beta^*_{\hat A\perp} = \beta^{*T}K\Sigma_{A^*}K^T\beta^* 
\]
\[
+ \beta_{A^*\perp}^{*T}\Sigma_{A^*\perp}\beta^*_{A^*\perp}-2\beta^{*T}_{A^*\perp}\Sigma_{A^*\perp}K\beta^*_{A^*}-2 \beta_{A^*\perp}^{*T}\Sigma_{A^*\perp}KK^T\beta_{A^*}^* + \beta_{A^*\perp}^{*T}K^T\Sigma_{A^*\perp}K\beta^*_{A^*\perp} + M_3
\]
Plugging in $K_{lin}$ for all $K^2$ terms, $M_4 = O\left(\left(\frac {\log n_B }{n_B}\right)^{3/2}\right)$
\[
\beta_{\hat A\perp}^{*T}\Sigma_{test}\beta^*_{\hat A\perp} = \beta^{*T}K_{lin}\Sigma_{A^*}K_{lin}^T\beta^* 
\]
\[
+ \beta_{A^*\perp}^{*T}\Sigma_{A^*\perp}\beta^*_{A^*\perp}-2\beta^{*T}_{A^*\perp}\Sigma_{A^*\perp}K\beta^*_{A^*}-2 \beta_{A^*\perp}^{*T}\Sigma_{A^*\perp}K_{lin}K_{lin}^T\beta_{A^*}^* + \beta_{A^*\perp}^{*T}K_{lin}^T\Sigma_{A^*\perp}K_{lin}\beta^*_{A^*\perp} + M_4
\]
\end{lemma}
\begin{lemma}\label{lem:BEFS_K_lin_scaling_prelims}
From \cref{lem:1st_stage_BEFS_K_expansion} define
\[
K_{lin} = (I - P_{A^*})\Delta_{R}H^{*T}(H^*H^{*T})^{-1}A^{*T}
\]
Additionally, from \cref{lem:basics}, $\mathbb{E}_{E_r, E_L}[\Delta_R \Delta_R^T|X^{(B)}] = \text{Tr}(\sigma_{r}^2I +H^T\Sigma_{{\ell_{H^*}}}H)(X^{(B)T}X^{(B)})^{-1}$ and for fixed $G$, $
\mathbb{E}_{E_r, E_L}[\Delta_R G\Delta_R^T|X^{(B)}] = \text{Tr}\left((\sigma_{r}^2I +H^T\Sigma_{{\ell_{H^*}}}H)G\right)(X^{(B)T}X^{(B)})^{-1}
$,\\
$\mathbb{E}_{E_r, E_L}[\Delta_R^T G\Delta_R|X^{(B)}] = \text{Tr}(G(X^{(B)T}X^{(B)})^{-1})\left(\sigma_{r}^2I +H^T\Sigma_{{\ell_{H^*}}}H\right)$.

We define a few preliminary scaling law quantities\\
\textbf{1)}
\[
\mathbb{E}_{E_L, E_R, X^{(B)}}[K_{lin}^T\Sigma_{A^*\perp}K_{lin}] = \mathbb{E}_{X^{(B)}}[\mathbb{E}_{E_L, E_R}[K_{lin}^T\Sigma_{A^*\perp}K_{lin} | X^{(B)}]] 
\]
\[
\mathbb{E}_{E_L, E_R}[K_{lin}^T\Sigma_{A^*\perp}K_{lin}|X^{(B)}] = \text{Tr}(\Sigma_{A^*\perp}(X^{(B)T}X^{(B)})^{-1}) A^*(H^*H^{*T})^{-1}H^*\left(\sigma_{r}^2I +H^{*T}\Sigma_{{\ell_{H^*}}}H^*\right)H^{*T}(H^*H^{*T})^{-1}A^{*T}
\]
\[
= \text{Tr}(\Sigma_{A^*\perp}(X^{(B)T}X^{(B)})^{-1})A^*(\sigma_r(H^*H^{*T})^{-1} + \Sigma_{{\ell_{H^*}}})A^{*T}
\]
And $\mathbb{E}_{X^{(B)}}[(X^{(B)T}X^{(B)})^{-1}] = \frac{1}{n_B- d_x -1} I$ so, 
\[
\mathbb{E}_{E_L, E_R, X^{(B)}}[K_{lin}^T\Sigma_{A^*\perp}K_{lin}] = \frac{1}{n_B -d_x -1}\text{Tr}(\Sigma_{A^*\perp})A^*(\sigma_r(H^*H^{*T})^{-1} + \Sigma_{{\ell_{H^*}}})A^{*T}
\]\\
\textbf{2)}
\[
\mathbb{E}_{E_L, E_R, X^{(B)}}[K_{lin}\Sigma_{A^*}K_{lin}^T] = \mathbb{E}_{E_L, E_R, }[\mathbb{E}_{X^{(B)}}[K_{lin}\Sigma_{A^*}K_{lin}^T| X^{(B)}]]
\]
\[
= \text{Tr}((\sigma_r^2I + H^{*T}\Sigma_{\ell_{H^*}} H^*)H^{*T}(H^*H^{*T})^{-1}A^{*T}\Sigma_{A^*} A^*(H^*H^{*T})^{-1}H^*) (I - P_{A^*})(X^{(B)T}X^{(B)})^{-1}(I - P_{A^*})
\]
And taking the expectation on $X^{(B)}$
\[
\mathbb{E}_{E_L, E_R, X^{(B)}}[K_{lin}\Sigma_{A^*}K_{lin}^T]= \frac{1}{n_B -d_x -1}\text{Tr}(\Sigma_{A^*}A^*(\sigma^2_r(H^*H^{*T})^{-1} + \Sigma_{{\ell_{H^*}}})A^{*T}) (I - P_{A^*})
\]
\textbf{3)}
\[
\mathbb{E}_{E_L, E_R, X^{(B)}}[K_{lin}K_{lin}^T] = \mathbb{E}_{E_L, E_R, X^{(B)}}[K_{lin}P_{A^*}K_{lin}^T] 
\]
Using 2),
\[
=\frac{1}{n_B -d_x -1}\text{Tr}(A^*(\sigma_r^2(H^*H^{*T})^{-1} + \Sigma_{{\ell_{H^*}}})A^{*T}) (I - P_{A^*})
\]
\end{lemma}
\begin{lemma}[Scaling of $\beta^{*T}_{\hat{A}\perp}\Sigma_{test}\beta^{*}_{\hat{A}\perp}$]\label{lem:BEFS_1st_stage_beta_norm_scaling}
Let $\boldsymbol{1}_{\mathcal{E}}^{BEFS-1}$ be the event indicator function and $\boldsymbol{1}_{\mathcal{E}^c}^{BEFS-1} $ be its complement. $\beta^{*T}_{\hat{A}\perp}\Sigma_{test}\beta^{*}_{\hat{A}\perp} = Z_{main} + M$ for leading terms $Z_{main}$.
\[
\mathbb{E}_{X^{(B)}, R^{(B)}}[\beta^{*T}_{\hat{A}\perp}\Sigma_{test}\beta^{*}_{\hat{A}\perp}] = \mathbb{E}[Z_{main}] + \mathbb{E}[M\boldsymbol{1}_{\mathcal{E}}^{BEFS-1}] + \mathbb{E}[(\beta^{*T}_{\hat{A}\perp}\Sigma_{test}\beta^{*}_{\hat{A}\perp} - Z_{main})\boldsymbol{1}_{\mathcal{E}^C}^{BEFS-1,}]
\]
First bounding:
\[
\mathbb{E}[(\beta^{*T}_{\hat{A}\perp}\Sigma_{test}\beta^{*}_{\hat{A}\perp} - Z_{main})\boldsymbol{1}_{\mathcal{E}^C}^{BEFS-1,}] = \mathbb{E}[\beta^{*T}_{\hat{A}\perp}\Sigma_{test}\beta^{*}_{\hat{A}\perp}\boldsymbol{1}_{\mathcal{E}^C}^{BEFS-1,}]  -\mathbb{E}[Z_{main}\boldsymbol{1}_{\mathcal{E}^C}^{BEFS-1,}]
\]
Since $P(\mathcal{E}_C^{BEFS-1}) < Cn_B^{-2} $ and
\[
\beta^{*T}_{\hat{A}\perp}\Sigma_{test}\beta^{*}_{\hat{A}\perp} \leq m_1
\]
\[
\mathbb{E}[\beta^{*T}_{\hat{A}\perp}\Sigma_{test}\beta^{*}_{\hat{A}\perp}\boldsymbol{1}_{\mathcal{E}^C}^{BEFS-1,}]  = o(n_B^{-1})
\]
Since \[
\|K_{lin}\|_{op}
\le
\|\Delta_R\|_{op}\,\|H^{*T}(H^*H^{*T})^{-1}A^{*T}\|_{op}
\le
\frac{\|\Delta_R\|_{op}}{\sqrt{C_k}}.
\]
and $\|K\|_{op}\le 1,$
\[
\mathbb{E}[|Z_{main}|\boldsymbol{1}_{\mathcal{E}^C}^{BEFS-1,}]\leq m_2 P(\mathcal{E}_C^{BEFS-1}) + m_3\mathbb{E}[\|\Delta_R\|^2_{op}\boldsymbol{1}_{\mathcal{E}^C}^{BEFS-1,}]
\]
\[
\leq m_2 P(\mathcal{E}_C^{BEFS-1}) + m_3\mathbb{E}[\|\Delta_R\|^4_{op}]\mathbb{E}[\boldsymbol{1}_{\mathcal{E}^C}^{BEFS-1,}] = o(n_B^{-1})
\]
Finally since $M = O\left(\left(\frac {\log n_B }{n_B}\right)^{3/2}\right) = o(n_B^{-1})$
\[
\mathbb{E}_{X^{(B)}, R^{(B)}}[\beta^{*T}_{\hat{A}\perp}\Sigma_{test}\beta^{*}_{\hat{A}\perp}] = \mathbb{E}[Z_{main}] + o(n_B^{-1})
\]
Now moving onto the main piece of $\mathbb{E}[Z_{main}]$
\[
\mathbb{E}_{X^{(B)}, R^{(B)}}[Z_{main}] =  \mathbb{E}_{X^{(B)}, R^{(B)}}[\beta^{*T}K_{lin}\Sigma_{A^*}K_{lin}^T\beta^*]+ \beta_{A^*\perp}^{*T}\Sigma_{A^*\perp}\beta^*_{A^*\perp}-2\mathbb{E}_{X^{(B)}, R^{(B)}}[\beta^{*T}_{A^*\perp}\Sigma_{A^*\perp}K\beta^*_{A^*}]
\]
\[
-2 \mathbb{E}_{X^{(B)}, R^{(B)}}[\beta_{A^*\perp}^{*T}\Sigma_{A^*\perp}K_{lin}K_{lin}^T\beta_{A^*}^*] + \mathbb{E}_{X^{(B)}, R^{(B)}}[\beta_{A^*\perp}^{*T}K_{lin}^T\Sigma_{A^*\perp}K_{lin}\beta^*_{A^*\perp}] 
\]
Using \cref{lem:K_expectation},
\[
-2\mathbb{E}_{X^{(B)}, R^{(B)}}[\beta^{*T}_{A^*\perp}\Sigma_{A^*\perp}K\beta^*_{A^*}] = 0
\]

Plugging in the scaling quantities in \cref{lem:BEFS_K_lin_scaling_prelims}, 
\[
\mathbb{E}_{E_R, E_L, X^{(B)}}[\beta^{*T}K_{lin}\Sigma_{A^*}K_{lin}^T\beta^*] = \frac{\|\beta^{*}_{A^*\perp}\|^2}{n_B-d_x -1}\text{Tr}(\Sigma_{A^*}A^*(\sigma^2_r(H^*H^{*T})^{-1} + \Sigma_{{\ell_{H^*}}})A^{*T})
\]
\[
\mathbb{E}_{E_R, E_L, X^{(B)}}[-2 \beta_{A^*\perp}^{*T}\Sigma_{A^*\perp}K_{lin}K_{lin}^T\beta_{A^*}^*] = -2\frac{\beta_{A^*\perp}^{*T}\Sigma_{A^*\perp}\beta_{A^*\perp}}{n_B-d_x -1}\text{Tr}(A^*(\sigma^2_r(H^*H^{*T})^{-1} + \Sigma_{{\ell_{H^*}}})A^{*T})
\]
\[\mathbb{E}_{E_R, E_L, X^{(B)}}[\beta_{A^*}^{*T}K_{lin}^T\Sigma_{A^*\perp}K_{lin}\beta^*_{A^*}]= \frac{1}{n_B -d_x -1}\text{Tr}(\Sigma_{A^*\perp})\beta^{*T}A^*(\sigma^2_r(H^*H^{*T})^{-1} + \Sigma_{{\ell_{H^*}}})A^{*T}\beta^*
\]
Call $\Sigma_{est} = A^*(\sigma^2_r(H^*H^{*T})^{-1} + \Sigma_{{\ell_{H^*}}})A^{*T}$, then the total leading order scaling law is:
\[
\mathbb{E}_{E_R, E_L, X^{(B)}}[
Z_{main}] =
\]
\[
\beta_{A^*\perp}^{*T}\Sigma_{A^*\perp}\beta^*_{A^*\perp} + \frac{1}{n_B - d_x -1}\left[(\|\beta^*_{A^*\perp}\|^2\text{Tr}(\Sigma_{A^*}\Sigma_{est}) - 2\beta_{A^*\perp}^{*T}\Sigma_{A^*\perp}\beta_{A^*\perp}\text{Tr}(\Sigma_{est}) + \beta^{*T}\Sigma_{est}\beta^{*}\text{Tr}(\Sigma_{A^*\perp})\right]
\]
Then the total scaling is
\[
\mathbb{E}_{X^{(B)}, R^{(B)}}[\beta^{*T}_{\hat{A}\perp}\Sigma_{test}\beta^{*}_{\hat{A}\perp}]  = 
\]
\[
\beta_{A^*\perp}^{*T}\Sigma_{A^*\perp}\beta^*_{A^*\perp} + \frac{1}{n_B - d_x -1}\left[(\|\beta^*_{A^*\perp}\|^2\text{Tr}(\Sigma_{A^*}\Sigma_{est}) - 2\beta_{A^*\perp}^{*T}\Sigma_{A^*\perp}\beta_{A^*\perp}\text{Tr}(\Sigma_{est}) + \beta^{*T}\Sigma_{est}\beta^{*}\text{Tr}(\Sigma_{A^*\perp})\right] + o(n_B^{-1})
\]
And simplifying the wishart denominator at first order
\[
\beta_{A^*\perp}^{*T}\Sigma_{A^*\perp}\beta^*_{A^*\perp} + \frac{1}{n_B }\left[(\|\beta^*_{A^*\perp}\|^2\text{Tr}(\Sigma_{A^*}\Sigma_{est}) - 2\beta_{A^*\perp}^{*T}\Sigma_{A^*\perp}\beta_{A^*\perp}\text{Tr}(\Sigma_{est}) + \beta^{*T}\Sigma_{est}\beta^{*}\text{Tr}(\Sigma_{A^*\perp})\right] + o(n_B^{-1})
\]
And under isotropic test, $\Sigma_{A^*} = P_{A^*}$ and $\Sigma_{A^*\perp} =(I -  P_{A^*})$ the leading scaling law becomes
\[
= \|\beta_{A^*\perp}\|^2 + \frac{1}{n_B}\left[( \beta^{*T}\Sigma_{est}\beta^{*}(d_x - d_{\ell_{H^*}}) - \|\beta^*_{A^*\perp}\|^2\text{Tr}(\Sigma_{est})\right] + o(n_B^{-1})
\]
\end{lemma}
\begin{lemma}[$\text{Tr}(J_{\hat A}\Sigma_{test}J_{\hat A})$ Second Order Expansion]\label{lem:BEFS_1st_stage_trace_second_order_expansion}
\[
\text{Tr}(J_{\hat A}\Sigma_{test}J_{\hat A}) = \text{Tr}(\Sigma_{test}(\alpha^2I + (1-\alpha)^2P_{\hat A})) = \alpha^2\text{Tr}(\Sigma_{test}) + (1-\alpha)^2\text{Tr}(\Sigma_{A^*}P_{\hat A} + \Sigma_{A^*\perp}P_{\hat A}) 
\]
\[
=\alpha^2\text{Tr}(\Sigma_{test}) + (1-\alpha^2)\text{Tr}(\Sigma_{A^*}P_{A^*}P_{\hat A}P_{A^*}) + (1-\alpha^2)\text{Tr}(\Sigma_{A^*\perp}(I - P_{A^*})P_{\hat A}(I - P_{A^*})) 
\]
Since by \cref{lem:BEFS_1st_stage_projection_bound}, $\|P_{\hat A} - P_{A^*}\| \leq 1/2$, then using \cref{lem:general_K_expansion} for $\|M_1\|_{op} = O(\|K\|^4)$
\[
\text{Tr}(\Sigma_{A^*}P_{A^*}P_{\hat A}P_{A^*})  = \text{Tr}(\Sigma_{A^*}(P_{A^*} - K^TK - B)) = \text{Tr}(\Sigma_{A^*}) - \text{Tr}(\Sigma_{A^*}K^TK) + M_1
\]
And using \cref{lem:general_K_expansion} for $\|M_2\|_{op} = O(\|K\|^4)$
\[
\text{Tr}(\Sigma_{A^*\perp}(I - P_{A^*})P_{\hat A}(I - P_{A^*})) = \text{Tr}(\Sigma_{A^*\perp}\left(KK^T +B \right) = \text{Tr}(\Sigma_{A^*\perp}KK^T) + M_2
\]
then for $\|M_3\|_{op} = O((1-\alpha^2)\|K\|^4)$,
\[
\text{Tr}(J_{\hat A}\Sigma_{test}J_{\hat A}) = \alpha^2\text{Tr}(\Sigma_{test})  + (1-\alpha^2)[\text{Tr}(\Sigma_{A^*}) -\text{Tr}(\Sigma_{A^*}K^TK) + \text{Tr}(\Sigma_{A^*\perp}KK^T)] + M_3
\]
\[
= \text{Tr}(\Sigma_{A^*}) + \alpha^2\text{Tr}(\Sigma_{A^*\perp}) + (1-\alpha^2)[\text{Tr}(K^T\Sigma_{A^*\perp}K) -\text{Tr}(K\Sigma_{A^*}K^T)] + M_3
\]
As a sanity check, plugging in $\Sigma_{test} = I = P_{A^*} + P_{A^*\perp}$ yields $d_{\ell_{H^*}}+ \alpha^2\text{Tr}(d_x - d_{\ell_{H^*}})$ which is the correct reduction. 

Using \cref{lem:1st_stage_BEFS_K_expansion}, for $M_4= O(\frac{\log n_B }{n_B})$
\[
K = K_{lin} + M_4
\]
So, for $M_5 = O\left((1-\alpha^2)(\frac{ \log n_B}{n_B})^{3/2}\right)$
\[
\text{Tr}(J_{\hat A}\Sigma_{test}J_{\hat A}) = \text{Tr}(\Sigma_{A^*}) + \alpha^2\text{Tr}(\Sigma_{A^*\perp}) + (1-\alpha^2)[\text{Tr}(K_{lin}^T\Sigma_{A^*\perp}K_{lin{}}) -\text{Tr}(K_{lin}\Sigma_{A^*}K_{lin}^T)] + M_5
\]
\end{lemma}
\begin{lemma}[Scaling law of $\text{Tr}(J_{\hat A}\Sigma_{test}J_{\hat A}) $]\label{lem:BEFS_1st_Stage_trace_scaling_law}
From \cref{lem:BEFS_1st_stage_trace_second_order_expansion}, for $\|M_1\|_{op} = O\left((1-\alpha^2)\left(\frac{\log n_B }{n_B}\right)^{3/2}\right)$
\[
\text{Tr}(J_{\hat A}\Sigma_{test}J_{\hat A}) = Z_{main} + M_1
\]
\[
Z_{main} =\text{Tr}(\Sigma_{A^*}) + \alpha^2\text{Tr}(\Sigma_{A^*\perp})+ (1-\alpha^2)\mathbb{E}_{E_L, E_R, X^{(B)}}[\text{Tr}(K_{lin}^T\Sigma_{A^*\perp}K_{lin}) -\text{Tr}(K_{lin}\Sigma_{A^*}K_{lin}^T)]
\]
\[
\mathbb{E}_{X^{(B)}, R^{(B)}}[\text{Tr}(J_{\hat A}\Sigma_{test}J_{\hat A})] = \mathbb{E}[Z_{main}] + \mathbb{E}[M\boldsymbol{1}_{\mathcal{E}}^{BEFS-1}] + \mathbb{E}[(\text{Tr}(J_{\hat A}\Sigma_{test}J_{\hat A}) - Z_{main})\boldsymbol{1}_{\mathcal{E}^C}^{BEFS-1,}]
\]
\[
\mathbb{E}[M\boldsymbol{1}_{\mathcal{E}}^{BEFS-1}] = O\left(\left(\frac{\log n_B }{n_B}\right)^2\right)
\]
\[
\mathbb{E}[(\text{Tr}(J_{\hat A}\Sigma_{test}J_{\hat A}) - Z_{main})\boldsymbol{1}_{\mathcal{E}^C}^{BEFS-1,}] \leq m_1P(\mathcal{E}_C^{BEFS-1}) + \mathbb{E}[|Z_{main}|^{2}]^{1/2}P(\mathcal{E}_C^{BEFS-1})^{1/2} = o((1-\alpha^2)n^{-1})
\]
Next solving for $\mathbb{E}[Z_{min}]$, 
Plugging in the terms using \cref{lem:BEFS_K_lin_scaling_prelims}, the expectation on the random parts of $Z_min$ are given by:
\[
(1-\alpha^2)\mathbb{E}_{E_L, E_R, X^{(B)}}[\text{Tr}(K_{lin}^T\Sigma_{A^*\perp}K_{lin}) -\text{Tr}(K_{lin}\Sigma_{A^*}K_{lin}^T)]
\]
\[
\mathbb{E}_{E_L, E_R, X^{(B)}}[\text{Tr}(K_{lin}^T\Sigma_{A^*\perp}K_{lin})] = \frac{1}{n_B -d_x -1}\text{Tr}(\Sigma_{A^*\perp})\text{Tr}\left(A^*(\sigma^2_r(H^*H^{*T})^{-1} + \Sigma_{{\ell_{H^*}}})A^{*T}\right)
\]
and 
\[
\mathbb{E}_{E_L, E_R, X^{(B)}}[\text{Tr}(K_{lin}\Sigma_{A^*}K_{lin}^T)] = \frac{1}{n_B -d_x -1}\text{Tr}(\Sigma_{A^*}A^*(\sigma^2_r(H^*H^{*T})^{-1} + \Sigma_{{\ell_{H^*}}})A^{*T})(d_x - d_{\ell_{H^*}})
\]
writing $\Sigma_{est} = A^*(\sigma^2_r(H^*H^{*T})^{-1} + \Sigma_{{\ell_{H^*}}})A^{*T}$
\[
\mathbb{E}_{E_L, E_R, X^{(B)}}[Z_{main}] = \text{Tr}(\Sigma_{A^*}) + \alpha^2\text{Tr}(\Sigma_{A^*\perp}) + \frac{(1-\alpha^2)}{n_B -d_x -1}[\text{Tr}(\Sigma_{A^*\perp})\text{Tr}(\Sigma_{est})-\text{Tr}(\Sigma_{A^*}\Sigma_{est})(d_x - d_{\ell_{H^*}})]
\]
So the total scaling is given by:
\[
\mathbb{E}_{E_L, E_R, X^{(B)}}[\text{Tr}(J_{\hat A}\Sigma_{test}J_{\hat A})]  = \text{Tr}(\Sigma_{A^*}) + \alpha^2\text{Tr}(\Sigma_{A^*\perp}) 
\]
\[
+ \frac{(1-\alpha^2)}{n_B -d_x -1}[\text{Tr}(\Sigma_{A^*\perp})\text{Tr}(\Sigma_{est})-\text{Tr}(\Sigma_{A^*}\Sigma_{est})(d_x - d_{\ell_{H^*}})] + o((1-\alpha^2)n_B^{-1})
\]
And simplifying the wishart denominator to first order
\[
\mathbb{E}_{E_L, E_R, X^{(B)}}[\text{Tr}(J_{\hat A}\Sigma_{test}J_{\hat A})]  = \text{Tr}(\Sigma_{A^*}) + \alpha^2\text{Tr}(\Sigma_{A^*\perp}) 
\]
\[
+ \frac{(1-\alpha^2)}{n_B}[\text{Tr}(\Sigma_{A^*\perp})\text{Tr}(\Sigma_{est})-\text{Tr}(\Sigma_{A^*}\Sigma_{est})(d_x - d_{\ell_{H^*}})] + o((1-\alpha^2)n_B^{-1})
\]
and under isotropic test, 
\[
\mathbb{E}_{E_L, E_R, X^{(B)}}[\text{Tr}(J_{\hat A}\Sigma_{test}J_{\hat A})] = d_{\ell_{H^*}} + \alpha^2(d_x - d_{\ell_{H^*}}) + o((1-\alpha^2)n_B^{-1})
\]
\end{lemma}
\begin{lemma}[$\text{Tr}(P_{\hat A\perp}\Sigma_{test})$ Second Order Expansion]\label{lem:BEFS_1st_stage_P_hat_A_perp_trace_second_order_expansion}
We have
\[\text{Tr}(P_{\hat A\perp}\Sigma_{test})=\text{Tr}(P_{\hat A\perp}\Sigma_{A^*})+\text{Tr}(P_{\hat A\perp}\Sigma_{A^*\perp})\]
\[\text{Tr}(P_{\hat A\perp}\Sigma_{test})=\text{Tr}(\Sigma_{A^*}P_{A^*}P_{\hat A\perp}P_{A^*})+\text{Tr}(\Sigma_{A^*\perp}(I-P_{A^*})P_{\hat A\perp}(I-P_{A^*}))\]
Since by \cref{lem:BEFS_1st_stage_projection_bound},
\[\|P_{\hat A}-P_{A^*}\|_{op}\leq \frac{1}{2}\]
we may use \cref{lem:general_K_expansion}. For $\|M_1\|_{op}=O(\|K\|^4)$,
\[\text{Tr}(\Sigma_{A^*}P_{A^*}P_{\hat A\perp}P_{A^*})=\text{Tr}(\Sigma_{A^*}(K^TK+B))\]
\[\text{Tr}(\Sigma_{A^*}P_{A^*}P_{\hat A\perp}P_{A^*})=\text{Tr}(\Sigma_{A^*}K^TK)+M_1\]
Similarly, for $\|M_2\|_{op}=O(\|K\|^4)$,
\[\text{Tr}(\Sigma_{A^*\perp}(I-P_{A^*})P_{\hat A\perp}(I-P_{A^*}))=\text{Tr}(\Sigma_{A^*\perp}(I-P_{A^*}-KK^T-B))\]
\[\text{Tr}(\Sigma_{A^*\perp}(I-P_{A^*})P_{\hat A\perp}(I-P_{A^*}))=\text{Tr}(\Sigma_{A^*\perp})-\text{Tr}(\Sigma_{A^*\perp}KK^T)+M_2\]
Therefore, for $\|M_3\|_{op}=O(\|K\|^4)$,
\[\text{Tr}(P_{\hat A\perp}\Sigma_{test})=\text{Tr}(\Sigma_{A^*\perp})+\text{Tr}(\Sigma_{A^*}K^TK)-\text{Tr}(\Sigma_{A^*\perp}KK^T)+M_3\]
Equivalently,
\[\text{Tr}(P_{\hat A\perp}\Sigma_{test})=\text{Tr}(\Sigma_{A^*\perp})+\text{Tr}(K\Sigma_{A^*}K^T)-\text{Tr}(K^T\Sigma_{A^*\perp}K)+M_3\]
Using \cref{lem:1st_stage_BEFS_K_expansion}, for
\[M_4=O\left(\frac{\log n_B}{n_B}\right)\]
we have
\[K=K_{lin}+M_4\]
Therefore, for
\[M_5=O\left(\left(\frac{\log n_B}{n_B}\right)^{3/2}\right)\]
we obtain
\[\text{Tr}(P_{\hat A\perp}\Sigma_{test})=\text{Tr}(\Sigma_{A^*\perp})+\text{Tr}(K_{lin}\Sigma_{A^*}K_{lin}^T)-\text{Tr}(K_{lin}^T\Sigma_{A^*\perp}K_{lin})+M_5\]
\end{lemma}

\begin{lemma}[Scaling law of $\text{Tr}(P_{\hat A\perp}\Sigma_{test})$]\label{lem:BEFS_1st_stage_P_hat_A_perp_trace_scaling_law}
From \cref{lem:BEFS_1st_stage_P_hat_A_perp_trace_second_order_expansion}, for
\[M_1=O\left(\left(\frac{\log n_B}{n_B}\right)^{3/2}\right)\]
we have
\[\text{Tr}(P_{\hat A\perp}\Sigma_{test})=Z_{main}+M_1\]
where
\[Z_{main}=\text{Tr}(\Sigma_{A^*\perp})+\text{Tr}(K_{lin}\Sigma_{A^*}K_{lin}^T)-\text{Tr}(K_{lin}^T\Sigma_{A^*\perp}K_{lin})\]
Taking expectation over the first-stage data,
\[\mathbb E_{X^{(B)},R^{(B)}}[\text{Tr}(P_{\hat A\perp}\Sigma_{test})]=\mathbb E[Z_{main}]+\mathbb E[M_1\boldsymbol 1_{\mathcal E^{BEFS-1}}]+\mathbb E[(\text{Tr}(P_{\hat A\perp}\Sigma_{test})-Z_{main})\boldsymbol 1_{\mathcal E^{BEFS-1,C}}]\]
On the good event,
\[\mathbb E[M_1\boldsymbol 1_{\mathcal E^{BEFS-1}}]=O\left(\left(\frac{\log n_B}{n_B}\right)^{3/2}\right)\]
On the complement, using the same argument as in \cref{lem:BEFS_1st_Stage_trace_scaling_law},
\[\mathbb E[(\text{Tr}(P_{\hat A\perp}\Sigma_{test})-Z_{main})\boldsymbol 1_{\mathcal E^{BEFS-1,C}}]=o(n_B^{-1})\]
Therefore,
\[\mathbb E_{X^{(B)},R^{(B)}}[\text{Tr}(P_{\hat A\perp}\Sigma_{test})]=\mathbb E[Z_{main}]+o(n_B^{-1})\]
Next, plugging in the terms from \cref{lem:BEFS_K_lin_scaling_prelims}, we have
\[\mathbb E_{E_L,E_R,X^{(B)}}[\text{Tr}(K_{lin}^T\Sigma_{A^*\perp}K_{lin})]=\frac{1}{n_B-d_x-1}\text{Tr}(\Sigma_{A^*\perp})\text{Tr}\left(A^*\left(\sigma_r^2(H^*H^{*T})^{-1}+\Sigma_{\ell_{H^*}}\right)A^{*T}\right)\]
and
\[\mathbb E_{E_L,E_R,X^{(B)}}[\text{Tr}(K_{lin}\Sigma_{A^*}K_{lin}^T)]=\frac{1}{n_B-d_x-1}\text{Tr}\left(\Sigma_{A^*}A^*\left(\sigma_r^2(H^*H^{*T})^{-1}+\Sigma_{\ell_{H^*}}\right)A^{*T}\right)(d_x-d_{\ell_{H^*}})\]
Writing
\[\Sigma_{est}=A^*\left(\sigma_r^2(H^*H^{*T})^{-1}+\Sigma_{\ell_{H^*}}\right)A^{*T}\]
we get
\[\mathbb E[Z_{main}]=\text{Tr}(\Sigma_{A^*\perp})-\frac{1}{n_B-d_x-1}\left[\text{Tr}(\Sigma_{A^*\perp})\text{Tr}(\Sigma_{est})-\text{Tr}(\Sigma_{A^*}\Sigma_{est})(d_x-d_{\ell_{H^*}})\right]\]
Hence the total scaling is
\[\mathbb E_{X^{(B)},R^{(B)}}[\text{Tr}(P_{\hat A\perp}\Sigma_{test})]=\text{Tr}(\Sigma_{A^*\perp})-\frac{1}{n_B-d_x-1}\left[\text{Tr}(\Sigma_{A^*\perp})\text{Tr}(\Sigma_{est})-\text{Tr}(\Sigma_{A^*}\Sigma_{est})(d_x-d_{\ell_{H^*}})\right]+o(n_B^{-1})\]
Simplifying the Wishart denominator to first order gives
\[\mathbb E_{X^{(B)},R^{(B)}}[\text{Tr}(P_{\hat A\perp}\Sigma_{test})]=\text{Tr}(\Sigma_{A^*\perp})-\frac{1}{n_B}\left[\text{Tr}(\Sigma_{A^*\perp})\text{Tr}(\Sigma_{est})-\text{Tr}(\Sigma_{A^*}\Sigma_{est})(d_x-d_{\ell_{H^*}})\right]+o(n_B^{-1})\]
\end{lemma}
\begin{lemma}[ $\|\beta^*_{\hat A\perp}\|^2\text{Tr}(J_{\hat A}\Sigma_{test}J_{\hat A})$ Second Order Expansion]\label{lem:befs_1st_stage_mixed_expansion}
\[
\|\beta^*_{\hat A\perp}\|^2\text{Tr}(J_{\hat A}\Sigma_{test}J_{\hat A})
\]
From \cref{lem:BEFS_stage_1_beta_norm_expansion}, $M_1 = O\left((\frac{\log n_B}{n_B})^{3/2}\right)$
\[
\|\beta^*_{\hat A\perp}\|^2 = \beta_{\hat A\perp}^{*T}\Sigma_{test}\beta^*_{\hat A\perp} = \beta^{*T}K_{lin}K_{lin}^T\beta^* 
\]
\[
+ \beta_{A^*\perp}^{*T}\beta^*_{A^*\perp}-2\beta^{*T}_{A^*\perp}K\beta^*_{A^*}-2 \beta_{A^*\perp}^{*T}K_{lin}K_{lin}^T\beta_{A^*}^* + \beta_{A^*\perp}^{*T}K_{lin}^TK_{lin}\beta^*_{A^*\perp} + M_1
\]
and \cref{lem:BEFS_1st_stage_trace_second_order_expansion}, for $M_2 = O\left((1-\alpha^2)(\frac{ \log n_B}{n_B})^{3/2}\right)$
\[
\text{Tr}(J_{\hat A}\Sigma_{test}J_{\hat A}) = \text{Tr}(\Sigma_{A^*}) + \alpha^2\text{Tr}(\Sigma_{A^*\perp}) + (1-\alpha^2)[\text{Tr}(K_{lin}^T\Sigma_{A^*\perp}K_{lin{}}) -\text{Tr}(K_{lin}\Sigma_{A^*}K_{lin}^T)] + M_2
\]
Combining terms, we can write, for $M_3 = O((\frac{\log n_B}{n_B})^{3/2}) + O((1-\alpha^2)(\frac{\log n_B}{n_B})^{3/2})$,
\[
\|\beta^*_{\hat A\perp}\|^2\text{Tr}(J_{\hat A}\Sigma_{test}J_{\hat A}) = 
\]
\[
(\beta^{*T}K_{lin}K_{lin}^T\beta^*  + \beta_{A^*\perp}^{*T}\beta^*_{A^*\perp}-2\beta^{*T}_{A^*\perp}K\beta^*_{A^*}-2 \beta_{A^*\perp}^{*T}K_{lin}K_{lin}^T\beta_{A^*}^* + \beta_{A^*\perp}^{*T}K_{lin}^TK_{lin}\beta^*_{A^*\perp})(\text{Tr}(\Sigma_{A^*}) + \alpha^2\text{Tr}(\Sigma_{A^*\perp}))
\]
\[
+ (1-\alpha^2)\|\beta_{A^*\perp}^{*T}\|^2 [\text{Tr}(K_{lin}^T\Sigma_{A^*\perp}K_{lin}) -\text{Tr}(K_{lin}\Sigma_{A^*}K_{lin}^T)] + M_3
\]
\end{lemma}
\begin{lemma}[Scaling of $\|\beta^*_{\hat A\perp}\|^2\text{Tr}(J_{\hat A}\Sigma_{test}J_{\hat A})$]\label{lem:BEFS_stage_1_mixed_scaling}
From \cref{lem:befs_1st_stage_mixed_expansion}, $M_1 = O((\frac{\log n_B}{n_B})^{3/2})$
\[
\|\beta^*_{\hat A\perp}\|^2\text{Tr}(J_{\hat A}\Sigma_{test}J_{\hat A}) = Z_{main} + M_1
\]
\[
\mathbb{E}[\|\beta^*_{\hat A\perp}\|^2\text{Tr}(J_{\hat A}\Sigma_{test}J_{\hat A})] = \mathbb{E}[Z_{main}] + \mathbb{E}[M_1\boldsymbol{1}_{\mathcal{E}}^{BEFS-1}] + \mathbb{E}[(\|\beta^*_{\hat A\perp}\|^2\text{Tr}(J_{\hat A}\Sigma_{test}J_{\hat A}) - Z_{main})\boldsymbol{1}_{\mathcal{E}^C}^{BEFS-1,}]
\]
\[
 \mathbb{E}[M_1\boldsymbol{1}_{\mathcal{E}}^{BEFS-1}] = O\left(\left(\frac{\log n_B}{n_B}\right)^{3/2}\right) + O\left((1-\alpha^2)\left(\frac{\log n_B}{n_B}\right)^{3/2}\right)
\]
\[
\mathbb{E}[(\|\beta^*_{\hat A\perp}\|^2\text{Tr}(J_{\hat A}\Sigma_{test}J_{\hat A}) - Z_{main})\boldsymbol{1}_{\mathcal{E}^C}^{BEFS-1,}] \leq m_1P(\mathcal{E}_C^{BEFS-1}) + \mathbb{E}[|Z_{main}|^{2}]^{1/2}P(\mathcal{E}_C^{BEFS-1})^{1/2} 
\]
\[
= o(n_B^{-1}) + o((1-\alpha^2)n_B^{-1})
\]
And using the intermediate results of the proofs in \cref{lem:BEFS_1st_Stage_trace_scaling_law} and \cref{lem:BEFS_1st_stage_beta_norm_scaling},
\[
\mathbb{E}[Z_{main}] = (\text{Tr}(\Sigma_{A^*}) + \alpha^2\text{Tr}(\Sigma_{A^*\perp}))(\beta_{A^*\perp}^{*T}\Sigma_{A^*\perp}\beta^*_{A^*\perp} + \frac{1}{n}[(\|\beta^*_{A^*\perp}\|^2\text{Tr}(\Sigma_{A^*}\Sigma_{est})
\]
\[
 - 2\beta_{A^*\perp}^{*T}\Sigma_{A^*\perp}\beta_{A^*\perp}\text{Tr}(\Sigma_{est}) + \beta^{*T}\Sigma_{est}\beta^{*}\text{Tr}(\Sigma_{A^*\perp})])
\]
\[
+  \frac{(1-\alpha^2)\|\beta^*_{A^*\perp}\|^2}{n_B}[\text{Tr}(\Sigma_{A^*\perp})\text{Tr}(\Sigma_{est})-\text{Tr}(\Sigma_{A^*}\Sigma_{est})(d_x - d_{\ell_{H^*}})] + o(n_B^{-1}) + o((1-\alpha^2)n_B^{-1})
\]
Therefore
\[
\mathbb{E}[\|\beta^*_{\hat A\perp}\|^2\text{Tr}(J_{\hat A}\Sigma_{test}J_{\hat A})] = (\text{Tr}(\Sigma_{A^*}) + \alpha^2\text{Tr}(\Sigma_{A^*\perp}))(\beta_{A^*\perp}^{*T}\Sigma_{A^*\perp}\beta^*_{A^*\perp} + \frac{1}{n_B}[(\|\beta^*_{A^*\perp}\|^2\text{Tr}(\Sigma_{A^*}\Sigma_{est})
\]
     \[
  - 2\beta_{A^*\perp}^{*T}\Sigma_{A^*\perp}\beta_{A^*\perp}\text{Tr}(\Sigma_{est}) + \beta^{*T}\Sigma_{est}\beta^{*}\text{Tr}(\Sigma_{A^*\perp})])\]
\[
+  \frac{(1-\alpha^2)\|\beta^*_{A^*\perp}\|^2}{n_B}[\text{Tr}(\Sigma_{A^*\perp})\text{Tr}(\Sigma_{est})-\text{Tr}(\Sigma_{A^*}\Sigma_{est})(d_x - d_{\ell_{H^*}})] + o(n_B^{-1}) + o((1-\alpha^2)n_B^{-1})
\]
And under isotropic test,
\[
\left(d_{{\ell_{H^*}}} + \alpha^2(d_x - d_{\ell_{H^*}})\right)(\|\beta_{A^*\perp}\|^2 + \frac{1}{n_B}\left[( \beta^{*T}\Sigma_{est}\beta^{*}(d_x - d_{\ell_{H^*}}) - \|\beta^*_{A^*\perp}\|^2\text{Tr}(\Sigma_{est})\right])+ o(n_B^{-1}) + o((1-\alpha^2)n_B^{-1})
\]
\end{lemma}
\subsection{BEFS - 2nd Stage}
Assume 
\[
\|\beta^*\| \leq C
\]
$n_T$ large enough such that:
\[
L_1(2)\sqrt{\frac{\log n_T}{n_T}}\leq 1/2
\]
Here we take the event $\mathcal{E}^{BEFS-2}$ to  be the following:
\[
\|\hat\Sigma - I\|_{op} \leq L_1(2)\sqrt{\frac{\log n_T}{n_T}}
\]
\[
\|\Delta_y\| \leq L_2(2)\sqrt{\frac{\log n_T}{n_T}}
\]
From \cref{lem:gaussian_bounds},
\[
P\left(\|\hat\Sigma - I\|_{op} \leq L_1(2)\sqrt{\frac{\log n_T}{n_T}}\right) \geq t_1(2)n_{B}^{-2}
\]
\[
P\left(\|\Delta_y\| \leq L_2(2)\sqrt{\frac{\log n_T}{n_T}}\right) \geq t_2(2)n_{B}^{-2}
\]
\begin{lemma}[BEFS Brain Encoding Second Stage Soft Constraint ]\label{lem:BEFS_2nd_stage_soft_beta} 
The optimization problem:
\[
\mathcal{L}(\beta) = \frac{1}{n_T}\|y^{(T)} - X^{(T)}\beta\|^2 + \lambda \|(I - P_{\hat A})\beta\|^2
\]
\[
\hat \beta^{BEFS} = \text{argmin}_{\beta}\mathcal{L}(\beta)
\]
Under event $\mathcal{E}^{BEFS, 2}$ has the following solution:
For $\alpha = \frac{1}{1+\lambda}, $$\beta^*_{\hat A_\perp} = (I - P_{\hat A})\beta^*$, $J_{\hat A} = P_{\hat A} + \alpha(I - P_{\hat A})$
\[
\hat \beta^{BEFS}=\beta^* - (1-\alpha)\beta^*_{\hat A\perp} + J_{\hat A}\Delta_y + (1-\alpha) J_{\hat A}(\hat \Sigma - I)\beta^*_{\hat A\perp}
\]
\[
- (1-\alpha)J_{\hat A}(\hat \Sigma - I)J_{\hat A}(\hat \Sigma - I)\beta^*_{\hat A\perp} +(1-\alpha)J_{\hat A}(\hat \Sigma - I)(I - P_{\hat A})\Delta_y + M
\]
\[
\|M\| \leq O\left(\lambda\left(\frac{\log n_T}{n_T}\right)^{3/2}\right)
\]
Note that the proof follows through with $P_{\hat A} = 0$ in which case it recovers ridge regression.\\
Proof:\\

Optimizing, the objective:
\[
\nabla_{\beta}\mathcal{L}(\beta) = \frac{2}{n_T}X^{(T)T}X^{(T)}\beta - \frac{2}{n_T}X^{(T)T}y^{(T)} + 2\lambda (I - P_{\hat A})\beta
\]
Setting equal to zero,
\[
\hat \beta^{BEFS} = \left(\frac{1}{n_T}X^{(T)T}X^{(T)} + \lambda(I - P_{\hat A})\right)^{-1}\frac{1}{n_T}X^{(T)T}y^{(T)}
\]
Substituting in $y^{(T)} = X^{(T)}\beta^* + e_y$
\[
\hat \beta^{BEFS} = \left(\frac{1}{n_T}X^{(T)T}X^{(T)} + \lambda(I - P_{\hat A})\right)^{-1}\frac{1}{n_T}X^{(T)T}(X^{(T)}\beta^* + e_y)
\]
Adding and subtracting $(\frac{1}{n_T}X^{(T)T}X^{(T)} + \lambda(I - P_{\hat A}))^{-1} \lambda(I - P_{\hat A})$,
\[
\hat \beta^{BEFS} = \beta^* + \left(\frac{1}{n_T}X^{(T)T}X^{(T)} + \lambda(I - P_{\hat A})\right)^{-1}\left(\frac{1}{n_T}X^Te_y -\lambda(I - P_{\hat A})\beta^*\right)
\]
\[
= \beta^* + \left(\hat \Sigma + \lambda(I - P_{\hat A})\right)^{-1}\left(\frac{1}{n_T}X^{(T)T}e_y -\lambda(I - P_{\hat A})\beta^*\right)
\]
\[
= \beta^* + \left(\hat \Sigma + \lambda(I - P_{\hat A})\right)^{-1}\left(\hat \Sigma\Delta_{y} -\lambda(I - P_{\hat A})\beta^*\right)
\]
Then since 
\[(\hat \Sigma + \lambda( I - P_{\hat A}))^{-1}\hat \Sigma = (\hat \Sigma + \lambda( I - P_{\hat A}))^{-1}\left((\hat \Sigma + \lambda( I - P_{\hat A})) - \lambda( I - P_{\hat A})\right) \]
\[
= I-\lambda(\hat \Sigma + \lambda( I - P_{\hat A}))^{-1}( I - P_{\hat A})
\]
We can simplify to:
\[
\hat \beta^{BEFS} = \beta^* + \Delta_{y} - \lambda(\hat \Sigma + \lambda( I - P_{\hat A}))^{-1}( I - P_{\hat A})(\beta^* + \Delta_{y})
\].
Now we take this to leading order by expanding $\hat \Sigma - I = G$
\[
\hat \beta^{BEFS} =\beta^* + \Delta_{y} - \lambda(I + \lambda( I - P_{\hat A}) + G)^{-1}( I - P_{\hat A})(\beta^* + \Delta_{y})
\]
$\hat \Sigma + \lambda(I - P_{\hat A})$ is invertible since $\|G\|_{op} \leq 1/2$, so $\hat \Sigma \succ 1/2$.

Define $J_{\hat A} = P_{\hat A} + \alpha(I - P_{\hat A})$ and $\alpha = \frac{1}{1+\lambda}$
and since
\[(I + \lambda\left(I - P_{\hat A})\right)^{-1} =  \left(P_{\hat A} + (1+\lambda)(I-P_{\hat A})\right)^{-1} = P_{\hat A} + \frac{1}{1+\lambda}(I - P_{\hat A}) = J_{\hat A}\]
\[
\left(I + \lambda(I - P_{\hat A}) + G\right)^{-1} = (I + J_{\hat A}G)^{-1}J_{\hat A}
\]
And the identity:
\[
(I+Z)^{-1}=I-Z+Z^2-Z^3(I+Z)^{-1}
\]
Then
\[
M_1 = - (J_{\hat A}G)^3(I + J_{\hat A}G)^{-1}J_{\hat A}
\]
\[
\left(I + \lambda(I - P_{\hat A}) + G\right)^{-1} = J_{\hat A} -  J_{\hat A}G J_{\hat A} +  J_{\hat A}G J_{\hat A}G J_{\hat A} + M_1
\]
Under $\mathcal{E}^{BEFS-2}$, since $\|J_{\hat A}\|_{op} =1$ and $G\leq 1/2$ and $G \leq \sqrt{\frac{\log n_T}{n_T}}$. Additionally using the Neumann series operator bound since $\|J_{\hat A}G\|_{op} \leq 1/2$, $\|(I + J_{\hat A}G)^{-1}\|_{op} \leq \frac{1}{1 - \|J_{\hat A}G\|_{op}} \leq 2$
\[
\|M_1\|_{op} \leq \|(J_{\hat A}G)^3\|_{op}\|(I + J_{\hat A}G)^{-1}\|_{op}\|J_{\hat A}\|_{op} \leq 2 \|G\|^{3}_{op} =O\left(\left(\frac{\log n_T}{n_T}\right)^{3/2}\right)
\]
So, 
\[
\hat \beta^{BEFS} = \beta^* + \Delta_y -\lambda(J_{\hat A} -  J_{\hat A}G J_{\hat A} +  J_{\hat A}G J_{\hat A}G J_{\hat A} + M_1)(I - P_{\hat A})(\beta^* + \Delta_y)
\]
Then, 
\[
\hat \beta^{BEFS} = \beta^* + \Delta_y -\lambda(J_{\hat A} -  J_{\hat A}G J_{\hat A} +  J_{\hat A}G J_{\hat A}G J_{\hat A} )(I - P_{\hat A})\beta^*
\]
\[
-\lambda(J_{\hat{A}} - J_{\hat A}G J_{\hat A})(I - P_{\hat A})\Delta_y + M_2
\] 
For 
\[
M_2 = -\lambda M_1(I - P_{\hat A})(\beta^* + \Delta_y) -\lambda J_{\hat A}GJ_{\hat A}GJ_{\hat A}(I - P_{\hat A})\Delta_y = -\lambda M_1(I - P_{\hat A})(\beta^* + \Delta_y) - (1-\alpha) J_{\hat A}GJ_{\hat A}G(I - P_{\hat A})\Delta_y
\]
\[
\|M_2\| \leq (1-\alpha)\|G\|^2_{op}\|\Delta_y\| + \lambda\|M_1\|_{op}(\|\beta^*\| + \|\Delta_y\|)  =O\left(\lambda\left(\frac{\log n_T}{n_T}\right)^{3/2}\right)
\]

And simplifying for $\beta^*_{\hat A_\perp} = (I - P_{\hat A})\beta^*$
\[
\hat \beta^{BEFS}=\beta^* - (1-\alpha)\beta^*_{\hat A\perp} + J_{\hat A}\Delta_y + (1-\alpha) J_{\hat A}(\hat \Sigma - I)\beta^*_{\hat A\perp}
\]
\[
- (1-\alpha)J_{\hat A}(\hat \Sigma - I)J_{\hat A}(\hat \Sigma - I)\beta^*_{\hat A\perp} +(1-\alpha)J_{\hat A}(\hat \Sigma - I)(I - P_{\hat A})\Delta_y + M_2
\]
\end{lemma}
\begin{lemma}[BEFS 2nd stage Scaling]\label{lem:BEFS_2nd_stage_scaling}
We want to find the scaling of \[
\mathbb{E}_{y_{test}, x_{test}, e_y,X^{(T)}}[ \|y_{test} - x_{test}^T\hat\beta^{BEFS}\|^2 |\hat{A}, e_y, X^{(T)}]  =\mathbb{E}_{X^{(T)}, y^{(T)}}[ (\hat\beta - \beta^*)^T\Sigma_{test}(\hat\beta - \beta^*) |\hat A]+ \sigma^2_{test}
\]
Suppose that under $\mathcal E^{BEFS-2}$, $\hat \beta^{BEFS} = \beta^* + v_{main} + M$ where $\|M\| = O\left(\lambda\left(\frac{\log n_T}{n_T}\right)^{3/2}\right)$ and $Z_{main} = O(\sqrt\frac{\log n_T}{ n_T})$
\[
\mathbb{E}_{X^{(T)}, y^{(T)}}[(\hat\beta^{BEFS} - \beta^*)^T\Sigma_{test}(\hat\beta^{BEFS} - \beta^*) \boldsymbol1_{\mathcal E}^{BEFS-2}| \hat A] + \mathbb{E}_{X^{(T)}, y^{(T)}}[(\hat\beta^{BEFS} - \beta^*)^T\Sigma_{test}(\hat\beta^{BEFS} - \beta^*) \boldsymbol1_{\mathcal E_c}^{BEFS-2}| \hat A] 
\]
Bounding the first term:
\[
\mathbb{E}_{X^{(T)}, y^{(T)}}[(\hat\beta^{BEFS} - \beta^*)^T\Sigma_{test}(\hat\beta^{BEFS} - \beta^*) \boldsymbol1_{\mathcal E}^{BEFS-2}| \hat A] = \mathbb{E}_{X^{(T)}, y^{(T)}}[(v_{main} +M)^T\Sigma_{test}(v_{main} +M) \boldsymbol1_{\mathcal E}^{BEFS-2}| \hat A]
\]
\[
=  \mathbb{E}_{X^{(T)}, y^{(T)}}[v_{main}^T\Sigma_{test}v_{main} \boldsymbol1_{\mathcal E}^{BEFS-2}| \hat A] +\mathbb{E}_{X^{(T)}, y^{(T)}}[M^T\Sigma_{test}v_{main} \boldsymbol1_{\mathcal E}^{BEFS-2}| \hat A]
\]
\[
+\mathbb{E}_{X^{(T)}, y^{(T)}}[v_{main}^T\Sigma_{test}M\boldsymbol1_{\mathcal E}^{BEFS-2}| \hat A]+\mathbb{E}_{X^{(T)}, y^{(T)}}[M^T\Sigma_{test}M\boldsymbol1_{\mathcal E}^{BEFS-2}| \hat A]
\]
Since $\|M\| = O\left(\lambda\left(\frac{\log n_T}{n_T}\right)^{3/2}\right)$ under the event,
\[
=  \mathbb{E}_{X^{(T)}, y^{(T)}}[v_{main}^T\Sigma_{test}v_{main} \boldsymbol1_{\mathcal E}^{BEFS-2}| \hat A] + o(\lambda n_T^{-1})
\]
Bounding the second term
\[
\mathbb{E}_{X^{(T)}, y^{(T)}}[(\hat\beta^{BEFS} - \beta^*)^T\Sigma_{test}(\hat\beta^{BEFS} - \beta^*) \boldsymbol1_{\mathcal E_c}^{BEFS-2}| \hat A] 
\]
Using the exact form of 
\[
\hat\beta^{BEFS} =\beta^*+\Delta_{y} - \lambda(\hat \Sigma + \lambda( I - P_{\hat A}))^{-1}( I - P_{\hat A})(\beta^* + \Delta_{y})
\]
\[
(\hat\beta^{BEFS} - \beta^*)^T\Sigma_{test}(\hat\beta^{BEFS} - \beta^*)
\le
2\|\Sigma_{test}\|_{op}\|\Delta_y\|^2
+
4\lambda^2\|\Sigma_{test}\|_{op}\|(\hat \Sigma + \lambda (I - P_{\hat A}))^{-1}\|_{op}^2\|\beta^*\|^2
\]
\[
+
4\lambda^2\|\Sigma_{test}\|_{op}\|(\hat \Sigma + \lambda (I - P_{\hat A}))^{-1}\|_{op}^2\|\Delta_y\|^2
\]
And $\|(\hat \Sigma + \lambda (I - P_{\hat A}))^{-1}\|_{op}^2\le \|\hat\Sigma^{-1}\|_{op}$. So,
\[
(\hat\beta^{BEFS} - \beta^*)^T\Sigma_{test}(\hat\beta^{BEFS} - \beta^*)\le
2\|\Sigma_{test}\|_{op}\|\Delta_y\|^2
+
4\lambda^2\|\Sigma_{test}\|_{op}\|\hat {\Sigma}^{-1}\|_{op}^2\|\beta^*\|^2+
4\lambda^2\|\Sigma_{test}\|_{op}\|\hat {\Sigma}^{-1}\|_{op}^2\|\Delta_y\|^2
\]
\[
\mathbb{E}_{X^{(T)}, y^{(T)}}[(\hat\beta^{BEFS} - \beta^*)^T\Sigma_{test}(\hat\beta^{BEFS} - \beta^*) \boldsymbol1_{\mathcal E_c}^{BEFS-2}| \hat A]  \leq
\]
Since by \cref{lem:gaussian_bounds},$\mathbb E[\|\Delta_y\|^4]=O(n_T^{-2})$ and $P(\mathcal E_c^{BEFS-2})^{1/2} = o(n_T^{-1})$
\[
\mathbb{E}_{X^{(T)}, y^{(T)}}[\|\Delta_y\|^2\boldsymbol1_{\mathcal E_c}^{BEFS-2}| \hat A] \leq \mathbb{E}_{X^{(T)}, y^{(T)}}[\|\Delta_y\|^4]^{1/2}P(\mathcal E_c^{BEFS-2})^{1/2} = o(n_T^{-2})
\]
and \cref{lem:gaussian_bounds} $\mathbb{E}[\|\hat \Sigma^{-1}\|_{op}] = O(1)$
\[
\mathbb{E}_{X^{(T)}, y^{(T)}}[\|\hat \Sigma^{-1}\|^2\boldsymbol1_{\mathcal E_c}^{BEFS-2}| \hat A]\leq\mathbb{E}_{X^{(T)}, y^{(T)}}[\|\hat \Sigma^{-1}\|_{op}^4]^{1/2}P(\mathcal E_c^{BEFS-2})^{1/2} = o(n_T^{-1})
\]
\[
\mathbb{E}_{X^{(T)}, y^{(T)}}[(\hat\beta^{BEFS} - \beta^*)^T\Sigma_{test}(\hat\beta^{BEFS} - \beta^*) \boldsymbol1_{\mathcal E_c}^{BEFS-2}| \hat A]  = o(\lambda^2n_T^{-1}) + o(n_T^{-2})
\]
\[
\mathbb{E}_{y_{test}, x_{test}, e_y,X^{(T)}}[ \|y_{test} - x_{test}^T\hat\beta^{BEFS}\|^2 |\hat{A}] =  \mathbb{E}_{X^{(T)}, y^{(T)}}[v_{main}^T\Sigma_{test}v_{main} \boldsymbol1_{\mathcal E}^{BEFS-2}| \hat A] + \sigma^2_{test}
\]
\[
+ o(\lambda n_T^{-1})+o(\lambda^2n_T^{-1}) + o(n_T^{-2})
\]
\end{lemma}
\begin{lemma}[BEFS Stage 2 Main Scaling]\label{lem:befs_stage_2_main_scaling}
\[
v_{main} = - (1-\alpha)\beta^*_{\hat A\perp} + J_{\hat A}\Delta_y + (1-\alpha) J_{\hat A}(\hat \Sigma - I)\beta^*_{\hat A\perp}
\]
\[
- (1-\alpha)J_{\hat A}(\hat \Sigma - I)J_{\hat A}(\hat \Sigma - I)\beta^*_{\hat A\perp} +(1-\alpha)J_{\hat A}(\hat \Sigma - I)(I - P_{\hat A})\Delta_y 
\]
Then for $m_1 = O\left(\lambda\left(\frac{\log n_T}{n_T}\right)^{-3/2}\right)$
\[
v_{main}^T\Sigma_{test}v_{main}
={}
(1-\alpha)^2\, \beta_{\hat A\perp}^{*T}\Sigma_{test}\beta^*_{\hat A\perp}
-2(1-\alpha)\,\beta_{\hat A\perp}^{*T}\Sigma_{test}J_{\hat A,\alpha}\Delta_y
\]
\[
-2(1-\alpha)^2\,\beta_{\hat A\perp}^{*T}\Sigma_{test}J_{\hat A,\alpha}(\hat\Sigma-I)\beta^*_{\hat A\perp}
+\Delta_y^T J_{\hat A,\alpha}^T\Sigma_{test}J_{\hat A,\alpha}\Delta_y+2(1-\alpha)\,\Delta_y^T J_{\hat A,\alpha}^T\Sigma_{test}J_{\hat A,\alpha}(\hat\Sigma-I)\beta^*_{\hat A\perp}
\]
\[
+(1-\alpha)^2\,\beta_{\hat A\perp}^{*T}(\hat\Sigma-I)^T
J_{\hat A,\alpha}^T\Sigma_{test}J_{\hat A,\alpha}
(\hat\Sigma-I)\beta^*_{\hat A\perp}
\]
\[
+2(1-\alpha)^2\,\beta_{\hat A\perp}^{*T}\Sigma_{test}
J_{\hat A,\alpha}(\hat\Sigma-I)J_{\hat A,\alpha}(\hat\Sigma-I)\beta^*_{\hat A\perp}
-2(1-\alpha)^2\,\beta_{\hat A\perp}^{*T}\Sigma_{test}
J_{\hat A,\alpha}(\hat\Sigma-I)(I-P_{\hat A})\Delta_y + m_1
\]
\[
= \omega + m_1
\]
Then 
\[
\mathbb{E}_{X^{(T)}, y^{(T)}}[v_{main}^T\Sigma_{test}v_{main}\boldsymbol1_{\mathcal E}^{BEFS-2}] = \mathbb{E}_{X^{(T)}, y^{(T)}}[w] - \mathbb{E}_{X^{(T)}, y^{(T)}}[w\boldsymbol1_{\mathcal E_c}^{BEFS-2}] +\mathbb{E}_{X^{(T)}, y^{(T)}}[m_1\boldsymbol1_{\mathcal E}^{BEFS-2}]
\]
Taking the conditional expectation on $e_{y}$ of the first term
\[
 \mathbb{E}_{e_y}[w|X^{(T)}]
\]
Dropping mean zero terms:
\[
= (1 - \alpha)^2\beta_{\hat A\perp}^{*T}\Sigma_{test}\beta^*_{ \hat A\perp} + \mathbb{E}_{e_{y}}[\Delta_{y}^TJ_{\hat A}\Sigma_{test}J_{\hat A}\Delta_{y}|X^{(T)}] + (1 - \alpha)^2\beta_{\hat A\perp}^{*T}(\hat \Sigma-I)J_{\hat A}\Sigma_{test}J_{\hat A}(\hat \Sigma-I)\beta^*_{ \hat A\perp}
\]
\[
- (1 - \alpha)^2\beta_{\hat A\perp}^{*T}\Sigma_{test}J_{\hat A}(\hat \Sigma-I)\beta^*_{\hat A\perp} - (1 - \alpha)^2\beta_{ \hat A\perp}^{*T}(\hat \Sigma-I)J_{\hat A}\Sigma_{test}\beta^*_{\hat A\perp}
\]
\[
+ (1-\alpha)^2\beta^{*T}_{\hat A\perp}\Sigma_{test}J_{\hat A}(\hat \Sigma - I)J_{\hat A}(\hat \Sigma - I)\beta^*_{\hat A\perp}+ (1-\alpha)^2\beta^{*T}_{\hat A\perp}(\hat \Sigma - I)J_{\hat A}(\hat \Sigma - I)J_{\hat A}\Sigma_{test}\beta^*_{\hat A\perp}
\]

Then taking expectation over $X^{(T)}$ drops the $(\hat \Sigma - I)$ linear terms since they are mean 0.
\[
 \mathbb{E}_{e_y, X^{(T)}}[w]
\]
\[
= (1 - \alpha)^2\beta_{\hat A\perp}^{*T}\Sigma_{test}\beta^*_{ \hat A\perp} + \mathbb{E}_{X^{(T)}}[\mathbb{E}_{e_{y}}[\Delta_{y}^TJ_{\hat A}\Sigma_{test}J_{\hat A}\Delta_{y}|X^{(T)}]] + (1 - \alpha)^2\mathbb{E}_{X^{(T)}}[\beta_{\hat A\perp}^{*T}(\hat \Sigma-I)J_{\hat A}\Sigma_{test}J_{\hat A}(\hat \Sigma-I)\beta^*_{ \hat A\perp}]
\]
\[
+ (1-\alpha)^2\mathbb{E}_{X^{(T)}}[\beta^{*T}_{\hat A\perp}\Sigma_{test}J_{\hat A}(\hat \Sigma - I)J_{\hat A}(\hat \Sigma - I)\beta^*_{\hat A\perp}]+ (1-\alpha)^2\mathbb{E}_{X^{(T)}}[\beta^{*T}_{\hat A\perp}(\hat \Sigma - I)J_{\hat A}(\hat \Sigma - I)J_{\hat A}\Sigma_{test}\beta^*_{\hat A\perp}] + M_3
\]
Since $\Delta_y | X^{(T)} \sim N(0, \frac{\sigma_{y}^2}{n_T}\hat \Sigma^{-1})$, we can simplify the conditional expectation to:
\[
\mathbb{E}_{e_{y}}[\Delta_{y}^TJ_{\hat A}\Sigma_{test}J_{\hat A}\Delta_{y}|X^{(T)}] = \frac{\sigma_{y}^2}{n_T}\text{Tr}(J_{\hat A}\Sigma_{test}J_{\hat A}\hat \Sigma^{-1})
\]
Finally, taking the expectation on $X^{(T)}$, using $E_{X^{(T)}}[\hat\Sigma^{-1}] = \frac{n_T}{n_T-d_x-1}\,I$, 
\[
\mathbb{E}_{X^{(T)}}[\mathbb{E}_{e_{y}}[\Delta_{y}^TJ_{\hat A}\Sigma_{test}J_{\hat A}\Delta_{y}|X^{(T)}]] = \frac{\sigma_{y}^2}{n_T - d_x -1}\text{Tr}(J_{\hat A}\Sigma_{test}J_{\hat A})
\]
Next, taking the expectations on the covariance error pieces using \cref{lem:basics},
\[
(1 - \alpha)^2\mathbb{E}_{X^{(T)}}[\beta_{\hat A\perp}^{*T}(\hat \Sigma-I)J_{\hat A}\Sigma_{test}J_{\hat A}(\hat \Sigma-I)\beta^*_{ \hat A\perp}] 
\]
\[
=  \frac{(1-\alpha)^2}{n_T}(\|\beta^*_{\hat A\perp}\|^2\text{Tr}(J_{\hat A}\Sigma_{test}J_{\hat A})+\beta^{*T}_{\hat A\perp}J_{\hat A}\Sigma_{test}J_{\hat A}\beta^{*}_{\hat A\perp})
\]
\[
= \frac{(1-\alpha)^2}{n_T}(\|\beta^*_{\hat A\perp}\|^2\text{Tr}(J_{\hat A}\Sigma_{test}J_{\hat A})+\alpha^2\beta^{*T}_{\hat A\perp}\Sigma_{test}\beta^{*}_{\hat A\perp})
\]

\[
(1-\alpha)^2\mathbb{E}_{X^{(T)}}[\beta^{*T}_{\hat A\perp}\Sigma_{test}J_{\hat A}(\hat \Sigma - I)J_{\hat A}(\hat \Sigma - I)\beta^*_{\hat A\perp}]
\]
\[
= \frac{(1-\alpha)^2}{n_T}(\text{Tr}(J_{\hat A})\beta^{*T}_{\hat A\perp}\Sigma_{test}J_{\hat A}\beta^{*}_{\hat A\perp} + \beta^{*T}_{\hat A\perp}\Sigma_{test}J^2_{\hat A, \alpha}\beta^{*}_{\hat A\perp})
\]
\[
= \frac{(1-\alpha)^2}{n_T}\left(\alpha\text{Tr}(J_{\hat A})+\alpha^2\right)\beta^{*T}_{\hat A\perp}\Sigma_{test}\beta^{*}_{\hat A\perp}
\]
\[
(1-\alpha)^2\mathbb{E}_{X^{(T)}}[\beta^{*T}_{\hat A\perp}(\hat \Sigma - I)J_{\hat A}(\hat \Sigma - I)J_{\hat A}\Sigma_{test}\beta^*_{\hat A\perp}]
\]
\[
= \frac{(1-\alpha)^2}{n_T}\left(\alpha\text{Tr}(J_{\hat A})+\alpha^2\right)\beta^{*T}_{\hat A\perp}\Sigma_{test}\beta^{*}_{\hat A\perp}
\]
Which gives
\[
 \mathbb{E}_{e_y, X^{(T)}}[w] = 
\]
\[
\sigma_{test}^2 + (1-\alpha)^2\beta^{*T}_{\hat A\perp}\Sigma_{test}\beta^{*}_{\hat A\perp}\left(1+ \frac{2\alpha\text{Tr}(J_{\hat A}) + 3\alpha^2}{n_T}\right) + \left(\frac{\sigma_y^2}{n_T - d_x -1} +\frac{(1-\alpha)^2\|\beta^*_{\hat A\perp}\|^2}{n_T}\right)\text{Tr}(J_{\hat A}\Sigma_{test}J_{\hat A})
\]
Next, working on
\[
\mathbb{E}_{X^{(T)}, y^{(T)}}[\omega\boldsymbol1_{\mathcal E_c}^{BEFS-2}] \leq \mathbb{E}_{X^{(T)}, y^{(T)}}[\omega^2]^{1/2}P(\mathcal E^{BEFS-2}_{c})^{1/2}\]
Note that each term of $\omega$ carries a $(1-\alpha)$ or $(1-\alpha)^2$ except for $\Delta_y^TJ_{\hat A}^T\Sigma_{test}J_{\hat A}\Delta_y$. So squared $\omega$ carries $(1-\alpha)$, $(1-\alpha)^2$,$(1-\alpha)^3$ and $(1-\alpha)^4$ terms. So,
\[
\mathbb{E}_{X^{(T)}, y^{(T)}}[\omega\boldsymbol1_{\mathcal E_c}^{BEFS-2}]  \leq \mathbb{E}_{X^{(T)}, y^{(T)}}[(\Delta_y^TJ_{\hat A}^T\Sigma_{test}J_{\hat A}\Delta_y)^2]^{1/2}P(\mathcal E^{BEFS-2}_{c})^{1/2} + m_1
\]
Where $m_1 = o((1-\alpha)n_T^{-1})+ o((1-\alpha)^2n_T^{-1})+o((1-\alpha)^3n_T^{-1})+ o((1-\alpha)^4n_T^{-1})$
Since $\mathbb{E}[\|\Delta_y\|^4] = O(n_T^{-2})$,
\[
\mathbb{E}_{X^{(T)}, y^{(T)}}[\omega\boldsymbol1_{\mathcal E_c}^{BEFS-2}] = o(n_T^{-2}) + o((1-\alpha)n_T^{-1})+ o((1-\alpha)^2n_T^{-1})+o((1-\alpha)^3n_T^{-1})+ o((1-\alpha)^4n_T^{-1})
\]
Finally, since $(1-\alpha) = \frac{\lambda}{1+\lambda} \leq \lambda$
\[
\mathbb{E}_{X^{(T)}, y^{(T)}}[\omega\boldsymbol1_{\mathcal E_c}^{BEFS-2}] = o(n_T^{-2}) + o(\lambda n_T^{-1})+ o(\lambda ^2n_T^{-1})+o(\lambda^3n_T^{-1})+ o(\lambda^4n_T^{-1})
\]
Finally, 
\[
\mathbb{E}_{X^{(T)}, y^{(T)}}[m_1\boldsymbol1_{\mathcal E}^{BEFS-2}] = O\left(\lambda\left(\frac{\log n_T}{n_T}\right)^{3/2}\right) = o(\lambda n_T^{-1})
\]
Therefore
\[
\mathbb{E}_{X^{(T)}, R^{(T)}}[v_{main}^T\Sigma_{test}v_{main}] = 
\]
\[
\sigma_{test}^2 + (1-\alpha)^2\beta^{*T}_{\hat A\perp}\Sigma_{test}\beta^{*}_{\hat A\perp}\left(1+ \frac{2\alpha\text{Tr}(J_{\hat A}) + 3\alpha^2}{n_T}\right) + \left(\frac{\sigma_y^2}{n_T - d_x -1} +\frac{(1-\alpha)^2\|\beta^*_{\hat A\perp}\|^2}{n_T}\right)\text{Tr}(J_{\hat A}\Sigma_{test}J_{\hat A})
\]
\[
+ o(\lambda n_T^{-1}) + o(\lambda^2 n_T^{-1}) + o(\lambda^3 n_T^{-1}) + o(\lambda^4 n_T^{-1}) + o(n_T^{-3})
\]
\end{lemma}
\begin{theorem}[BEFS Stage 2 Scaling with remainder bound ]\label{thm:befs_stage_2_scaling}
Combining \cref{lem:BEFS_2nd_stage_scaling} and \cref{lem:befs_stage_2_main_scaling} gives \[
\mathbb{E}_{y_{test}, x_{test}, e_y,X^{(T)}}[ \|y_{test} - x_{test}^T\hat\beta^{BEFS}\|^2 |\hat{A}] = 
\]
\[
\sigma_{test}^2 + (1-\alpha)^2\beta^{*T}_{\hat A\perp}\Sigma_{test}\beta^{*}_{\hat A\perp}\left(1+ \frac{2\alpha\text{Tr}(J_{\hat A}) + 3\alpha^2}{n_T}\right) + \left(\frac{\sigma_y^2}{n_T - d_x -1} +\frac{(1-\alpha)^2\|\beta^*_{\hat A\perp}\|^2}{n_T}\right)\text{Tr}(J_{\hat A}\Sigma_{test}J_{\hat A})
\]
\[
+ o(n_T^{-2}) + o(\lambda n_T^{-1})+ o(\lambda ^2n_T^{-1})+o(\lambda^3n_T^{-1})+ o(\lambda^4n^{-1})
\]
\end{theorem}
\begin{theorem}[Total Scaling Law]\label{thm:total_scaling}
From \cref{thm:befs_stage_2_scaling} we have 
 \[
\mathbb{E}_{y_{test}, x_{test}, e_y,X^{(T)}}[ \|y_{test} - x_{test}^T\hat\beta^{BEFS}\|^2 |\hat{A}] = 
\]
\[
\sigma_{test}^2 + (1-\alpha)^2\beta^{*T}_{\hat A\perp}\Sigma_{test}\beta^{*}_{\hat A\perp}\left(1+ \frac{2\alpha\text{Tr}(J_{\hat A}) + 3\alpha^2}{n_T}\right) + \left(\frac{\sigma_y^2}{n_T - d_x -1} +\frac{(1-\alpha)^2\|\beta^*_{\hat A\perp}\|^2}{n_T}\right)\text{Tr}(J_{\hat A}\Sigma_{test}J_{\hat A})
\]
\[
+ o(n_T^{-2}) + o(\lambda n_T^{-1})+ o(\lambda ^2n_T^{-1})+o(\lambda^3n_T^{-1})+ o(\lambda^4n_T^{-1})
\]
Taking the expectation on $\hat A$, (note that $\text{Tr}(J_{\hat A})$ is constant),
\[
\mathbb{E}_{y_{test}, x_{test}, X^{(B)}, R^{(B)},e_y,X^{(T)}}[ \|y_{test} - x_{test}^T\hat\beta^{BEFS}\|^2]
\]
\[\sigma_{test}^2 + (1-\alpha)^2\mathbb{E}_{X^{(B)}, R^{(B)}}[\beta^{*T}_{\hat A\perp}\Sigma_{test}\beta^{*}_{\hat A\perp}]\left(1+ \frac{2\alpha\text{Tr}(J_{\hat A}) + 3\alpha^2}{n_T}\right)
\]
\[
+
\frac{\sigma_y^2}{n_T-d_x-1}
\mathbb{E}\!\left[\mathrm{Tr}(J_{\hat A,\alpha}\Sigma_{test}J_{\hat A,\alpha})\right]
+
\frac{(1-\alpha)^2}{n_T}
\mathbb{E}\!\left[\|\beta^*_{\hat A\perp}\|^2\,
\mathrm{Tr}(J_{\hat A,\alpha}\Sigma_{test}J_{\hat A,\alpha})\right]
\]
\[
+ o(n_T^{-2}) + o(\lambda n_T^{-1})+ o(\lambda ^2n_T^{-1})+o(\lambda^3n_T^{-1})+ o(\lambda^4n_T^{-1})
\]
Plugging in \cref{lem:BEFS_1st_stage_beta_norm_scaling}, \cref{lem:befs_1st_stage_mixed_expansion}, \cref{lem:BEFS_1st_Stage_trace_scaling_law},
\[
\mathbb{E}_{X^{(B)}, R^{(B)}}[\|
\beta_{\hat A\perp}^{*}\|^2]\approx \gamma_{I}(n_B) = \|\beta_{A^*\perp}\|^2 + \frac{1}{n_B}\left[( \beta^{*T}\Sigma_{est}\beta^{*}(d_x - d_{\ell_{H^*}}) - \|\beta^*_{A^*\perp}\|^2\text{Tr}(\Sigma_{est})\right]
\]
and
\[
\mathbb{E}_{X^{(B)}, R^{(B)}}[
\beta_{\hat A\perp}^{*T}\Sigma_{test}\beta^*_{\hat A\perp}]\approx \gamma_{\Sigma_{test}}(n_B)
  = \beta_{A^*\perp}^{*T}\Sigma_{A^*\perp}\beta^*_{A^*\perp} 
\]
\[
+ \frac{1}{n_B }\left[(\|\beta^*_{A^*\perp}\|^2\text{Tr}(\Sigma_{A^*}\Sigma_{est}) - 2\beta_{A^*\perp}^{*T}\Sigma_{A^*\perp}\beta_{A^*\perp}\text{Tr}(\Sigma_{est}) + \beta^{*T}\Sigma_{est}\beta^{*}\text{Tr}(\Sigma_{A^*\perp})\right]
\]

\[
\varepsilon_{\Sigma_{test}}^{BEFS, Soft}(n_B,n_T)
=
\sigma^2_{test}
+
(1-\alpha)^2\gamma_{\Sigma_{test}}(n_B)\left(1+ \frac{2\alpha\left(d_{\ell_{H^*}} + \alpha(d_x - d_{\ell_{H^*}})\right) + 3\alpha^2}{n_T}\right)
\]
\[
+
\frac{\sigma_y^2}{n_T-d_x-1}
\left(
\mathrm{Tr}(\Sigma_{A^*}) + \alpha^2\mathrm{Tr}(\Sigma_{A^*\perp})
+
\frac{1-\alpha^2}{n_B}
\left[
\mathrm{Tr}(\Sigma_{A^*\perp})\mathrm{Tr}(\Sigma_{est})
-
\mathrm{Tr}(\Sigma_{A^*}\Sigma_{est})(d_x-d_{\ell_{H^*}})
\right]
\right) 
\]
\[
+
\frac{(1-\alpha)^2\gamma_I(n_B)}{n_T}
\left(\mathrm{Tr}(\Sigma_{A^*}) + \alpha^2\mathrm{Tr}(\Sigma_{A^*\perp})\right)
\]
\[
+
\frac{(1-\alpha^2)(1-\alpha)^2\|\beta^*_{A^*\perp}\|^2}{n_Bn_T}
\left[
\mathrm{Tr}(\Sigma_{A^*\perp})\mathrm{Tr}(\Sigma_{est})
-
\mathrm{Tr}(\Sigma_{A^*}\Sigma_{est})(d_x-d_{\ell_{H^*}})
\right]
\]
\[
+
o\!\left(\frac{\lambda^2}{n_B}\right)
+
o\!\left(\frac{1-\alpha^2}{n_Bn_T}\right)
+
o\!\left(\frac{\lambda^2}{n_Bn_T}\right)
+
o(n_T^{-2})
+
o(\lambda n_T^{-1})
+
o(\lambda^2 n_T^{-1})
+
o(\lambda^3 n_T^{-1})
+
o(\lambda^4 n_T^{-1}).
\]
And in the isotropic test case
\[
\varepsilon^{BEFS, Soft}_{I}(n_B,n_T)
=
\sigma_{test}^2
+
(1-\alpha)^2\gamma_I(n_B)\left(1+ \frac{2\alpha\left(d_{\ell_{H^*}} + \alpha(d_x - d_{\ell_{H^*}})\right) + 3\alpha^2}{n_T}\right)
\]
\[
+
\left(
\frac{\sigma_y^2}{n_T-d_x-1}
+
\frac{(1-\alpha)^2\gamma_I(n_B)}{n_T}
\right)
\left(d_{\ell_{H^*}} + \alpha^2(d_x-d_{\ell_{H^*}})\right)
\]
\[
+
o\!\left(\frac{\lambda^2}{n_B}\right)
+
o\!\left(\frac{1-\alpha^2}{n_Bn_T}\right)
+
o\!\left(\frac{\lambda^2}{n_Bn_T}\right)
+
o(n_T^{-2})
+
o(\lambda n_T^{-1})
+
o(\lambda^2 n_T^{-1})
+
o(\lambda^3 n_T^{-1})
+
o(\lambda^4 n_T^{-1}).
\]
Note that if we only care about the remainder up to $o(n_T^{-1})+o(n_B^{-1}) + o(n_T^{-1}n_B^{-1})$, then we can drop the $d_x -1$ in the denominator.
\end{theorem}
\subsection{Optimal BEFS $\lambda$ Schedule}
\begin{theorem}[Asymptotically Optimal $\lambda$ Schedule]\label{thm:optimal_lambda}
Here we assume that $\beta^*_{A^*\perp} \neq 0$ so $\gamma_{I}(n_B) \neq 0$ and we take $n_B$ to be fixed. $\mathbb{E}_{X^{(B)}, R^{(B)}}\left[\|\beta^*_{\hat A\perp}\|^2\right]$ is clearly bounded since $\beta^*$ is bounded and $\hat A$ is orthornormal. Then from the exact form under $n_B$ of BEFS stage 2,
\[
\varepsilon^{BEFS}_{I}(n_T, n_B)
=
\sigma_{test}^2
+
(1-\alpha)^2\mathbb{E}_{X^{(B)}, R^{(B)}}\left[\|\beta^*_{\hat A\perp}\|^2\right]\left(1+ \frac{2\alpha\left(d_{\ell_{H^*}} + \alpha(d_x - d_{\ell_{H^*}})\right) + 3\alpha^2}{n_T}\right)
\]
\[
+
\left(
\frac{\sigma_y^2}{n_T-d_x-1}
+
\frac{(1-\alpha)^2\mathbb{E}_{X^{(B)}, R^{(B)}}\left[\|\beta^*_{\hat A\perp}\|^2\right]}{n_T}
\right)
\left(d_{\ell_{H^*}} + \alpha^2(d_x-d_{\ell_{H^*}})\right)
\]
\[
+ o(n_T^{-2}) + o(\lambda n_T^{-1})+ o(\lambda ^2n_T^{-1})+o(\lambda^3n_T^{-1})+ o(\lambda^4n_T^{-1})
\]
Then clearly $(1-\alpha)^2\mathbb{E}_{X^{(B)}, R^{(B)}}\left[\|\beta^*_{\hat A\perp}\|^2\right]$ must go to zero in $n_T$. Otherwise there is a higher risk floor than $\sigma^2_{test}$.
\[
\alpha=1-\lambda+\lambda^2+O(\lambda^3),\quad
\alpha^2=1-2\lambda+3\lambda^2+O(\lambda^3),\quad
1-\alpha=\lambda-\lambda^2+O(\lambda^3),
\]
So in the small $\lambda < 1$ regime,
\[
\varepsilon^{BEFS}_{I}(n_B,n_T,\lambda)
=
\sigma_{test}^2
+
\frac{\sigma_y^2 d_x}{n_T-d_x-1}
+
\mathbb{E}_{X^{(B)}, R^{(B)}}\left[\|\beta^*_{\hat A\perp}\|^2\right]\lambda^2
-
\frac{2\sigma_y^2 (d_x - d_{\ell_{H^*}})}{n_T}\lambda
\]
\[
+
O\!\left(\frac{\lambda^2}{n_T}\right)
+
O(\lambda^3)
+
o\!\left(\frac{\lambda^2}{n_B}\right)
+
o\!\left(\frac{\lambda}{n_Bn_T}\right)
+ o(n_T^{-2}) + o(\lambda n_T^{-1})
\]
To balance the leading terms, it must be $\lambda_{opt}(n_T) = \frac{c}{n_T} + o(n_T^{-1})$
\[
\varepsilon^{BEFS}_{I}(n_B,n_T,\lambda(n_B,n_T,c))
=
\sigma_{test}^2
+
\frac{\sigma_y^2 d_x}{n_T -d_x-1}
+
\frac{1}{n_T^2}[\mathbb{E}_{X^{(B)}, R^{(B)}}\left[\|\beta^*_{\hat A\perp}\|^2\right]c^2
-
2\sigma_y^2 (d_x - d_{\ell_{H^*}})c]
\]
\[
+ o(n_T^{-2}) + o(n_T^{-2}n_B^{-1})
\]
Optimizing  over $c$ gives 
\[
c_{min} = 
\frac{\sigma_y^2(d_x-d_{\ell_{H^*}})}{\mathbb{E}_{X^{(B)}, R^{(B)}}\left[\|\beta^*_{\hat A\perp}\|^2\right]}
\]
So the risk becomes:
\[
\varepsilon^{BEFS}_{I}(n_B,n_T,\lambda_{opt})
=
\sigma_{test}^2
+
\frac{\sigma_y^2 d_x}{n_T - d_x-1}
-
\frac{\sigma_y^4(d_x-d_{\ell_{H^*}})^2}{\gamma_I(n_B) + o(n^{-1}_B)}\frac{1}{n_T^2}
+
o(n_T^{-2})
\]
\end{theorem}
\begin{lemma}[Value Function]\label{lem:value}
Suppose we have 
\[
\varepsilon^{BEFS}_{\Sigma_{test}}(n_B, n_T, \lambda^{opt}(n_T)) = \varepsilon^{TOS}_{\Sigma_{test}}(n_T) - \sigma_y^2\text{Tr}(\Sigma_{test})\frac{s(n_B)}{n_{T}^2} + o(n_T^{-2})
\]
Define the value function
\[
\varepsilon^{TOS}_{\Sigma_{test}}(n_T + V_{\Sigma_{test}}(n_T, n_B)) = \varepsilon^{BEFS}_{\Sigma_{test}}(n_B, n_T, \lambda^{opt}(n_T))
\]
We show that 
\[
V(n_B, n_T) = s(n_B) + o_{n_T}(1)
\]
Proof:
\[
\varepsilon^{BEFS}_{\Sigma_{test}}(n_B, n_T, \lambda^{opt}(n_T)) = \varepsilon^{TOS}_{\Sigma_{test}}(n_T) - \sigma_y^2\text{Tr}(\Sigma_{test})\frac{s(n_B)}{n_{T}^2} + o(n_T^{-2})
\]
Then,
\[
\varepsilon^{TOS}_{\Sigma_{test}}(n_T + V_{\Sigma_{test}}(n_T, n_B)) = \varepsilon^{TOS}_{\Sigma_{test}}(n_T) - \sigma_y^2\text{Tr}(\Sigma_{test})\frac{s(n_B)}{n_{T}^2} + o(n_T^{-2})
\]
Canceling the constants on each side,
\[
\frac{1}{V(n_B, n_T)+n_T - d_x-1} = \frac{1}{n_T - d_x-1} - \frac{s(n_B)}{n_T^2} + o(n_T^{-2})
\]
Expanding the wishart denominator
\[
\frac{1}{V(n_B, n_T)+n_T - d_x-1} = \frac{1}{n_T} + \frac{-s(n_B)+d_x+1}{n_T^2} + o(n_T^{-2})
\]
\[
V(n_B, n_T) = n_{T}\left(1+\frac{-s(n_B)+d_x+1}{n_T} + o(n_T^{-1})\right)^{-1} +d_x+1
\]
\[
V(n_B, n_T) = n_{T}\left(1+\frac{s(n_B)-d_x-1}{n_T} + o(n_T^{-1})\right) -n+d_x+1 = s(n_B) + o_{n_T}(1)
\]
\end{lemma}
\begin{theorem}[BEFS brain samples to TOS task samples exchange rate]\label{thm:exchange_rate}
    \[
\varepsilon^{BEFS}_{I}(n_B,n_T,\lambda^{opt}(n_T)) = \varepsilon^{TOS}_{I} - \frac{\sigma_y^4(d_x - d_{\ell_{H^*}})^2}{\gamma_{I}(n_B) + o(n^{-1}_{B})}\frac{1}{n_T^2} + o(n_T^{-2})
\]
Applying \cref{lem:value},
\[
V_{I}(n_B, n_T) = \sigma_y^2\frac{(d_x-d_{{\ell_{H^*}}})^2}{d_x}\frac{1}{
\gamma_{I}(n_B) +o(n_{B}^{-1})}+ o_{n_T}(1)
\]
Which we call the value of brain data. We can also write the value as an exchange rate $V_{I}(n_B) = \rho_{I}(n_B)n_{B}$
\[
\rho_{I}(n_B, n_T) = \sigma_y^2\frac{(d_x-d_{{\ell_{H^*}}})^2}{d_x}\frac{1}{
n_Bm^2
+
\Big(
(d_x - d_{\ell_{H^*}})\beta^{*T}\Sigma_{est}\beta^*
-
m^2 \mathrm{Tr}(\Sigma_{est})\Big ) + o_{n_B}(1)} + o_{n_T}(1/n_B)
\]
So the value increases with brain data, but the exchange rate decreases.
\end{theorem}
\subsection{Robustness}
\begin{theorem}[Robustness under $\lambda_{opt}$]\label{thm:robustness_exact}
\[
\varepsilon^{BEFS}_{\Sigma_{test}}(n_T, n_B) = \sigma_{test}^2 + (1-\alpha)^2\mathbb{E}_{X^{(B)}, R^{(B)}}[\beta^{*T}_{\hat A\perp}\Sigma_{test}\beta^{*}_{\hat A\perp}]\left(1+ \frac{2\alpha(d_{{\ell_{H^*}}} + \alpha(d_x - d_{\ell_{H^*}})) + 3\alpha^2}{n_T}\right)
\]
\[
 + \frac{\sigma_y^2\mathbb{E}_{X^{(B)}, R^{(B)}}[\text{Tr}(J_{\hat A}\Sigma_{test}J_{\hat A})]}{n_T - d_x -1}+\frac{(1-\alpha)^2\mathbb{E}\left[\|\beta^*_{\hat A\perp}\|^2\text{Tr}(J_{\hat A}\Sigma_{test}J_{\hat A})\right]}{n_T}
\]
\[
+ o(\lambda n_T^{-1}) + o(\lambda^2 n_T^{-1}) + o(\lambda^3 n_T^{-1}) + o(\lambda^4 n_T^{-1}) + o(n_T^{-3})
\]
Using $\lambda_{opt}(n_B,n_T) = \frac{1}{n_T}\frac{\sigma_y^2(d_x-d_{\ell_{H^*}})}{\mathbb{E}_{X^{(B)}, R^{(B)}}[\|\beta^*_{A^*\perp^2}\|^2])} + o(n_T^{-1})$, plugging this schedule into the risk:
\[
\alpha=1-\lambda+\lambda^2+O(\lambda^3),\quad
\alpha^2=1-2\lambda+3\lambda^2+O(\lambda^3),\quad
1-\alpha=\lambda-\lambda^2+O(\lambda^3),
\]
\[
\varepsilon_{\Sigma_{test}}^{BEFS}(n_B,n_T,\lambda_{opt})
=
\varepsilon_{\Sigma_{test}}^{TOS}(n_T)
\]
\[
+
\frac{\sigma_y^4(d_x-d_{\ell_{H^*}})^2}{n_T^2}
\frac{
\mathbb E_{X^{(B)},R^{(B)}}
\left[
\beta_{\hat A\perp}^{*T}
\Sigma_{test}
\beta_{\hat A\perp}^*
\right]
}
{
\left(
\mathbb E_{X^{(B)},R^{(B)}}
[
\|\beta^*_{\hat A\perp}\|^2
]
\right)^2
}
\]
\[
-
\frac{2\sigma_y^4(d_x-d_{\ell_{H^*}})}{n_T^2}
\frac{
\mathbb E_{X^{(B)},R^{(B)}}
\left[
\mathrm{Tr}(P_{\hat A\perp}\Sigma_{test})
\right]
}
{
\mathbb E_{X^{(B)},R^{(B)}}
[
\|\beta^*_{\hat A\perp}\|^2
]
}
+
o(n_T^{-2})
\]
Writing into a form such that the sign condition is clear:
\[
\varepsilon_{\Sigma_{test}}^{BEFS}(n_B,n_T,\lambda_{opt})
=
\varepsilon_{\Sigma_{test}}^{TOS}(n_T)
\]
\[-\frac{\sigma_y^4(d_x-d_{\ell_{H^*}})^2}
{
n_T^2
\mathbb E_{X^{(B)},R^{(B)}}
\left[
\|\beta^*_{\hat A\perp}\|^2
\right]
}
\left[
2
\frac{
\mathbb E_{X^{(B)},R^{(B)}}
\left[
\mathrm{Tr}(P_{\hat A\perp}\Sigma_{test})
\right]
}
{
d_x-d_{\ell_{H^*}}
}
-\frac{
\mathbb E_{X^{(B)},R^{(B)}}
\left[
\beta_{\hat A\perp}^{*T}
\Sigma_{test}
\beta_{\hat A\perp}^*
\right]
}
{
\mathbb E_{X^{(B)},R^{(B)}}
\left[
\|\beta^*_{\hat A\perp}\|^2
\right]
}
\right]
+
o(n_T^{-2})
\]
So there is a net scaling improvement when the test distribution mass on the "missed" $\beta^*$ direction doesn't have exceptionally large mass (twice the size) compared to the average covariance mass in the null space of the learned encoding model features. 
\[
\text{negative sign when:} \quad
\frac{
\mathbb E_{X^{(B)},R^{(B)}}
\left[
\beta_{\hat A\perp}^{*T}
\Sigma_{test}
\beta_{\hat A\perp}^*
\right]
}
{
\mathbb E_{X^{(B)},R^{(B)}}
\left[
\|\beta^*_{\hat A\perp}\|^2
\right]
}
<2
\frac{
\mathbb E_{X^{(B)},R^{(B)}}
\left[
\mathrm{Tr}(P_{\hat A\perp}\Sigma_{test})
\right]
}
{
d_x-d_{\ell_{H^*}}
}
\]
\end{theorem}
\begin{lemma}[Value of brain data under test distribution shift]\label{lem:robustness_value}
From \cref{thm:robustness_exact},
\[
\varepsilon_{\Sigma_{test}}^{BEFS}(n_B,n_T,\lambda_{opt})
=
\varepsilon_{\Sigma_{test}}^{TOS}(n_T)
\]
\[+\frac{\sigma_y^4(d_x-d_{\ell_{H^*}})^2}
{
n_T^2
\mathbb E_{X^{(B)},R^{(B)}}
\left[
\|\beta^*_{\hat A\perp}\|^2
\right]
}
\left[
\frac{
\mathbb E_{X^{(B)},R^{(B)}}
\left[
\beta_{\hat A\perp}^{*T}
\Sigma_{test}
\beta_{\hat A\perp}^*
\right]
}
{
\mathbb E_{X^{(B)},R^{(B)}}
\left[
\|\beta^*_{\hat A\perp}\|^2
\right]
}
-
2
\frac{
\mathbb E_{X^{(B)},R^{(B)}}
\left[
\mathrm{Tr}(P_{\hat A\perp}\Sigma_{test})
\right]
}
{
d_x-d_{\ell_{H^*}}
}
\right]
+
o(n_T^{-2})
\]
Then by \cref{lem:value},
\[
V_{\Sigma_{test}}(n_B,n_T) = \frac{d_x}{\text{Tr}(\Sigma_{test})}\frac{(d_x - d_{\ell_{H^*}})^2}{\mathbb E_{X^{(B)},R^{(B)}}
\left[
\|\beta^*_{\hat A\perp}\|^2
\right]}\Big[
\frac{
\mathbb E_{X^{(B)},R^{(B)}}
\left[
\beta_{\hat A\perp}^{*T}
\Sigma_{test}
\beta_{\hat A\perp}^*
\right]
}
{
\mathbb E_{X^{(B)},R^{(B)}}
\left[
\|\beta^*_{\hat A\perp}\|^2
\right]
}
\]
\[
-
2
\frac{
\mathbb E_{X^{(B)},R^{(B)}}
\left[
\mathrm{Tr}(P_{\hat A\perp}\Sigma_{test})
\right]
}
{
d_x-d_{\ell_{H^*}}
}
\Big] + o_{n_T}(1)
\]
\[
= \frac{d_x}{\text{Tr}(\Sigma_{test})}V_{I}(n_B)\left[
\frac{
\mathbb E_{X^{(B)},R^{(B)}}
\left[
\beta_{\hat A\perp}^{*T}
\Sigma_{test}
\beta_{\hat A\perp}^*
\right]}{\mathbb E_{X^{(B)},R^{(B)}}
\left[
\|\beta^*_{\hat A\perp}\|^2
\right]}
-
2
\frac{
\mathbb E_{X^{(B)},R^{(B)}}
\left[
\mathrm{Tr}(P_{\hat A\perp}\Sigma_{test})
\right]
}
{
d_x-d_{\ell_{H^*}}
}\right] + o_{n_T}(1)
\]
\[
= \frac{d_x}{\text{Tr}(\Sigma_{test})}V_{I}(n_B)\Big[2
\frac{\text{Tr}(\Sigma_{A^*\perp})-\frac{1}{n_B}\left[\text{Tr}(\Sigma_{A^*\perp})\text{Tr}(\Sigma_{est})-\text{Tr}(\Sigma_{A^*}\Sigma_{est})(d_x-d_{\ell_{H^*}})\right]+o(n_B^{-1})}{d_x-d_{\ell_{H^*}}}
\]
\[
-\frac{
\gamma_{\Sigma_{test}}(n_B)+o(n_B^{-1})}{\gamma_{I}(n_B)+o(n_B^{-1})}\Big] +o_{n_T}(1)
\]
This is the most explicit form of the value function. However, its not the most interpretable.
Expanding the value multiplicative term to first order:
call 
\[
s_{\beta^*_{A^*\perp}}
=
\frac{\beta_{A^*\perp}^{*T}\Sigma_{A^*\perp}\beta^*_{A^*\perp}}{\|\beta^*_{A^*\perp}\|^2},
\qquad
\bar s
=
\frac{\mathrm{Tr}(\Sigma_{A^*\perp})}{d_x-d_{\ell_{H^*}}}
\]
\[
V_{\Sigma_{test}}(n_B)
=
V_I(n_B)
\frac{d_x}{\mathrm{Tr}(\Sigma_{test})}
\left[
2\bar s
-
s_{\beta^*_{A^*\perp}}
\right.
+
\frac{1}{n_B}
\Big[
\mathrm{Tr}(\Sigma_{A^*}\Sigma_{est})
+
\left(
s_{\beta^*_{A^*\perp}}
-
2\bar s
\right)
\mathrm{Tr}(\Sigma_{est})
\]
\[
-
(d_x-d_{\ell_{H^*}})
\frac{\beta^{*T}\Sigma_{est}\beta^*}{\|\beta^*_{A^*\perp}\|^2}
\left(
\bar s
-
s_{\beta^*_{A^*\perp}}
\right)\Big]
 +
o(n_B^{-1})
\Big]+o_{n_T}(1)
\]
Writing as an exchange rate $V_{\Sigma_{test}}(n_B) = \rho_{\Sigma_{test}}(n_B)n_{B}$
\[
\rho_{\Sigma_{test}}(n_B) = \rho_{I}(n_B)\frac{d_x}{\mathrm{Tr}(\Sigma_{test})}
\left[
2\bar s
-
s_{\beta^*_{A^*\perp}}
\right.
+
\frac{1}{n_B}
\Big[
\mathrm{Tr}(\Sigma_{A^*}\Sigma_{est})
+
\left(
s_{\beta^*_{A^*\perp}}
-
2\bar s
\right)
\mathrm{Tr}(\Sigma_{est})
\]
\[
-
(d_x-d_{\ell_{H^*}})
\frac{\beta^{*T}\Sigma_{est}\beta^*}{\|\beta^*_{A^*\perp}\|^2}
\left(
\bar s
-
s_{\beta^*_{A^*\perp}}
\right)\Big]
 +
o(n_B^{-1})
\Big]+o_{n_T}(1)
\]
\end{lemma}
\begin{lemma}[Balanced Test Distribution Brain Value]\label{lem:brain_value_balanced_robustness}
Under the condition that the test input distribution is \textit{balanced} $\bar s = s_{\beta^*_{A^*\perp}}$,
\[
s_{\beta^*_{A^*\perp}}
=
\frac{\beta_{A^*\perp}^{*T}\Sigma_{A^*\perp}\beta^*_{A^*\perp}}{\|\beta^*_{A^*\perp}\|^2},
\qquad
\bar s
=
\frac{\mathrm{Tr}(\Sigma_{A^*\perp})}{d_x-d_{\ell_{H^*}}}
\]
 meaning that the mass placed on the beta direction not captured by the encoding map is the same as the average covariance mass, then
\[
V_{\Sigma_{test}}(n_B)
=
V_I(n_B)
\frac{d_x}{\mathrm{Tr}(\Sigma_{test})}
[
\bar s
+
\frac{1}{n_B}
\Big[
\mathrm{Tr}(\Sigma_{A^*}\Sigma_{est})
-\bar s
\mathrm{Tr}(\Sigma_{est})\Big]
 +
o(n_B^{-1})
\Big]+o_{n_T}(1)
\]
And the exchange rate:
\[
\rho_{\Sigma_{test}}(n_B)
=
\rho_I(n_B)
\frac{d_x}{\mathrm{Tr}(\Sigma_{test})}
[
\bar s
+
\frac{1}{n_B}
\Big[
\mathrm{Tr}(\Sigma_{A^*}\Sigma_{est})
-\bar s
\mathrm{Tr}(\Sigma_{est})\Big]
 +
o(n_B^{-1})
\Big]+o_{n_T}(1)
\]
\end{lemma}
\begin{theorem}[Isotropic Value is From Nullspace]\label{thm:isotropic_value_nullspace}
From \cref{lem:robustness_value},
When $\Sigma_{test} = P_{A^*}$
\[
V_{P_{A^*}}(n_B,n_T) = V_{I}(n_B,n_T)\frac{d_x}{d_{\ell_{H^*}}}[\frac{1}{n_B}\text{Tr}(\Sigma_{A^*}\Sigma_{est}) + o(n_B^{-1})]
\]
And 
\[
V_{P_{A^*\perp}}(n_B,n_T) = V_{I}(n_B)\frac{d_x}{d_x - d_{\ell_{H^*}}}[1 - \frac{1}{n_B}\text{Tr}(\Sigma_{est}) + o(n_B^{-1})] + o_{n_T}(1)
\]
And note also from \cref{lem:robustness_value} that to this same order,
\[
V_{I}(n_B,n_T) = V_{P_{A^*}+P_{A^*\perp}}(n_B,n_T)  = \frac{d_{\ell_{H^*}}}{d_x}V_{P_{A^*}}(n_B,n_T) + \frac{d_x - d_{\ell_{H^*}}}{d_x}V_{P_{A^*\perp}}(n_B,n_T)
\]
Since $V_{P_{A^*}}$ vanishes at large $n_B$, then the value of brain data comes from the nullspace value.
\end{theorem}
\begin{theorem}[On subspace scaling]\label{thm:robustness_on_subspace}
\cref{lem:BEFS_1st_stage_P_hat_A_perp_trace_scaling_law}
\[\mathbb E_{X^{(B)},R^{(B)}}[\text{Tr}(P_{\hat A\perp}\Sigma_{A^*})]=\frac{d_x-d_{\ell_{H^*}}}{n_B}\left[\text{Tr}(\Sigma_{A^*}\Sigma_{est})\right]+o(n_B^{-1})\]
\cref{lem:BEFS_1st_stage_beta_norm_scaling}
\[
\mathbb E_{X^{(B)},R^{(B)}}
\left[
\beta_{\hat A\perp}^{*T}
\Sigma_{A^*}
\beta_{\hat A\perp}^*
\right] = \frac{1}{n_B }\left[(\|\beta^*_{A^*\perp}\|^2\text{Tr}(\Sigma_{A^*}\Sigma_{est}) \right] + o(n_B^{-1})
\]
And under isotropic test, $\Sigma_{A^*} = P_{A^*}$ and $\Sigma_{A^*\perp} =(I -  P_{A^*})$ the leading scaling law becomes
\[
\mathbb E_{X^{(B)},R^{(B)}}
\left[
\|\beta_{\hat A\perp}^{*T}\|^2
\right] = \|\beta_{A^*\perp}\|^2 + \frac{1}{n_B}\left[( \beta^{*T}\Sigma_{est}\beta^{*}(d_x - d_{\ell_{H^*}}) - \|\beta^*_{A^*\perp}\|^2\text{Tr}(\Sigma_{est})\right] + o(n_B^{-1})
\]
\[
= \gamma_I+ o(n_B^{-1})
\]
\[
\varepsilon_{\Sigma_{test}}^{BEFS}(n_B,n_T,\lambda_{opt})
=
\varepsilon_{\Sigma_{test}}^{TOS}(n_T)
\]
\[+
\frac{\sigma_y^4(d_x-d_{\ell_{H^*}})^2}
{
n_T^2(\gamma_{I}(n_B)+ o(n_B^{-1}))
}\left[
\frac{\mathrm{Tr}(\Sigma_{A^*}\Sigma_{est})}{n_B}
\left(
\frac{
\|\beta^*_{A^*\perp}\|^2
}
{
(\gamma_{I}(n_B)+ o(n_B^{-1}))
}
-
2
\right)\right]
+
o(n_T^{-2}) + o(n_B^{-1}n_T^{-2})
\]
Note that this decays in $n_B$. In the large brain data regime,
\[
\frac{\mathrm{Tr}(\Sigma_{A^*}\Sigma_{est})}{n_B}
\left(
\frac{
\|\beta^*_{A^*\perp}\|^2
}
{
(\gamma_{I}(n_B)+ o(n_B^{-1}))
}
-
2
\right) =
-
\frac{1}{n_B}
\mathrm{Tr}(\Sigma_{A^*}\Sigma_{est})
+
o(n_B^{-1})
\]
So the correction is a vanishing but negative sign correction in large $n_B$. In the infinite brain data limit,
\[
\lim_{n_B\to \infty}\varepsilon_{\Sigma_{test}}^{BEFS}(n_B,n_T,\lambda_{opt}) = \varepsilon_{\Sigma_{test}}^{TOS}(n_T)
\]
\end{theorem}
\begin{theorem}[Off Subspace Scaling]\label{thm:robustness_off_subspace}
\cref{lem:BEFS_1st_stage_beta_norm_scaling}
\[\mathbb E_{X^{(B)},R^{(B)}}[\text{Tr}(P_{\hat A\perp}\Sigma_{A^*\perp})]=\text{Tr}(\Sigma_{A^*\perp})-\frac{1}{n_B}\text{Tr}(\Sigma_{A^*\perp})\text{Tr}(\Sigma_{est})+o(n_B^{-1})\]
\cref{lem:BEFS_1st_stage_beta_norm_scaling}
\[
\mathbb E_{X^{(B)},R^{(B)}}
\left[
\beta_{\hat A\perp}^{*T}
\Sigma_{A^*\perp}
\beta_{\hat A\perp}^*
\right] = \beta_{A^*\perp}^{*T}\Sigma_{A^*\perp}\beta^*_{A^*\perp} 
\]
\[
+ \frac{1}{n_B }\big[ - 2\beta_{A^*\perp}^{*T}\Sigma_{A^*\perp}\beta_{A^*\perp}\text{Tr}(\Sigma_{est}) + \beta^{*T}\Sigma_{est}\beta^{*}\text{Tr}(\Sigma_{A^*\perp})\big] + o(n_B^{-1})
\]
\[
\mathbb E_{X^{(B)},R^{(B)}}
\left[
\|\beta_{\hat A\perp}^{*T}\|^2
\right] = \|\beta_{A^*\perp}\|^2 + \frac{1}{n_B}\left[( \beta^{*T}\Sigma_{est}\beta^{*}(d_x - d_{\ell_{H^*}}) - \|\beta^*_{A^*\perp}\|^2\text{Tr}(\Sigma_{est})\right] + o(n_B^{-1})
\]
\[
\varepsilon_{\Sigma_{A^*\perp}}^{BEFS}(n_B,n_T,\lambda_{opt})
=
\varepsilon_{\Sigma_{A^*\perp}}^{TOS}(n_T)+\frac{\sigma_y^4(d_x-d_{\ell_{H^*}})^2}
{
n_T^2(\gamma_{I}(n_B)+ o(n_B^{-1}))
}
\]
\[
\left[
\frac{
\beta_{A^*\perp}^{*T}\Sigma_{A^*\perp}\beta^*_{A^*\perp}
+
\frac{1}{n_B}
\left[
-2\beta_{A^*\perp}^{*T}\Sigma_{A^*\perp}\beta^*_{A^*\perp}\mathrm{Tr}(\Sigma_{est})
+
\beta^{*T}\Sigma_{est}\beta^*\mathrm{Tr}(\Sigma_{A^*\perp})
\right]
+
o(n_B^{-1})
}
{
\|\beta^*_{A^*\perp}\|^2
+
\frac{1}{n_B}
\left[
\beta^{*T}\Sigma_{est}\beta^*(d_x-d_{\ell_{H^*}})
-
\|\beta^*_{A^*\perp}\|^2\mathrm{Tr}(\Sigma_{est})
\right]
+
o(n_B^{-1})
}
\right.
\]
\[
\left.
-
2
\frac{
\mathrm{Tr}(\Sigma_{A^*\perp})
-
\frac{1}{n_B}\mathrm{Tr}(\Sigma_{A^*\perp})\mathrm{Tr}(\Sigma_{est})
+
o(n_B^{-1})
}
{
d_x-d_{\ell_{H^*}}
}
\right]
+
o(n_T^{-2})
\]
Note that this does not vanish in $n_B$. Taking large $n_B$,
\[
s_{\beta^*_{A^*\perp}}
=
\frac{\beta_{A^*\perp}^{*T}\Sigma_{A^*\perp}\beta^*_{A^*\perp}}{\|\beta^*_{A^*\perp}\|^2},
\qquad
\bar s
=
\frac{\mathrm{Tr}(\Sigma_{A^*\perp})}{d_x-d_{\ell_{H^*}}}
\]
Then the inner expression simplifies to:
\[
s_{\beta^*_{A^*\perp}}-2\bar s
+
\frac{(d_x-d_{\ell_{H^*}})}{n_B}
\left[
\frac{\beta^{*T}\Sigma_{est}\beta^*}{\|\beta^*_{A^*\perp}\|^2}
\left(
\bar s-s_{\beta^*_{A^*\perp}}
\right)
+
\frac{\mathrm{Tr}(\Sigma_{est})}{d_x-d_{\ell_{H^*}}}
\left(
2\bar s-s_{\beta^*_{A^*\perp}}
\right)
\right]
+
o(n_B^{-1}).
\]
Which in large $n_B$ has a negative sign if the missing direction is not overly represented in the test covariance. If the test is balanced such that $\bar s = s_{\beta^*_{A^*\perp}}$,
\[
\varepsilon_{\Sigma_{A^*\perp}}^{BEFS}(n_B,n_T,\lambda_{opt})
=
\varepsilon_{\Sigma_{A^*\perp}}^{TOS}(n_T)
\]
\[
+\frac{\sigma_y^4(d_x-d_{\ell_{H^*}})^2}
{
n_T^2(\gamma_{I}(n_B)+ o(n_B^{-1}))
}\left[\bar s\left(\frac{\text{Tr}(\Sigma_{est})}{n_B} -1 \right)\right]
+
o(n_T^{-2}n_B^{-1}) + o(n_T^{-2})
\]
So the sign becomes negative when $n_B > \text{Tr}(\Sigma_{est})$. And in the infinite brain data limit, 
\[
\lim_{n_B \to \infty}\varepsilon_{\Sigma_{A^*\perp}}^{BEFS}(n_B,n_T,\lambda_{opt})
=
\varepsilon_{\Sigma_{A^*\perp}}^{TOS}(n_T)+\frac{\sigma_y^4(d_x-d_{\ell_{H^*}})\mathrm{Tr}(\Sigma_{A^*\perp})}
{
n_T^2\|\beta_{A^*\perp}\|^2
} + o(n_T^{-2})
\]
However, if $s_{B^*_{A^*\perp}} > 2\bar s$, the sign becomes negative and brain data contributes an asymptotically negative equivalent task data samples.
\end{theorem}
\subsection{BEFS Budget Scaling}
\begin{theorem}[Budget Scaling]\label{thm:budget_scaling}
\[
\varepsilon_{I}^{BEFS}(n^{opt}_B, n_{T}^{opt}|\mathcal{B}) = \min_{n_B, n_T}\varepsilon^{BEFS}_{I}(n_B,n_T,\lambda_{opt}(n_T)) \quad c_Bn_B + c_Tn_T \leq \mathcal{B}
\]
\[
\varepsilon^{BEFS}_{I}(n_B,n_T,\lambda_{opt})
=
\sigma_{test}^2
+
\frac{\sigma_y^2 d_x}{n_T - d_x -1}
-
\frac{\sigma_y^4(d_x-d_{\ell_{H^*}})^2}{\gamma_I(n_B)}\frac{1}{n_T^2}
+
o(n_T^{-2})+ o(n_T^{-2}n_B^{-1})
\]
Taking the continuous relaxation of the problem, \[
0 < n_T \le \frac{\mathcal{B}}{c_2},
\qquad
n_B = \frac{\mathcal{B}-c_2 n_T}{c_1}\ge 0
\]
\[
\gamma_{I}(n_B|B,n_T) = \|\beta_{A^*\perp}\|^2 + \frac{c_1}{\mathcal{B}-c_2n_T}\left[( \beta^{*T}\Sigma_{est}\beta^{*}(d_x - d_{\ell_{H^*}}) - \|\beta^*_{A^*\perp}\|^2\text{Tr}(\Sigma_{est})\right]
\]
So the total optimization becomes
\[
\varepsilon^{BEFS}_{I}(n_B,n_T|\mathcal{B},\lambda_{opt}(n_T)) = \sigma_{test}^2 
+
\frac{\sigma_y^2 d_x}{n_T - d_x -1}
-
\frac{\sigma_y^4(d_x-d_{\ell_{H^*}})^2}{\gamma_I(n_B|\mathcal{B},n_T)}\frac{1}{n_T^2}
+
o(n_T^{-2})+ o(n_T^{-2}(\mathcal{B} - c_Tn_T)^{-1})
\]
Call $z = c_Bn_B$, then $n_T = (\mathcal{B} - z)/c_T$
\[
\gamma_{I}(z) = \|\beta_{A^*\perp}\|^2 + \frac{c_B}{z}\left[( \beta^{*T}\Sigma_{est}\beta^{*}(d_x - d_{\ell_{H^*}}) - \|\beta^*_{A^*\perp}\|^2\text{Tr}(\Sigma_{est})\right]
\]
then 
\[
\varepsilon^{BEFS}_{I}(n_T, n_B|\lambda_{opt}(n_T), \mathcal{B} )
=
\frac{\sigma_y^2d_xc_T}{\mathcal{B}-z-c_T(d_x+1)}
\]
\[
-
\frac{\sigma_y^4(d_x-d_{\ell_{H^*}})^2 c_T^2}{(\mathcal{B}-z)^2}
\frac{1}{ \|\beta_{A^*\perp}\|^2 + \frac{c_B}{z}\left[( \beta^{*T}\Sigma_{est}\beta^{*}(d_x - d_{\ell_{H^*}}) - \|\beta^*_{A^*\perp}\|^2\text{Tr}(\Sigma_{est})\right]}
\]
\[
+
o((\mathcal{B}-z)^{-2}) + o(z^{-1}(\mathcal{B}-z)^{-2})
\]
Then clearly $z = o(B)$ in order to drop the risk asymptotically. Now, operating in the $z = o(B)$ regime, 
\[
\frac{1}{\mathcal{B}-z-c_T(d_x+1)}
=
\frac{1}{\mathcal{B}}
+
\frac{z+c_T(d_x+1)}{\mathcal{B}^2}
+
o(\mathcal{B}^{-2}).\]
\[
\frac{1}{(\mathcal{B}-z)^2}
=
\frac{1}{\mathcal{B}^2} +o(\mathcal{B}^{-2}).
\]
\[
\varepsilon^{BEFS}_{I}(n_T, n_B|\lambda_{opt}(n_T), \mathcal{B} )
=
\sigma_{test}^2
+
\frac{\sigma_y^2 d_x c_T}{\mathcal{B}}
\]
\[
+
\frac{1}{\mathcal{B}^2}
\left[
\sigma_y^2 d_xc_Tz+\sigma^2_yd_xc^2_T(d_x+1)
-
\frac{\sigma_y^4(d_x-d_{\ell_{H^*}})^2 c_T^2}{\|\beta_{A^*\perp}\|^2 + \frac{c_B}{z}\left[( \beta^{*T}\Sigma_{est}\beta^{*}(d_x - d_{\ell_{H^*}}) - \|\beta^*_{A^*\perp}\|^2\text{Tr}(\Sigma_{est})\right]}
\right]
+
o(\mathcal{B}^{-2})
\]
Rewriting back into $n_B$,
\[
\varepsilon^{BEFS}_{I}(n_T, n_B|\lambda_{opt}(n_T), \mathcal{B} )
=
\sigma_{test}^2
+
\frac{\sigma_y^2 d_x c_T}{\mathcal{B}}
\]
\[
+\frac{1}{\mathcal{B}^2}
\left[
\sigma_y^2 d_xc_Tc_Bn_B+\sigma^2_yd_xc^2_T(d_x+1)
-
\frac{n_B\sigma_y^4(d_x-d_{\ell_{H^*}})^2 c_T^2}{n_B\|\beta_{A^*\perp}\|^2 + ( \beta^{*T}\Sigma_{est}\beta^{*}(d_x - d_{\ell_{H^*}}) - \|\beta^*_{A^*\perp}\|^2\text{Tr}(\Sigma_{est})}
\right]
+
o(\mathcal{B}^{-2})
\]
Minimizing over $n_B$, take the equation:
\[
f(n_B) = \kappa_1n_B - \frac{\kappa_2n_B}{\kappa_3n_B + \kappa_4}
\]
Differentiating:
\[
\frac{df(n_B)}{d n_B} = \kappa_1 - \frac{\kappa_2\kappa_4}{(\kappa_3n_B + \kappa_4)^2}
\]
Solving for the minimum
\[
n_B^* = \frac{1}{\kappa_3}\left(\sqrt\frac{\kappa_2\kappa_4}{\kappa_1} - \kappa_4\right)
\]
Which is greater than zero when 
\[
 c_B
< c_T\left(\frac{d_x - d_{\ell_{H^*}}}{d_x}\right)
\frac{\sigma_y^2}{\left[\beta^{*T}\Sigma_{est}\beta^{*} - \|\beta^*_{A^*\perp}\|^2\frac{\text{Tr}(\Sigma_{est})}{d_x - d_{\ell_{H^*}}}\right]}
\]
Giving the expression:
\[
n_B^{opt}
=
\frac{1}{\|\beta^*_{A^*\perp}\|^2}
\big[
\sigma_y(d_x-d_{\ell_{H^*}})
\sqrt{
\frac{c_T}{d_x c_B}\ ( \beta^{*T}\Sigma_{est}\beta^{*}(d_x - d_{\ell_{H^*}}) - \|\beta^*_{A^*\perp}\|^2\text{Tr}(\Sigma_{est})
}
\]
\[
- ( \beta^{*T}\Sigma_{est}\beta^{*}(d_x - d_{\ell_{H^*}}) - \|\beta^*_{A^*\perp}\|^2\text{Tr}(\Sigma_{est})
\big] + o(1)
\]
Plugging in $n_B^{opt}$ into $f(n_B)$, 
\[
\varepsilon_I^{BEFS}(n_B^{opt},n_T^{opt}|\mathcal{B})
=
\sigma_{test}^2
+
\frac{\sigma_y^2 d_x c_T}{\mathcal{B}}
+ \frac{\sigma^2_yd_xc^2_T(d_x+1)}{ \mathcal{B}^2}
-
\frac{1}{\|\beta^*_{A^*\perp}\|^2 \mathcal{B}^2}
\Big(
\sigma_y^2(d_x-d_{\ell_{H^*}})c_T
\]
\[
-
\sigma_y\sqrt{d_x c_B c_T ( \beta^{*T}\Sigma_{est}\beta^{*}(d_x - d_{\ell_{H^*}}) - \|\beta^*_{A^*\perp}\|^2\text{Tr}(\Sigma_{est})}
 \Big)^2
+
o(\mathcal{B}^{-2})
\]
\end{theorem}
\begin{theorem}[Effective Extra TOS Budget From Brain Data]\label{thm:extra_budget}
Under a fixed budget in a continuous relaxed TOS scaling:
\[
\varepsilon_I^{TOS}(n^{opt}_T|\mathcal{B}) = \varepsilon_I^{TOS}(\mathcal{B}/c_T)
=
\sigma_{test}^2
+
\frac{\sigma_y^2d_x}{\mathcal{B}/c_T-d_x-1}
\]
So adding a fixed amount to the budget $\Delta B$ under large budget gives the quadratic correction:
\[
\varepsilon_{I}^{BEFS}(n^{opt, BEFS}_{T}, n^{opt, BEFS}_B |\mathcal{B}) = \varepsilon_I^{TOS}(n^{opt, TOS}_T |\mathcal{B} + \Delta \mathcal{B})= \varepsilon_I^{TOS}\left(\mathcal{B}/c_T + \frac{\Delta \mathcal{B}}{c_T}\right)
\]
\[
=\sigma_{test}^2
+
\frac{\sigma_y^2d_xc_T}{\mathcal{B}}
+
\frac{\sigma_y^2d_xc_T^2(d_x+1)}{\mathcal{B}^2}
-
\frac{\sigma_y^2d_xc_T^2}{\mathcal{B}^2}\left(\frac{\Delta \mathcal{B}}{c_T}\right)
+
o(\mathcal{B}^{-2})
\]
Equating to \cref{thm:budget_scaling} and solving for $\Delta \mathcal{B}$ using the same argument as \cref{lem:value},
\[
\Delta \mathcal{B}
=
c_T\frac{\sigma_y^2(d_x-d_{\ell_{H^*}})^2}{d_x\,\|\beta^*_{A^{*\perp}}\|^2}
\left[
1-
\sqrt{ \frac{c_B}{c_T} \frac{d_x}{d_x - d_{\ell_{H^*}}}\frac{1}{\sigma_y^2}
\left(
\beta^{*T}\Sigma_{est}\beta^*
-
\|\beta^*_{A^{*\perp}}\|^2\frac{\mathrm{Tr}(\Sigma_{est})}{d_x-d_{\ell_{H^*}}}
\right)
}
\right]^2 + o_{\mathcal{B}}(1)
\]
\end{theorem}
\subsection{BEFS- Hard Constraint}
\begin{lemma}[BEFS-Second Stage Hard Constraint]\label{lem:BEFS_2nd_stage_hard_beta} Suppose we have a fixed map $\hat A$ and we want to learn a task map estimator restricted to being on top of $\hat A$.
\[
\hat w^{BEFS, Hard} = \text{argmin}_{w} \frac{1}{n}\|y^{(T)} - X^{(T)}\hat Aw\|^2
\]
Such that $\hat \beta^{BEFS, Hard} = \hat A\hat w^{BEFS, Hard}$.
Clearly this is an OLS problem. Let $Z = X^{(T)}\hat A$, then this has the OLS solution. 
\[
\hat w^{BEFS, Hard} = (Z^TZ)Z^Ty
\]
$\beta = (I - P_{\hat A})\beta + P_{\hat A}\beta$
\[
y^{(T)} = X^{(T)}\beta^* + e_y  = X^{(T)}(I - P_{\hat A})\beta^* + X^{(T)}P_{\hat A}\beta^* + e_{y}
\]
\[
\hat \beta^{BEFS, Hard} = \frac{1}{n_T}\hat A(\hat A^T\hat\Sigma \hat A)^{-1}\hat A^TX^{(T)T}y^{(T)}
\]
Define $w^*_{\hat A} $ as $ P_{\hat A}\beta^* = \hat Aw^*_{\hat A}$
\[
= \frac{1}{n_T}\hat A(\hat A^T\hat\Sigma \hat A)^{-1}\hat A^TX^{(T)T}(X^{(T)}(\beta^* -  P_{\hat A}\beta^*) +  X^{(T)}\hat A w_{\hat A} + e_y)
\]
\[
= P_{\hat A}\beta^* + \hat A(\hat A^T\hat\Sigma \hat A)^{-1}\hat A^T\hat \Sigma (I - P_{\hat A})\beta^*  + \frac{1}{n_T}\hat A(\hat A^T\hat\Sigma \hat A)^{-1}\hat A^T X^{(T)T}e_y
\]
Note that this means $\hat \beta$ is biased since 
\[
\hat \beta - \beta^* = -(I - P_{\hat A})\beta^* + \hat A(\hat A^T\hat\Sigma \hat A)^{-1}\hat A^T\hat \Sigma (I - P_{\hat A})\beta^*  + \frac{1}{n_T}\hat A(\hat A^T\hat\Sigma \hat A)^{-1}\hat A^T X^Te_y
\] and at high samples the second terms vanish. 
\end{lemma}
\begin{lemma}[BEFS - Hard Constraint Scaling Law]\label{lem:BEFS_2nd_stage_soft_scaling}
Assume $x_{test} \sim N(0, \Sigma_{test})$ and $y_{test} = x_{test}^T\beta^* + \eta_{test}$ for $\eta_{test} \sim N(0,\sigma_{test}^2)$. We want to solve (for independent $\hat A$):
\[
\hat \beta^{BEFS, Hard} = \hat A\hat w , \quad \mathbb{E}_{e_y, X, y_{test}, x_{test}}[ \|y_{test} - x_{test}^T\hat\beta^{BEFS, Hard}\|^2 |\hat{A}] 
\]
Taking the expectation over the test distribution:
\[
\mathbb{E}_{y_{test}, x_{test}}[ \|y_{test} - x_{test}^T\hat\beta\|^2 |\hat{A}, e_y, X^{(T)}]  = (\hat\beta - \beta)^T\Sigma_{test}(\hat\beta - \beta) + \sigma^2_{test}
\]
Call $( I - P_{\hat A})\beta^* = \beta^*_{\hat A \perp}$
\[
\hat \beta^{BEFS, Hard} - \beta = -\beta_{\hat A \perp} + \frac{1}{n}\hat{A}(\hat{A}^T\hat\Sigma\hat{A})^{-1}\hat{A}^TX^{(T)T}(e_y + X^{(T)}\beta_{\hat A \perp})
\]
Call $Z =X^{(T)}\hat A$, then $x_{i}^T\hat A$ is independent from $x_{i}^T\beta_{\hat A \perp}$ because $\beta_{\hat A \perp}^T\hat A = 0$ and $x_i$ is gaussian. So call 
\[
F(Z) = \hat A(Z^TZ)^{-1}Z^T
\]
\[
\mathbb{E}_{y_{test}, x_{test}}[ \|y_{test} - x_{test}^T\hat\beta^{BEFS, Hard}\|^2 |\hat{A}, e_y, X^{(T)}] = \beta_{\hat A \perp}^{*T}\Sigma_{test}\beta^*_{\hat A \perp}
\]
\[
+ (e_y + X^{(T)}\beta^*_{\hat A \perp})^TF(Z)^T\Sigma_{test}F(Z)(e_y + X^{(T)}\beta^*_{\hat A \perp})
\]
\[
+ \beta_{\hat A \perp}^{*T}X^{(T)T}\Sigma_{test}F(Z)(e_{y} + X^{(T)}\beta^*_{\hat A \perp}) + (e_y + X^{(T)}\beta^*_{\hat A \perp})^TF(Z)^T\Sigma_{test}X^{(T)}\beta^*_{\hat A\perp}
\]
Taking the expectation over $e_y, X^{(T)}$, the last terms drop because $X^{(T)}\beta^*_{\hat A \perp}$ and $e_y$ are mean zero. 
\[
\mathbb{E}_{y_{test}, x_{test}, X,e_{y}}[ \|y_{test} - x_{test}^T\hat\beta\|^2 | \hat A, Z] = \beta_{\hat A \perp}^{*T}\Sigma_{test}\beta^*_{\hat A \perp}
\]
\[
 + \mathbb{E}_{X,e_y}[(e_y + X^{(T)}\beta^*_{\hat A \perp})^TF(Z)^T\Sigma_{test}F(Z)(e_y + X^{(T)}\beta^*_{\hat A \perp}) | \hat A, Z]
\]
Using the expectation of a quadratic form:
\[
 \mathbb{E}_{X^{(T)},e_y}[(e_y + X^{(T)}\beta^*_{\hat A \perp})^TF(Z)^T\Sigma_{test}F(Z)(e_y + X^{(T)}\beta^*_{\hat A \perp}) | \hat A, Z] 
\]
Since $X^{(T)}, e_y$ are independent
\[
=\text{Tr}(F(Z)^T\Sigma_{test}F(Z)\mathbb{E}_{e_y, X^{(T)}}[(e_y + X^{(T)}\beta_{\hat A \perp})(e_y + X^{(T)}\beta_{\hat A \perp})^T] = (\sigma^2_{y} +\|\beta^*_{\hat A \perp}\|^2)\text{Tr}(F(Z)^T\Sigma_{test}F(Z))
\]
Finally, taking the expectation on $Z$,
\[
\mathbb{E}_Z[\text{Tr}(F(Z)^T\Sigma_{test}F(Z))] = \text{Tr}\left(\Sigma_{test}\mathbb{E}_Z[F(Z)F(Z)^T]\right) = \text{Tr}(\Sigma_{test}\hat A\mathbb{E}_{Z}[(Z^TZ)^{-1}]\hat A^T)
\]
$Z_{i} = X^{(T)}_{i}\hat A$ so $Z_{i} \sim N(0, \hat A^T \hat A)$ and $Z^TZ\sim \text{Wishart}(\hat A^T \hat A, n_T)$ so $(Z^TZ)^{-1} \sim \text{Inv-Wishart}((\hat A^T \hat A)^{-1}, n_T)$ which has expectation $\frac{1}{n_T - \hat d_{\ell_{H^*}} -1}(\hat A^T \hat A)^{-1}$.
\[
\text{Tr}(\Sigma_{test}\hat A\mathbb{E}_{Z}[(Z^TZ)^{-1}]\hat A^T) = \frac{1}{n_T - \hat d_{\ell_{H^*}} -1}\text{Tr}(\Sigma_{test} P_{\hat A})
\]
So the total scaling is given by: 
\[
\mathbb{E}_{e_y, X^{(T)}, y_{test}, x_{test}}[ \|y_{test} - x_{test}^T\hat\beta^{BEFS, Hard}\|^2 |\hat{A}] = \beta_{\hat A\perp}^{*T}\Sigma_{test}\beta^*_{\hat A\perp} + \frac{(\sigma_y^2 + \|\beta^*_{\hat A\perp}\|^2)}{n_T - \hat d_{\ell_{H^*}} -1}\text{Tr}(\Sigma_{test} P_{\hat A})
\]
Taking the wishart denominator to first order in large $n_T$,
\[
\mathbb{E}_{e_y, X^{(T)}, y_{test}, x_{test}}[ \|y_{test} - x_{test}^T\hat\beta^{BEFS, Hard}\|^2 |\hat{A}] = \beta_{\hat A\perp}^{*T}\Sigma_{test}\beta^*_{\hat A\perp} + \frac{\sigma_y^2 +\|\beta^*_{\hat A\perp}\|^2}{n_T}\text{Tr}(\Sigma_{test} P_{\hat A}) + o(n^{-1}_T)
\]
\end{lemma}
\begin{theorem}[Large $\lambda$ BEFS Scales as $BEFS-Hard$]\label{thm:BEFS_hard_lambda}
    From \cref{thm:befs_stage_2_scaling}, for fixed constant $\lambda$, $\alpha = \frac{1}{1+\lambda}$ and the wishart denominator pushed into the remainder:
    \[
    \mathbb{E}_{y_{test}, x_{test}, e_y,X^{(T)}}[ \|y_{test} - x_{test}^T\hat\beta^{BEFS}\|^2 |\hat{A}] = 
\sigma_{test}^2 + (1-\alpha)^2\beta^{*T}_{\hat A\perp}\Sigma_{test}\beta^{*}_{\hat A\perp}\left(1+ \frac{2\alpha\text{Tr}(J_{\hat A}) + 3\alpha^2}{n_T}\right)
\]
\[
+ \frac{\sigma_y^2 + (1-\alpha)^2\|\beta^*_{\hat A\perp}\|^2}{n_T}\text{Tr}(J_{\hat A}\Sigma_{test}J_{\hat A})+ o(n_T^{-1})
\]
Taking $\lambda$ large such that $\alpha \approx 0$
\[
\mathbb{E}_{e_y, X^{(T)}, y_{test}, x_{test}}[ \|y_{test} - x_{test}^T\hat\beta^{BEFS}\|^2 |\hat{A}]\approx \beta_{\hat A\perp}^{*T}\Sigma_{test}\beta^*_{\hat A\perp} + \frac{\sigma_y^2 +\|\beta^*_{\hat A\perp}\|^2}{n_T}\text{Tr}(\Sigma_{test} P_{\hat A}) + o(n^{-1}_T)
\]
\[
= \mathbb{E}_{e_y, X^{(T)}, y_{test}, x_{test}}[ \|y_{test} - x_{test}^T\hat\beta^{BEFS, Hard}\|^2 |\hat{A}]
\]
\end{theorem}

\end{document}